\PassOptionsToPackage{table,dvipsnames}{xcolor}
\documentclass{ifm-tr}

\usepackage{amsmath,amsfonts,amssymb,mathtools}
\usepackage[utf8]{inputenc}
\usepackage{xspace}
\usepackage{url}
\usepackage{booktabs}
\usepackage{makecell}
\usepackage{xcolor}
\usepackage{adjustbox}
\usepackage{wrapfig}
\newlength{\arxivcolwidth}
\let\arxivfigure\figure \let\endarxivfigure\endfigure
\renewenvironment{figure}[1][tbp]%
  {\arxivfigure[#1]\centering\begin{minipage}{\arxivcolwidth}\setlength{\columnwidth}{\linewidth}}%
  {\end{minipage}\endarxivfigure}
\AtBeginEnvironment{table}{\centering}

\makeatletter
\def\arxiv@inline#1{\def\@captype{#1}\xdef\arxiv@type{#1}\@ifnextchar[{\arxiv@gobbleopt}{}}
\def\arxiv@gobbleopt[#1]{}
\newcommand\arxiv@inlinefloats{%
  \def\figure{\arxiv@inline{figure}}\let\endfigure\relax
  \def\table{\arxiv@inline{table}}\let\endtable\relax
  \setlength{\columnwidth}{\linewidth}}
\newcommand\arxiv@pairshrink{%
  \renewcommand\footnotesize{\fontsize{8}{9.5}\selectfont}%
  \renewcommand\scriptsize{\fontsize{7}{8}\selectfont}%
  \renewcommand\tiny{\fontsize{5}{6}\selectfont}}
\newenvironment{floatpair}[3][t]%
  {\arxivfigure[#1]\def\arxiv@wb{#3}\noindent
   \begin{minipage}[t]{#2\textwidth}\vspace{0pt}\arxiv@inlinefloats\arxiv@pairshrink\centering}%
  {\end{minipage}\endarxivfigure}
\newenvironment{fixedpair}[2]%
  {\par\vspace{\intextsep}\begingroup\centering\def\arxiv@wb{#2}%
   \begin{minipage}[t]{#1\textwidth}\vspace{0pt}\arxiv@inlinefloats\arxiv@pairshrink\centering}%
  {\end{minipage}\par\endgroup\vspace{\intextsep}}
\newcommand\pairnext{\end{minipage}\hfill
  \begin{minipage}[t]{\arxiv@wb\textwidth}\vspace{0pt}\arxiv@inlinefloats\arxiv@pairshrink\centering}
\newenvironment{sidecapfloat}[2][t]%
  {\arxivfigure[#1]\centering\begin{minipage}{#2\textwidth}\arxiv@inlinefloats\centering}%
  {\end{minipage}\endarxivfigure}
\newenvironment{wrapside}[2][R]%
  {\wrapfigure{#1}{#2\textwidth}\arxiv@inlinefloats\arxiv@pairshrink\centering}%
  {\endwrapfigure}
\makeatother

\usepackage{cuted} %
\usepackage{colortbl} %
\definecolor{tabzebra}{gray}{0.955}
\definecolor{tabours}{RGB}{229,238,252}
\newcommand{\tabgroup}[2]{\multicolumn{#1}{@{}l}{\textit{\color{black!60}#2}}}
\newcommand{\tabdouble}{\specialrule{0.3pt}{0pt}{1.1pt}\specialrule{0.3pt}{0pt}{0pt}}
\newcommand{\tabdash}[1]{\tabdashi#1\relax}
\def\tabdashi#1-#2\relax{\multispan{#2}\xleaders\hbox to 2.4pt{\hss\vrule height 0.3pt depth 0pt width 1.2pt\hss}\hfill\cr\noalign{\vskip-0.3pt}}
\newcommand{\tabcmidend}{\noalign{\vskip-\cmidrulewidth}}
\newcommand{\tabstyle}{\setlength{\aboverulesep}{0pt}\setlength{\belowrulesep}{0pt}\setlength{\extrarowheight}{1.1pt}\setlength{\arrayrulewidth}{0.3pt}\setlength{\heavyrulewidth}{0.85pt}\setlength{\cmidrulewidth}{0.3pt}}
\usepackage{multirow} %
\usepackage{stfloats} %

\usepackage{currfile} %

\usepackage{caption} %
\newcommand{\figlab}[1]{{\footnotesize #1}}%
\newcommand{\figdesc}[1]{{\footnotesize\color{black!75}#1}}%
\newsavebox{\rimgbox}
\newcommand{\rimg}[2][]{\sbox{\rimgbox}{\includegraphics[#1]{#2}}\begin{tikzpicture}[baseline=0pt]\clip[rounded corners=2.2pt] (0,0) rectangle (\wd\rimgbox,\ht\rimgbox);\node[inner sep=0pt,anchor=south west] at (0,0) {\usebox{\rimgbox}};\end{tikzpicture}}
\newcommand{\figours}[1]{\makebox[0pt]{{\setlength{\fboxsep}{1.2pt}\colorbox{tabours}{\footnotesize\strut #1}}}}%

\usepackage{tikz}
\usetikzlibrary{arrows.meta}

\usepackage{pifont}
\newcommand{\cmark}{\ding{51}}
\newcommand{\xmark}{\ding{55}}

\makeatletter
\DeclareRobustCommand\onedot{\futurelet\@let@token\@onedot}
\def\@onedot{\ifx\@let@token.\else.\null\fi\xspace}
\def\eg{\emph{e.g}\onedot} 
\def\ie{\emph{i.e}\onedot}

\makeatother

\let\cite\citep
\makeatletter
\renewcommand*{\hyper@natlinkbreak}[2]{#1}
\makeatother

\title{Simulator-as-Proxy World Model for Robot Manipulation}

\author[1]{Tingting Liao}
\authorsep{,}
\author[1,2]{Mingjia Huo}
\authorsep{,}
\author[1,3,*]{Hao Li}
\authorsep{,}
\author[1]{Zhengzhong Liu}
\authorsep{,}
\author[1]{Eric P. Xing}
\authorsep{,}
\author[1]{Guangyi Liu}

\affiliation[1]{Mohamed bin Zayed University of Artificial Intelligence}
\affiliationsep{\\}
\affiliation[2]{University of California, San Diego}
\affiliationsep{\\}
\affiliation[3]{Pinscreen}

\metadata[Project Page]{\href{https://sim-pwm.pages.dev/}{\texttt{\color{urlcolor}https://sim-pwm.pages.dev}}}
\metadata[Code]{\href{https://github.com/TingtingLiao/SimPWM}{\texttt{\color{urlcolor}https://github.com/TingtingLiao/SimPWM}}}

\abstract{%
Video world models for robots are increasingly conditioned on a simulated or rendered proxy of the scene, on the assumption that the video will show what the proxy shows.
Testing this assumption requires a proxy whose state is explicit and can be perturbed.
We present \textbf{SimPWM} (\textbf{Sim}ulator-as-\textbf{P}roxy \textbf{W}orld \textbf{M}odel): given a calibrated camera, it represents, from the first frame alone, the robot by its 3D model and the objects the gripper may touch as 3D points; a rule-based simulator with no trained parameters moves them under the actions, and a video model renders the result.
SimPWM returns the realistic video together with the simulated state.
We then freeze the simulated objects while the arm moves as recorded and check whether the video still shows the object lifted.
On DROID episodes from unseen setups, the answer differs for the arm and for the objects.
With each of three base video models, the arm follows the proxy (robot IoU $0.75$--$0.78$, against at most $0.69$ for end-to-end models); rendering the robot alone already gives this.
What the simulated objects add is the object state: against a robot-only proxy, they cut missed pick-ups from $54\%$ to $38\%$, and the video follows them only with some base models: on rigid objects, freezing them lowers the lift rate of MiniMax-H3, fine-tuned on episodes whose simulation agrees with the real video, from $42\%$ to $10\%$, but that of LTX-2.3, trained on the same episodes, only from $57\%$ to $49\%$, as if it inferred the grasp from the arm motion; more than half of the remaining lifts of MiniMax-H3 are of cloth and rope, which the simulator carries rigidly.
With any base model, the simulated state records what the simulation did; under perturbed actions, a video that disagrees with it reveals that the video ignores the simulated state.
}

\begin{document}
\maketitle

\renewcommand{\thefootnote}{\fnsymbol{footnote}}
\footnotetext[1]{Corresponding author.}
\renewcommand{\thefootnote}{\arabic{footnote}}

\section{Introduction}
\label{sec:intro}
\begin{floatpair}[t]{0.485}{0.485}
\begin{figure}[t]
\centering
\newlength{\ptw}\newlength{\pth}\newlength{\ptl}
\setlength{\ptl}{9pt}
\setlength{\ptw}{\dimexpr(\columnwidth-\ptl-6pt)/3\relax}
\setlength{\pth}{0.79\ptw}%
\definecolor{tzbad}{RGB}{192,57,43}
\definecolor{tzok}{RGB}{30,132,73}
\newcommand{\pimg}[4]{\begin{scope}\clip[rounded corners=2pt] (#1,#2) rectangle ++(\ptw,-\pth);\node[anchor=north west, inner sep=0pt] at (#1,#2) {\includegraphics[width=\ptw,height=\pth]{fig/perturb/#3}};\end{scope}
  \node[anchor=south east, fill=white, fill opacity=0.85, text opacity=1, inner sep=1.2pt, rounded corners=1pt, font=\scriptsize] at ({#1+\ptw-1.5pt},{#2-\pth+1.5pt}) {#4};}
\begin{tikzpicture}[x=1pt,y=1pt]
\pgfmathsetlengthmacro{\xa}{\ptl+2pt}
\pgfmathsetlengthmacro{\xb}{\ptl+2pt+\ptw+2pt}
\pgfmathsetlengthmacro{\xc}{\ptl+2pt+2*\ptw+4pt}
\pgfmathsetlengthmacro{\ya}{0pt}
\pgfmathsetlengthmacro{\yb}{-\pth-2pt}
\pgfmathsetlengthmacro{\yc}{-2*\pth-4pt}
\foreach \x/\t in {\xa/Proxy, \xb/SimPWM-H3-C, \xc/SimPWM-LTX-2.3-C} {\node[anchor=base, font=\footnotesize\bfseries] at ({\x+0.5*\ptw},3pt) {\t};}
\foreach \y/\t in {\ya/Recorded, \yb/No close, \yc/Frozen objects} {\node[rotate=90, anchor=south, font=\scriptsize\bfseries] at (\ptl,{\y-0.5*\pth}) {\t};}
\pimg{\xa}{\ya}{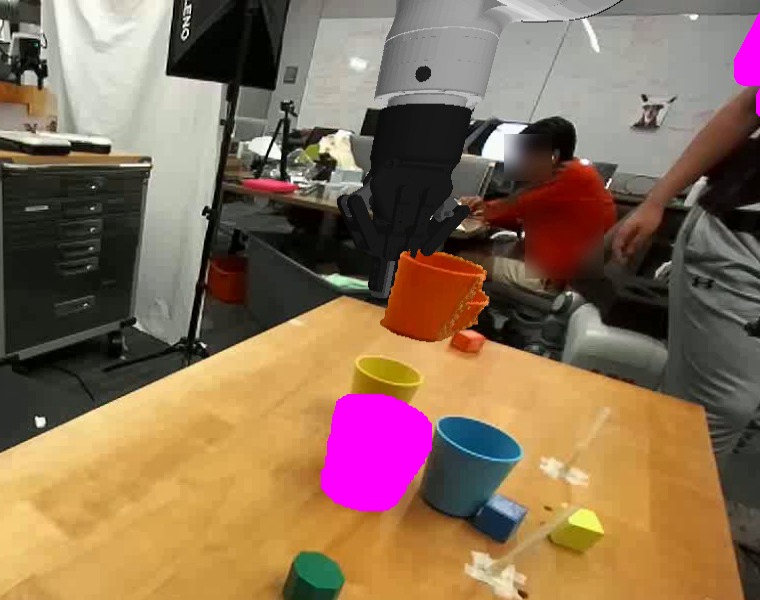}{cup carried}
\pimg{\xb}{\ya}{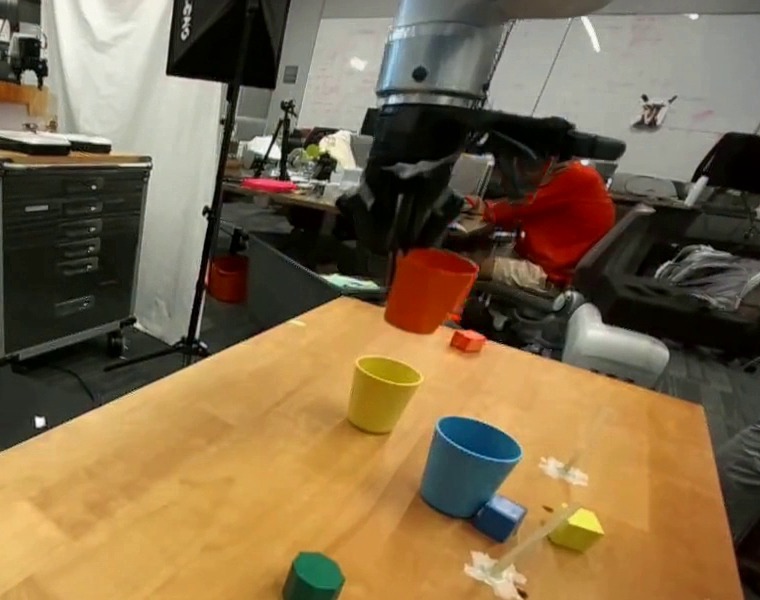}{\textcolor{tzok}{\cmark} lifted}
\pimg{\xc}{\ya}{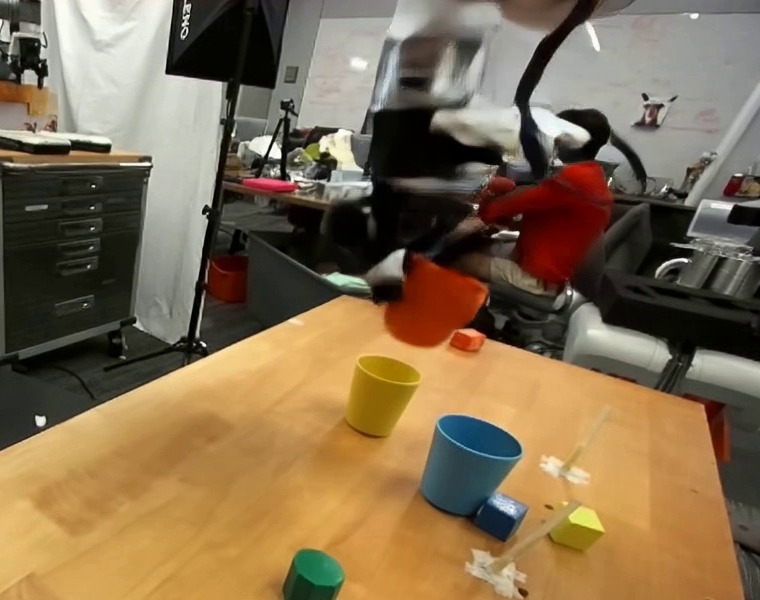}{\textcolor{tzok}{\cmark} lifted}
\pimg{\xa}{\yb}{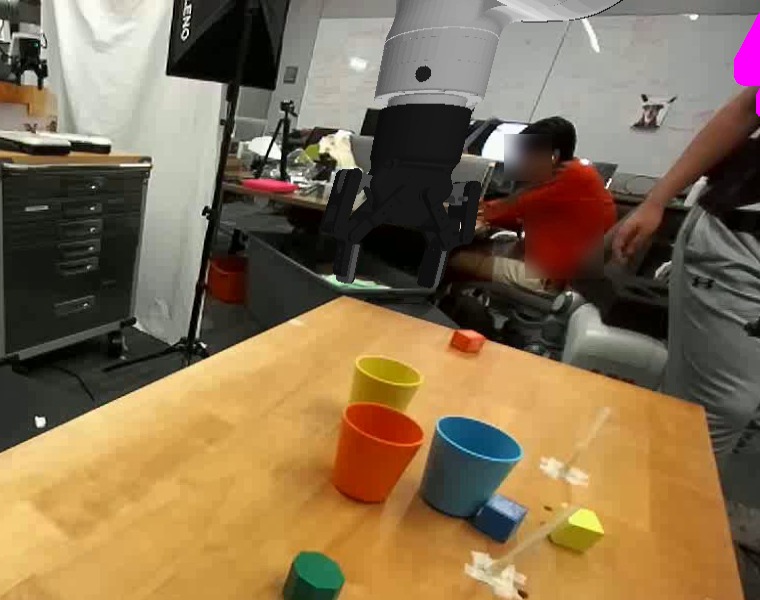}{open, cup still}
\pimg{\xb}{\yb}{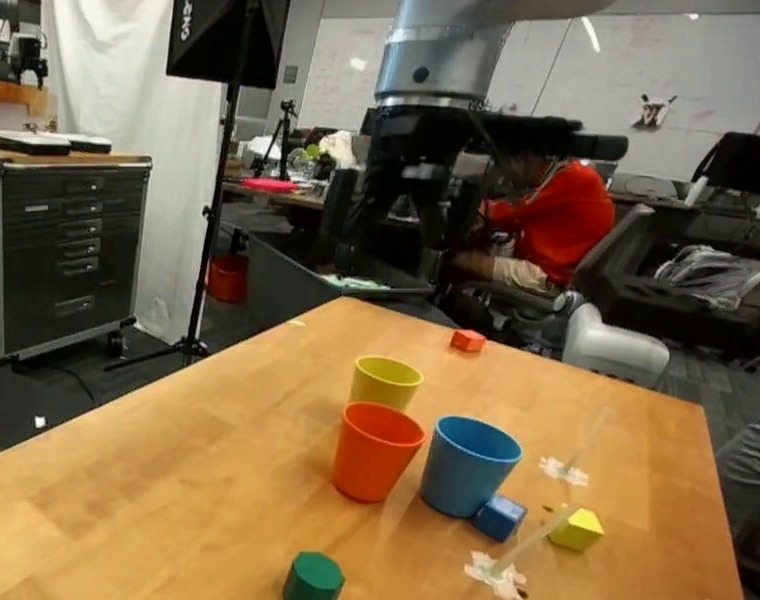}{\textcolor{tzok}{\cmark} stays}
\pimg{\xc}{\yb}{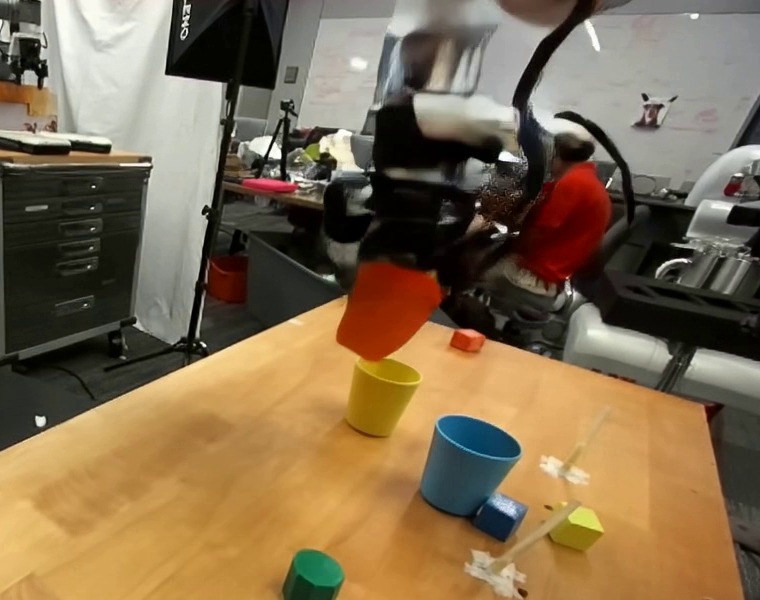}{\textcolor{tzbad}{\xmark} lifted}
\pimg{\xa}{\yc}{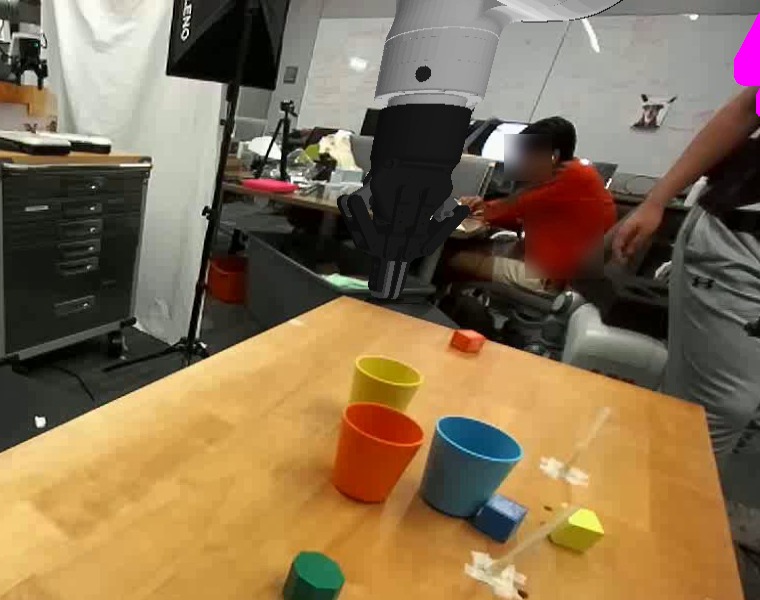}{closed, cup still}
\pimg{\xb}{\yc}{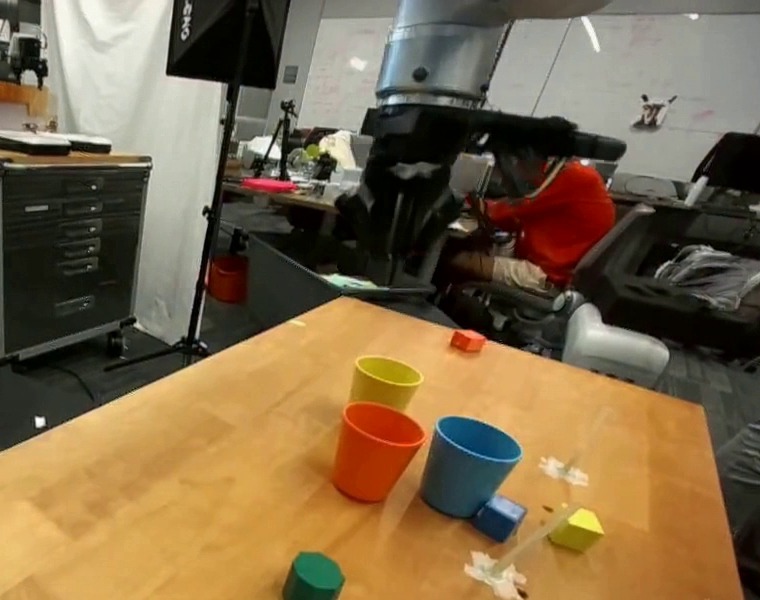}{\textcolor{tzok}{\cmark} stays}
\pimg{\xc}{\yc}{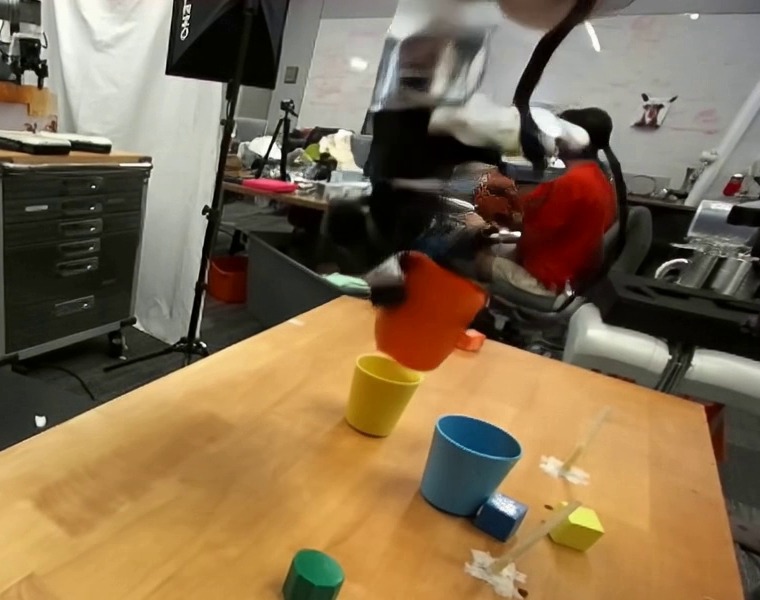}{\textcolor{tzbad}{\xmark} lifted}
\end{tikzpicture}
\caption{\textbf{Following the simulated state.} Under the recorded actions (top), both base models lift the cup as the proxy does. With the gripper held open (middle) or the simulated objects frozen while the arm moves as recorded (bottom), the proxy leaves the cup in place; SimPWM-H3-C does too, whereas SimPWM-LTX-2.3-C still lifts it.}
\label{fig:perturb}
\end{figure}

\pairnext
\begin{figure}[t]
\centering
\newlength{\tzw}\newlength{\tzh}\newlength{\tzg}
\setlength{\tzg}{3pt}
\setlength{\tzw}{\dimexpr(\columnwidth-2\tzg-4pt)/3\relax}
\setlength{\tzh}{0.86\tzw}%
\definecolor{tzbad}{RGB}{192,57,43}
\definecolor{tzok}{RGB}{30,132,73}
\definecolor{tzbox}{RGB}{232,238,252}
\newcommand{\tzimg}[3]{\begin{scope}\clip[rounded corners=2.2pt] (#1,#2) rectangle ++(\tzw,-\tzh);\node[anchor=north west, inner sep=0pt] at (#1,#2) {\includegraphics[width=\tzw,height=\tzh]{fig/teaser/#3}};\end{scope}}
\newcommand{\tzhead}[3]{\node[anchor=base, font=\footnotesize\bfseries, align=center] at ({#1+\hw},{#2+3.5pt}) {#3};}
\newcommand{\tznote}[4]{\node[anchor=north, text=#3, font=\scriptsize] at ({#1+\hw},{#2-\tzh-1.5pt}) {#4};}
\begin{tikzpicture}[x=1pt,y=1pt]
\useasboundingbox (0,14pt) rectangle (\columnwidth,{-2*\tzh-29pt-14pt});
\pgfmathsetlengthmacro{\xa}{2pt}
\pgfmathsetlengthmacro{\xb}{2pt+\tzw+\tzg}
\pgfmathsetlengthmacro{\xc}{2pt+2*\tzw+2*\tzg}
\pgfmathsetlengthmacro{\hw}{0.5*\tzw}
\pgfmathsetlengthmacro{\yr}{-\tzh-29pt}
\fill[tzbox, rounded corners=2.5pt] ({\xb-2pt},{\yr+12pt}) rectangle ({\xc+\tzw+2pt},{\yr-\tzh-12pt});
\tzhead{\xa}{0}{Input}
\tzhead{\xb}{0}{Video world model}
\tzhead{\xc}{0}{Agentic Real2Sim}
\tzimg{\xa}{0}{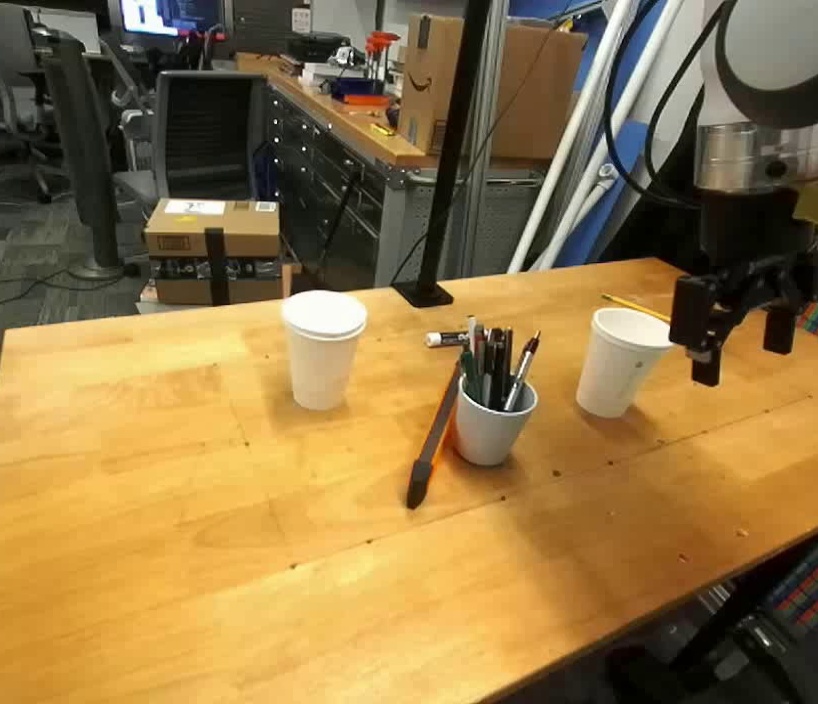}
\tzimg{\xb}{0}{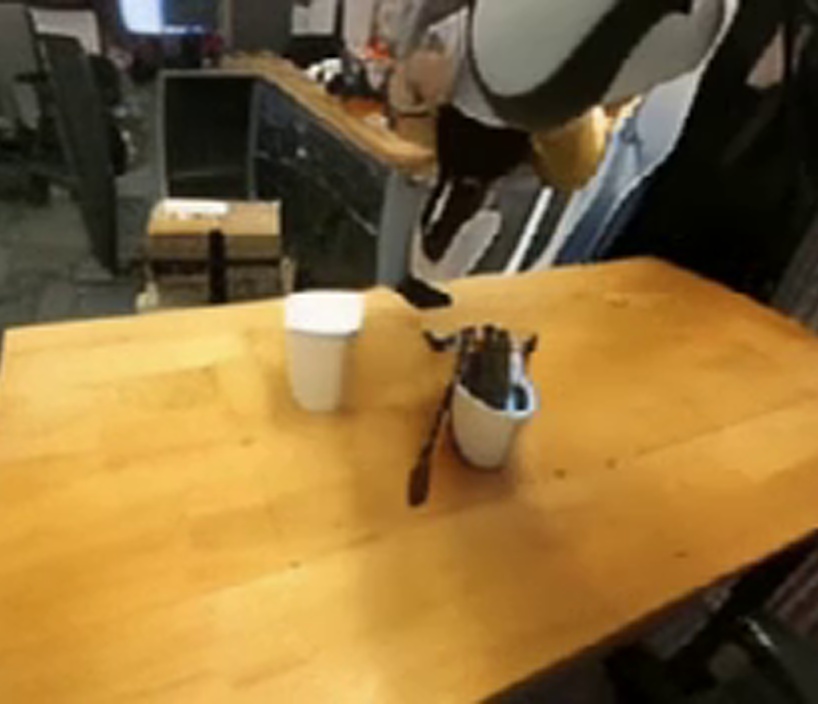}
\tzimg{\xc}{0}{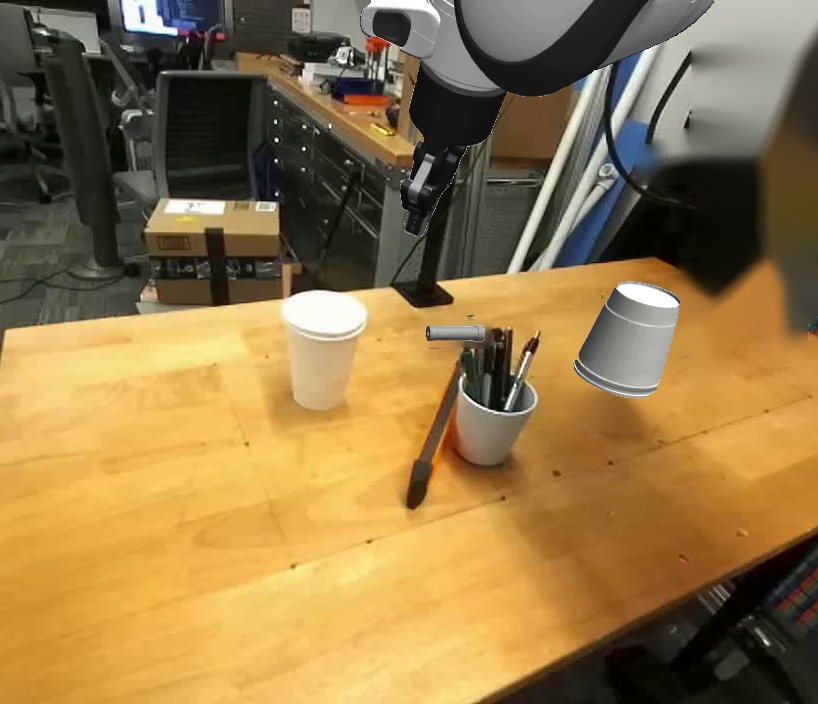}
\tznote{\xa}{0}{black!60}{$+$ actions $a_{1:T}$}
\tznote{\xb}{0}{tzbad}{\xmark\ cup vanishes}
\tznote{\xc}{0}{tzbad}{\xmark\ cup knocked over}
\tzhead{\xa}{\yr}{Ground truth}
\tzhead{\xb}{\yr}{Ours: simulator}
\tzhead{\xc}{\yr}{Ours: video model}
\tzimg{\xa}{\yr}{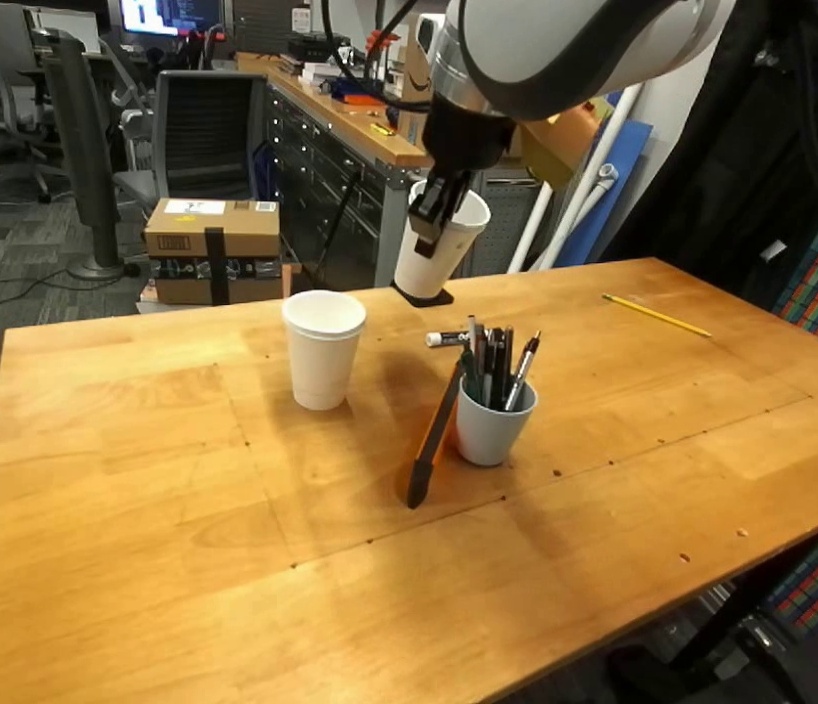}
\tzimg{\xb}{\yr}{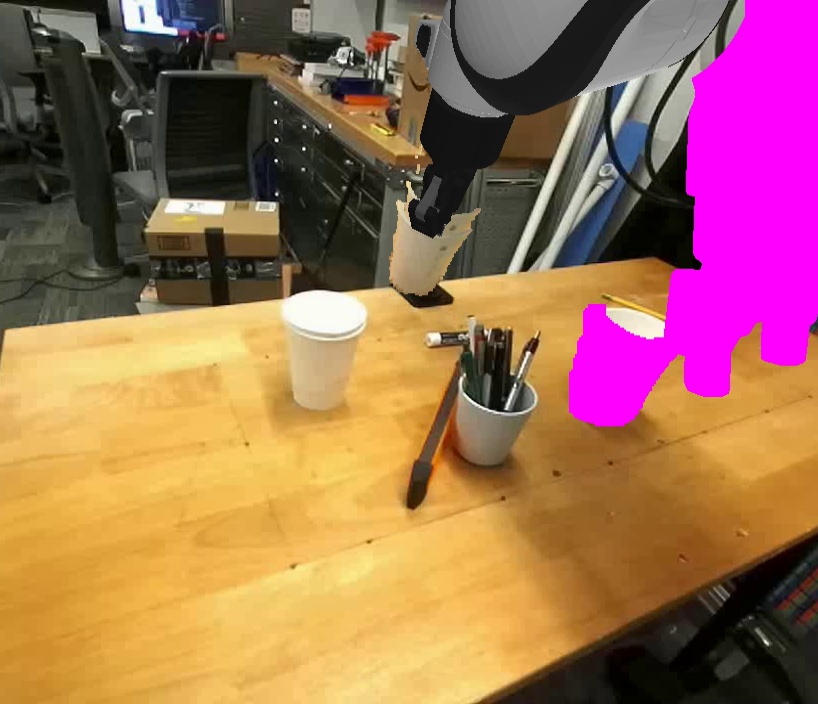}
\tzimg{\xc}{\yr}{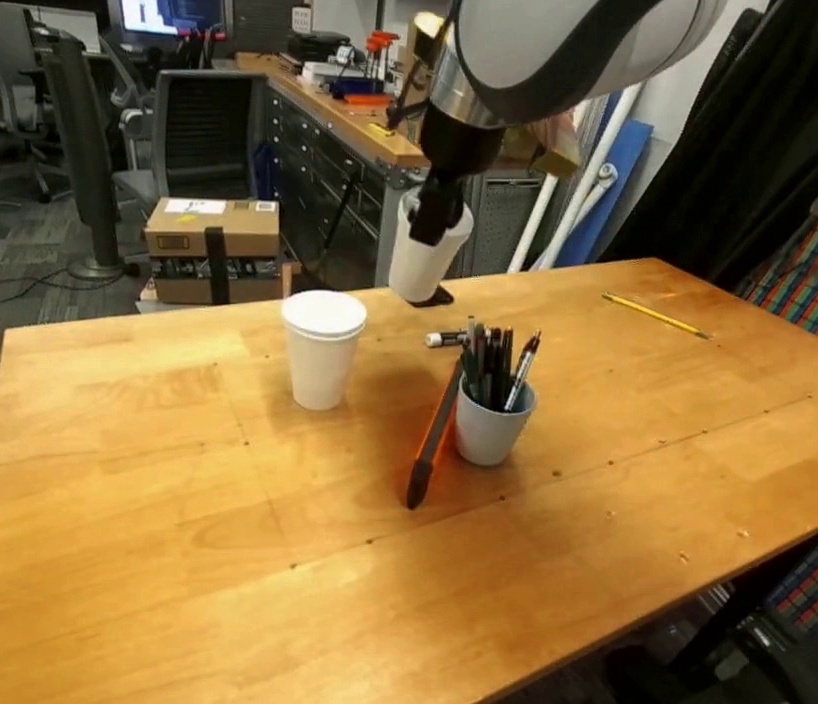}
\tznote{\xb}{\yr}{tzok}{\cmark\ explicit: cup carried}
\tznote{\xc}{\yr}{tzok}{\cmark\ realistic: cup carried}
\end{tikzpicture}
\caption{\textbf{SimPWM outputs: realistic video and explicit simulated state.} SimPWM-H3-C carries the cup as in the real video; Ctrl-World loses it and Agentic Real2Sim (render composited on the input background) knocks it over.}
\label{fig:teaser}
\end{figure}
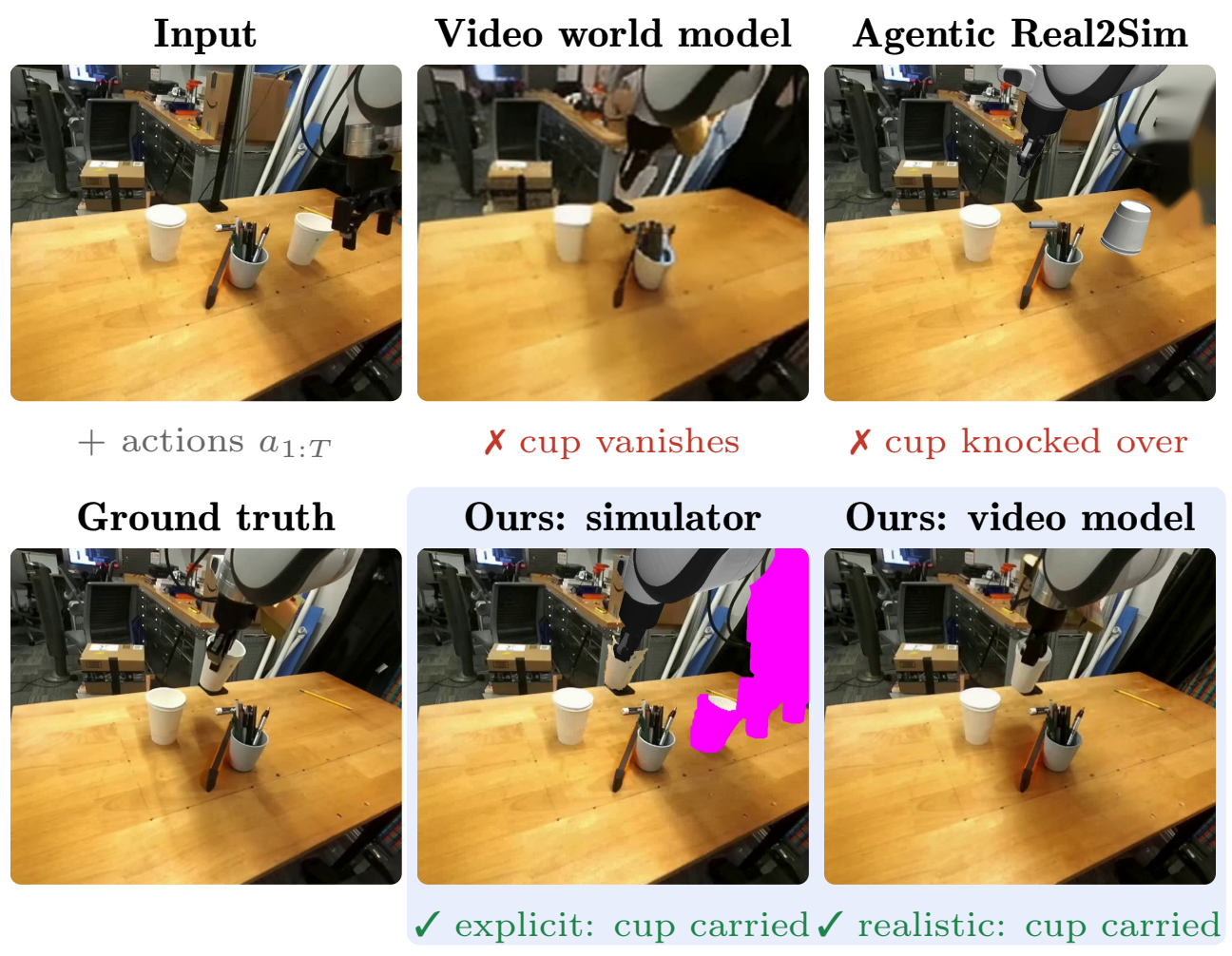

\end{floatpair}

Action-conditioned video world models take an observation and a sequence of robot actions and render the future~\cite{yang2024unisim,agarwal2025cosmos,zhu2024irasim}.
They must follow the actions strictly: the arm must move exactly as commanded, an object may move only when the gripper grasps it, and it must stay where it is released.
To this end they are increasingly conditioned on a simulated or rendered proxy of the scene, such as the commanded robot rendered from its model~\cite{kim2026rofacto}, a 4D occupancy sequence~\cite{yang2026orv}, structural controls such as depth~\cite{nvidia2025cosmostransfer} or a physics simulation~\cite{liu2024physgen,li2025wonderplay}, on the assumption that the video will show what the proxy shows.
Testing this assumption requires a proxy whose state is explicit and can be perturbed.
\hyperref[fig:perturb]{Fig.~\ref*{fig:perturb}} does so: with the simulated cup frozen while the arm moves as recorded, one video model leaves the cup in place, as the proxy does, and another still lifts it.

Such a proxy also combines the strengths of the two current kinds of models, which fail in opposite ways (\hyperref[fig:teaser]{Fig.~\ref*{fig:teaser}}).
\emph{End-to-end video world models}~\cite{zhu2024irasim,guo2025ctrlworld,agarwal2025cosmos} focus mostly on visual quality: their frames look realistic, but they often do not follow the actions.
The arm drifts from the commanded trajectory, and grasped objects split, stay behind or dissolve.
Because the consequence is never represented, these errors are hard to detect or correct.
\emph{Simulators} compute the consequences from kinematics and contact, but they need a scene.
A scene rebuilt from a real image, by hand, automatically~\cite{ranawaka2026simfoundry} or by agents~\cite{chen2026agentic}, does not match it exactly, and a few centimetres decide whether a grasp succeeds; rigid-body physics also mishandles ropes and cloth, and the rendered video does not look real.

We combine the two into a model that returns two outputs: \emph{a realistic video from a video model, and an explicit state of what the actions cause, from a simulator, that the video can be tested against}.
\emph{What moves where} (the arm, the grasped object and where it goes) depends mainly on coarse geometry and the commanded motion, which a simple simulator computes explicitly.
\emph{How it looks}, from appearance and shadows to how a rope or a cloth hangs (\hyperref[fig:gallery]{Fig.~\ref*{fig:gallery}}), depends on fine physics and appearance, which video models learn well from data.
Accordingly, we credit arm following to the robot rendering, which prior work also uses~\cite{kim2026rofacto}, and evaluate the simulator only on what it adds: which object moves, and whether the video follows it.
Our model, \textbf{SimPWM} (\textbf{Sim}ulator-as-\textbf{P}roxy \textbf{W}orld \textbf{M}odel), is built on this split (\hyperref[fig:overview]{Fig.~\ref*{fig:overview}}).
From the first frame it separates the scene into three parts, the robot as its 3D model, the objects the gripper may touch as 3D points, and the background (\emph{real-to-sim}); a simple simulator moves the robot and the objects under the actions with explicit rules for grasp, carry, release and gravity (\hyperref[fig:simulator]{Fig.~\ref*{fig:simulator}}); and the simulated scene is rendered per frame as a \emph{proxy}: the first frame, with the robot and every moved object redrawn where the simulator puts them.
A video model that takes a reference video, fine-tuned to render such proxies, turns them into a realistic video (\emph{sim-to-real}); we instantiate it with three base models, MiniMax-H3, LTX-2.3 and Cosmos-Transfer~2.5.
The proxy pins down the arm whatever the base model, more closely than end-to-end models, but the rendered robot alone does this; the simulated objects add the pick-ups that the robot alone misses.

\begin{sidecapfloat}[t]{0.6}
\begin{figure}[t]
\centering
\setlength{\tabcolsep}{1.2pt}
\newlength{\galw}\setlength{\galw}{\dimexpr(\columnwidth-8\tabcolsep)/4\relax}%
\newcommand{\galimg}[1]{\rimg[width=\galw]{fig/gallery/#1}}
\newcommand{\galrow}[1]{\galimg{#1_gt.jpg} & \galimg{#1_rgb.jpg} & \galimg{#1_depth.jpg} & \galimg{#1_gen.jpg}}
\renewcommand{\arraystretch}{0}
\begin{tabular}{@{}cccc@{}}
\figlab{GT} & \figlab{RGB proxy} & \figlab{Depth proxy} & \figours{Generated} \\[3pt]
\galrow{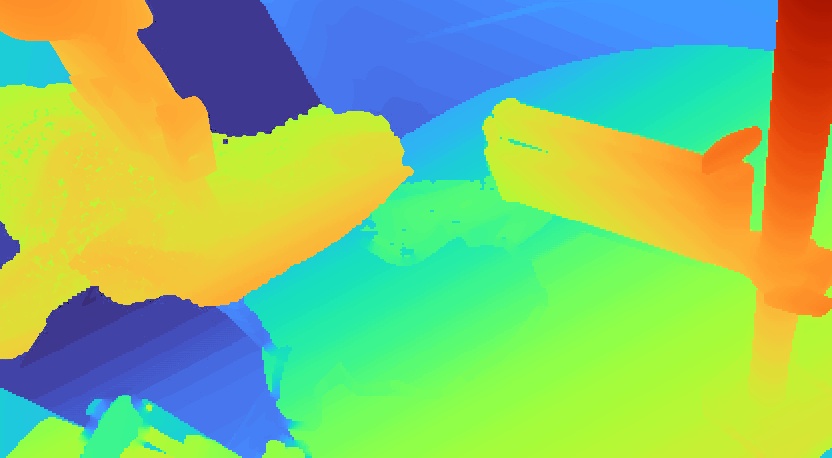} \\[2.5pt]
\multicolumn{4}{c}{\figdesc{Put the black t-shirt on the black stand}} \\[3pt]
\galrow{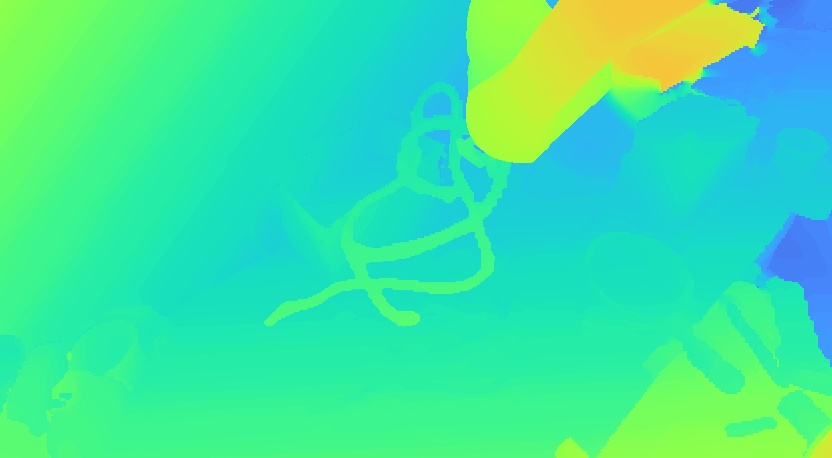} \\[2.5pt]
\multicolumn{4}{c}{\figdesc{Pick up the white ropes and put them in the box}} \\[3pt]
\galrow{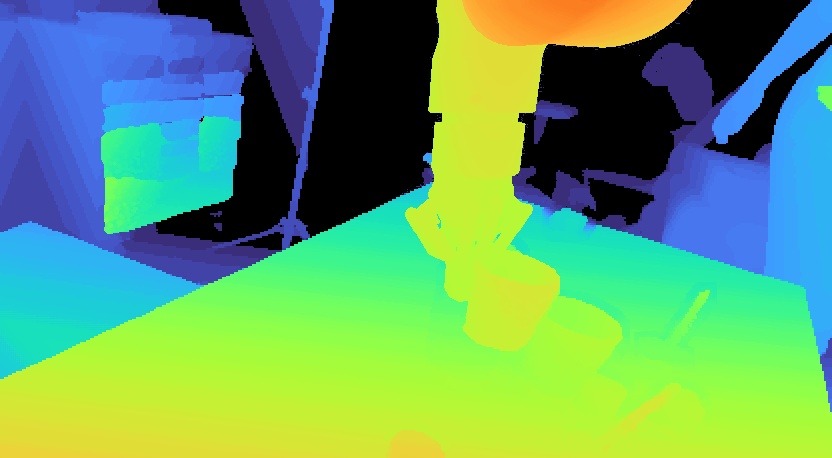} \\[2.5pt]
\multicolumn{4}{c}{\figdesc{Pick up the orange cup and place it into the yellow cup}} \\
\end{tabular}
\caption{\textbf{Results of SimPWM} on test episodes from unseen setups (generated: SimPWM-H3-C). The simulator carries cloth and rope rigidly; the video model lets them hang (magenta: unseen).}
\label{fig:gallery}
\end{figure}

\end{sidecapfloat}
Unlike the arm, the object state is followed only by some base models (\hyperref[sec:evaluator]{Sec.~\ref*{sec:evaluator}}): with the simulated objects frozen, MiniMax-H3 mostly leaves the object in place, even without fine-tuning, whereas LTX-2.3 lifts it as if inferring the grasp from the arm motion.
With any base model, the simulated state is returned as an explicit record of what the simulation did; under perturbed actions, a video that disagrees with it reveals that the video ignores the simulated state. What the simulator adds is that whether the video follows the objects can be tested by perturbing the simulation, a test that end-to-end models, having no simulated state, do not admit.

Three choices make this work without training the simulator: (i) objects are point sets lifted from the first frame, without categories or CAD models, so the scene matches the image by construction; (ii) the actions correct perception, since an object the gripper closes on must lie at the gripper's depth; and (iii) robot-specific knowledge enters only through a configuration file, so the same pipeline runs on single-arm and bimanual robots.

\paragraph{Contributions.}
(i) A real-to-sim-to-real world model that returns a realistic video together with an explicit state of what the actions cause, simulated at test time from the first frame; under perturbed actions the state reveals videos that ignore it.
(ii) A class-agnostic, image-aligned point-set scene whose depth is corrected by the commanded gripper closure, so that carried objects are placed more accurately.
(iii) Perturbation and frozen-object tests showing that every base model follows the rendered arm, whereas the simulated object state is followed only by a base model whose prior favours the reference video (MiniMax-H3); against a robot-only proxy, the simulated objects reduce missed pick-ups by $16.5$ points ($[8.3, 25.6]$) and account for $18.8$ points ($[10, 28]$) of the drop under frozen objects.
(iv) An evaluation on $293$ DROID episodes from unseen setups against end-to-end video world models and a learned 3D world model.
SimPWM also runs on WorldArena and is compared with an agentic real-to-sim pipeline in the appendix.

\section{Related Work}
\label{sec:related}

\paragraph{Robot world models.}
Action-conditioned video generators serve as learned simulators for games and robots~\cite{finn2016unsupervised,ebert2018visualforesight,ha2018worldmodels,hafner2023dreamerv3,wu2022daydreamer,bruce2024genie,valevski2025gamengen,yang2024unisim,du2023unipi,ko2024avdc,wu2024ivideogpt,zhu2024irasim,agarwal2025cosmos,liao2026memorizon}.
For manipulation, Ctrl-World and WorldGym roll out and rank policies~\cite{guo2025ctrlworld,quevedo2026worldgym}, RoboScape adds depth and keypoint objectives~\cite{shang2025roboscape}, EnerVerse-AC projects end-effector actions into the image~\cite{jiang2025enerverseac}, and DreamDojo pretrains on human video with latent actions~\cite{gao2026dreamdojo}.
These models are trained end to end, so the effect of an action is implicit in the weights.
Other models learn how the 3D scene moves, as particles, scene flow, point flows or voxel latents~\cite{whitney2024hdvpd,guo2025flowdreamer,huang2026pointworld,wang2026worldsim3d}; SimPWM computes it with rules that need no training (we compare with PointWorld in \hyperref[sec:exp_droid]{Secs.~\ref*{sec:exp_droid}} and~\ref{sec:exp_render}).

\paragraph{Physics and proxies for video generation.}
\begin{table}[t]
\centering\footnotesize\setlength{\tabcolsep}{5pt}
\renewcommand{\arraystretch}{1.08}
\caption{\textbf{Simulation-driven video generation.} What each method simulates, what conditions the video, and whether it tests that the video follows the simulated state (sim.: simulation; R2S: Real2Sim).}
\label{tab:related}
\tabstyle
\begin{tabular}{l||l|l|l|c}
\toprule
\textbf{Method} & \textbf{Input} & \textbf{Simulated} & \textbf{Condition} & \textbf{Tests} \\
\tabdouble
\tabgroup{5}{Simulation drives a video model} \\
PhysGen~\cite{liu2024physgen} & image, forces & 2D rigid bodies & simulated motion & -- \\
\rowcolor{tabzebra} WonderPlay~\cite{li2025wonderplay} & image, 3D actions & objects, materials & coarse sim.\ video & -- \\
ORV~\cite{yang2026orv} & 4D occupancy & -- (given) & occupancy & -- \\
\rowcolor{tabzebra} RoFacto~\cite{kim2026rofacto} & actions & robot only & render, depth & -- \\
\tabdouble
\tabgroup{5}{Scene dynamics without a video model} \\
PointWorld~\cite{huang2026pointworld} & RGB-D, actions & points (learned) & -- (3D only) & -- \\
\rowcolor{tabzebra} Agentic R2S~\cite{chen2026agentic} & recorded video & MuJoCo (agents) & MuJoCo render & -- \\
\tabdouble
\rowcolor{tabours} SimPWM (ours) & image, actions & robot, grasped objects (rules) & RGB render of both & \checkmark \\
\bottomrule
\end{tabular}
\end{table}

Closest to our approach (\hyperref[tab:related]{Table~\ref*{tab:related}}) are methods that drive a video generator with a physics simulation of a single image~\cite{liu2024physgen,li2025wonderplay}; SimPWM shares this split but simulates a robot: its inputs are joint and gripper commands, it decides grasps rather than applying forces, and it needs no material estimation.
Video can also be rendered from structural conditions such as label maps, depth, edges or masks~\cite{wang2018vid2vid,mallya2020worldconsistent,zhang2023controlnet,esser2023gen1,wang2023videocomposer,guo2024sparsectrl,nvidia2025cosmostransfer,li2025manipdreamer} or 3D tracking videos~\cite{gu2025das}, for robots from 4D occupancy~\cite{yang2026orv} or a render of the commanded robot~\cite{kim2026rofacto}.
These conditions are given, extracted from a reference video, or cover only the robot; in SimPWM they are simulated at test time from the first frame and the actions, so they encode what the actions do to objects. Our robot-only proxy is the closest analogue of RoFacto (\hyperref[sec:exp_render]{Secs.~\ref*{sec:exp_render}} and~\ref{sec:evaluator}).

\paragraph{Real-to-sim.}
Evaluating policies in simulation requires matching the simulated scene to the real one~\cite{li2024simpler}.
Agentic Real2Sim turns a DROID recording into a MuJoCo twin with vision-language agents~\cite{chen2026agentic}; others build scenes or objects from videos or scans~\cite{torne2024rialto,ranawaka2026simfoundry,jiang2025phystwin,yu2025r2r2r}.
These need recorded videos or scans, and their scenes need not match the image; SimPWM lifts an image-aligned point set from the first frame, poses the robot with MuJoCo~\cite{todorov2012mujoco}, and leaves appearance to the video model.

\section{Method}
\label{sec:method}
\begin{figure*}[t]
\centering
\includegraphics[width=\linewidth]{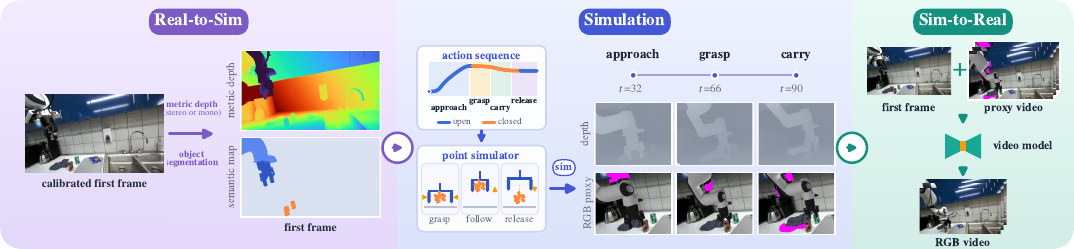}
\caption{\textbf{Overview of SimPWM.} The first frame is lifted into a point set (real-to-sim), moved under the actions by a rule-based simulator (blue: gripper open, orange: closed), and its proxy is rendered into video (sim-to-real). Steps $t$ within the $97$ evaluated steps; video model: SimPWM-H3-C.}
\label{fig:overview}
\end{figure*}

SimPWM takes the first frame $x_0$ of a calibrated camera and the action sequence $a_{1:T}$ (arm joint angles and gripper closure) and generates the video in the three stages of \hyperref[fig:overview]{Fig.~\ref*{fig:overview}} (\hyperref[sec:method_init]{Secs.~\ref*{sec:method_init}}--\ref{sec:method_s2r}).
Geometry is expressed in the robot base frame, with gravity along $-z$.
The camera parameters are inputs; scene perception uses only the first frame.
On DROID we fit the camera per episode to the robot's appearance, aligning the rendered robot mask with the segmented one on $20$ frames, $5$ inside the evaluated window.
The accuracy of the rendered arm therefore rests on this calibration; fitting only on frames outside the window lowers the robot IoU of the proxy from $0.787$ to $0.768$, and its grasp success, not significantly, from $69.5\%$ to $68.0\%$.

\paragraph{Scene state.}
The robot is its mesh model, posed by forward kinematics.
Everything else is points: each object is a point set $\{p_i\}_{i=1}^{N}\subset\mathbb{R}^3$, \ie its visible surface plus a back side completed along camera rays, and the static background is the first-frame depth map together with a support plane.

\subsection{Real-to-Sim}
\label{sec:method_init}
\paragraph{Metric depth.}
On DROID the external camera is a stereo camera: we estimate disparity on the rectified first-frame pair with FoundationStereo~\cite{wen2025foundationstereo} and convert it to metric depth with the focal length and baseline.
Pixels without stereo support, such as the border that only one camera sees and occlusions, take monocular metric depth from Depth Anything~V2~\cite{yang2024depthv2}, scaled to the stereo depth by the median ratio on the valid non-robot pixels.
Without a stereo pair (\hyperref[sec:exp_droid]{Sec.~\ref*{sec:exp_droid}}), the monocular depth is used alone, scaled with the robot as a ruler.

\paragraph{Support and objects.}
The support is a gravity-aligned plane fitted by RANSAC to the non-robot points; among strong horizontal candidates we take the one closest to the lowest point that the effector reaches during the actions.
Objects are what rises above the support.
We segment them with SAM~2~\cite{ravi2024sam2} and SAM~3~\cite{carion2025sam3} proposals, split large holes in the support with SAM~2 point prompts, and add a segment at every point where the gripper closes.

\paragraph{Action-conditioned depth refinement (action snap).}
\label{sec:method_refine}
For each window in which the gripper closes, we project the lowest effector pose into the first frame; if that pixel lies on an object, the object must be at the effector's depth there.
Scaling the object's points about the camera centre by a factor $s$ leaves its image footprint unchanged, so we choose $s$ to match the two depths (within $s\in[0.35,2.5]$).

\subsection{Rule-Based Point Simulator}
\label{sec:method_engine}
The simulator moves the scene from frame $t$ to $t+1$ given the robot's next state $a_{t+1}$.
It models only what decides pick-and-place and uses only the object point sets and the effector frame $(o,R)$ and keypoints (finger pads and tool-centre point) given by forward kinematics; its thresholds are set on development episodes whose setups are excluded from the test set.

\textbf{Grasp.} When the gripper closes and holds nothing, it grasps the object that fits between the open jaws and has the most points, at least $k$, between its pads; an object wider than the jaws (\eg a cloth) is pinched at its edge.

\textbf{Carry.} The grasped points are stored in the effector frame at the grasp step $\tau$, $Q=(P_\tau-o_\tau)R_\tau$, and follow it rigidly, $P_t=QR_t^{\top}+o_t$.

\begin{wrapside}{0.4}
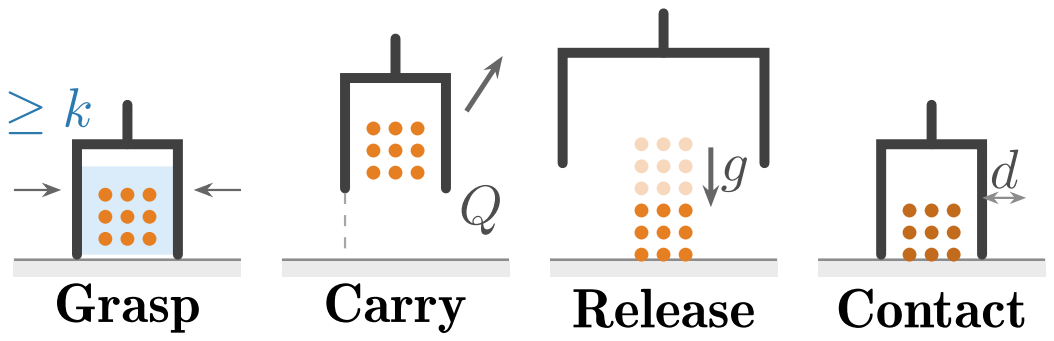
\begin{figure}[t]
\centering
\newcommand{\simgripper}[2]{%
  \draw[line width=1.8pt, black!75, line cap=round] (-#1,#2) -- (-#1,{#2+0.7}) -- (#1,{#2+0.7}) -- (#1,#2);
  \draw[line width=1.8pt, black!75, line cap=round] (0,{#2+0.7}) -- (0,{#2+0.95});}
\newcommand{\simobj}[2]{%
  \foreach \px in {-0.14,0,0.14}{\foreach \py in {0,0.14,0.28}{\fill[#1] ({\px},{#2+\py}) circle (1.25pt);}}}
\newcommand{\simtable}{\draw[black!45, line width=0.9pt] (-0.72,0) -- (0.72,0); \fill[black!8] (-0.72,0) rectangle (0.72,-0.1);}
\resizebox{\linewidth}{!}{%
\begin{tikzpicture}[x=1cm,y=1cm,font=\small]
\definecolor{objc}{RGB}{230,126,34}
\definecolor{regc}{RGB}{52,152,219}
\begin{scope}
  \simtable
  \fill[regc!18, rounded corners=1pt] (-0.32,0.04) rectangle (0.32,0.6);
  \simobj{objc}{0.14}
  \simgripper{0.32}{0.04}
  \draw[-{Stealth[length=3.5pt]}, black!60] (-0.72,0.45) -- (-0.42,0.45);
  \draw[-{Stealth[length=3.5pt]}, black!60] (0.72,0.45) -- (0.42,0.45);
  \node[text=regc!80!black] at (-0.5,0.95) {$\geq k$};
  \node at (0,-0.3) {\textbf{Grasp}};
\end{scope}
\begin{scope}[xshift=1.7cm]
  \simtable
  \begin{scope}[yshift=0.42cm]
    \simobj{objc}{0.14}
    \simgripper{0.32}{0.04}
  \end{scope}
  \draw[black!35, dash pattern=on 2pt off 2pt] (-0.32,0.08) -- (-0.32,0.44);
  \draw[-{Stealth[length=4pt]}, line width=0.9pt, black!60] (0.45,0.95) -- (0.68,1.3);
  \node[anchor=west, text=black!70, inner sep=1pt] at (0.36,0.32) {$Q$};
  \node at (0,-0.3) {\textbf{Carry}};
\end{scope}
\begin{scope}[xshift=3.4cm]
  \simtable
  \simgripper{0.64}{0.62}
  \begin{scope}[opacity=0.3]\simobj{objc}{0.46}\end{scope}
  \simobj{objc}{0.04}
  \draw[-{Stealth[length=4pt]}, line width=0.9pt, black!60] (0.3,0.72) -- (0.3,0.34);
  \node[anchor=west, text=black!70, inner sep=1pt] at (0.32,0.55) {$g$};
  \node at (0,-0.3) {\textbf{Release}};
\end{scope}
\begin{scope}[xshift=5.1cm]
  \simtable
  \simobj{objc!85!black}{0.04}
  \simgripper{0.32}{0.04}
  \draw[black!45, <->, >={Stealth[length=3pt]}] (0.32,0.4) -- (0.6,0.4);
  \node[anchor=south, text=black!70, inner sep=1pt] at (0.46,0.42) {$d$};
  \node at (0,-0.3) {\textbf{Contact}};
\end{scope}
\end{tikzpicture}}
\caption{\textbf{Rule-based point simulator:} grasp, rigid carry, release with gravity, and the pad distance $d$ (diagnostic only).}
\label{fig:simulator}
\end{figure}

\end{wrapside}
\textbf{Release and gravity.} When the gripper opens, the object falls onto a height map of the scene and stops at the first contact; an object placed by the depth refinement stays put until it is first grasped.

A bimanual robot runs one copy per arm.

\subsection{Sim-to-Real}
\label{sec:method_s2r}
\paragraph{Proxy rendering.}
\label{sec:method_render}
Each frame of the simulated scene is rendered into the proxy: the robot from its posed mesh, the objects as point splats and the background from the first-frame depth.
The video model receives the \emph{RGB proxy}, in which every point keeps its first-frame colour and the robot is a shaded mesh; where the robot or a held object uncovers part of its first-frame footprint, the first frame shows nothing, and we fill these pixels with a marker colour (magenta).
The same rendering gives a depth map (the \emph{depth proxy}, compared in \hyperref[sec:exp_render]{Sec.~\ref*{sec:exp_render}}) and a semantic map, in which only the robot and the objects that the gripper held get their own label; the latter is not an input to the video model, but gives the \emph{readout}: whether, and which object, the gripper holds at each step (\hyperref[sec:evaluator]{Sec.~\ref*{sec:evaluator}}).

\paragraph{Video model.}
The proxy fixes where the robot is and where the simulator moves each object in every frame.
The video model supplies the rest: appearance, lighting and shadows, the background hidden in the first frame, and the physical detail that the simulator lacks, such as how a rope or a cloth hangs.
Any video model that takes a reference video can render the proxy; we fine-tune three, each receiving the proxy as the reference video, the real first frame $x_0$ and a text prompt naming the task.
\emph{SimPWM-H3} is a LoRA (rank $32$) on MiniMax-H3~\cite{minimax2026h3} ($33$B parameters), whose Ref2VA variant also takes $x_0$ as a reference image and as the first latent frame of the target; it generates $158$ frames at $24$\,fps, to which the $97$ recorded steps are resampled (so a clip lasts $6.5$\,s instead of $6.4$\,s).
\emph{SimPWM-LTX-2.3} is an in-context LoRA (IC-LoRA)~\cite{huang2024iclora} on LTX-2.3 ($22$B), an update of LTX-2~\cite{hacohen2026ltx2}, generating $97$ frames at $15$\,fps.
\emph{SimPWM-Cosmos-Transfer} is a control branch on Cosmos-Transfer~2.5~\cite{nvidia2025cosmospredict25} ($2.4$B), trained like SimPWM-LTX-2.3.
All three are trained on the $3{,}000$ training episodes of \hyperref[sec:exp_setup]{Sec.~\ref*{sec:exp_setup}}.
Our main model, \emph{SimPWM-H3-C}, is trained in the same way on the \emph{consistent subset}: the $845$ training episodes whose proxy agrees with the real video; \emph{SimPWM-LTX-2.3-C} is LTX-2.3 trained on it for comparison. \emph{H3 zero-shot} is MiniMax-H3 with our proxy and no fine-tuning; the \emph{ablation models} are SimPWM-H3-C at $640{\times}352$, one per proxy variant. \hyperref[tab:variants]{Table~\ref*{tab:variants}} lists all variants.

\section{Experiments}
\label{sec:exp}
\begin{figure*}[t]
\centering
\footnotesize
\setlength{\tabcolsep}{3.6pt}
\renewcommand{\arraystretch}{1.08}
\captionof{table}{\textbf{Comparison with baselines.} Same first frame and actions for all methods; metrics against the real video at $320{\times}192$ and $5$\,Hz. Size: parameters of the video model. $^\dagger$Trained on DROID including at least $234$ of our $293$ test episodes. $^\ddagger$MiniMax-H3 with our proxy, without fine-tuning. Cosmos-Predict~2.5 and SimPWM-Cosmos-Transfer share a base model. \mbox{-C}: trained on the consistent subset; the other fine-tuned SimPWM rows on all $3{,}000$ training episodes (\hyperref[tab:variants]{Table~\ref*{tab:variants}}). Best in bold.}
\label{tab:droid_compare}
\tabstyle
\begin{tabular}{l||c|ccc|ccc|c}
\toprule
\multirow{2}{*}{\textbf{Method}} & \multirow{2}{*}{\textbf{Size}} & \multicolumn{3}{c|}{\textbf{Video quality}} & \multicolumn{3}{c|}{\textbf{Motion}} & \textbf{Physics} \\
\cmidrule(lr){3-5}\cmidrule(lr){6-8}\tabcmidend
 & & PSNR$\uparrow$ & LPIPS$\downarrow$ & FVD$\downarrow$ & Robot$\uparrow$ & Traj.$\uparrow$ & Depth$\downarrow$ & Phys-IQ$\uparrow$ \\
\tabdouble
\tabgroup{9}{Video world models conditioned on the actions} \\
Ctrl-World$^\dagger$~\cite{guo2025ctrlworld} & 1.5B & 23.23 & .094 & 105.9 & .690 & .613 & .182 & \textbf{77.8} \\
\rowcolor{tabzebra} DreamDojo~\cite{gao2026dreamdojo} & 2B & 19.74 & .142 & 155.5 & .304 & .271 & .320 & 67.8 \\
Cosmos-Predict~2.5~\cite{nvidia2025cosmospredict25} & 2.1B & 20.65 & .128 & 134.3 & .398 & .384 & .272 & 66.8 \\
\tabdouble
\tabgroup{9}{SimPWM: video models conditioned on the simulator proxy} \\
\rowcolor{tabours} SimPWM-Cosmos-Transfer & 2.4B & 23.73 & .074 & 86.1 & .754 & .750 & .155 & 75.8 \\
\rowcolor{tabours} SimPWM-LTX-2.3 & 22B & 23.66 & .073 & 91.7 & \textbf{.777} & .764 & \textbf{.148} & 69.9 \\
\rowcolor{tabours} SimPWM-LTX-2.3-C & 22B & 23.36 & .076 & 89.4 & .773 & \textbf{.773} & .151 & 69.9 \\
H3 zero-shot$^\ddagger$ & 33B & 21.65 & .105 & 194.1 & .718 & .686 & .184 & 66.4 \\
\rowcolor{tabours} SimPWM-H3 & 33B & 23.81 & \textbf{.069} & 86.1 & .768 & .765 & .155 & 76.5 \\
\rowcolor{tabours} \textbf{SimPWM-H3-C} & 33B & \textbf{23.84} & \textbf{.069} & \textbf{85.6} & \textbf{.777} & .770 & .153 & 76.0 \\
\bottomrule
\end{tabular}
\par\vspace{10pt}
{\centering
\setlength{\tabcolsep}{1.2pt}
\newlength{\qw}\setlength{\qw}{\dimexpr(\textwidth-12\tabcolsep-11pt)/7\relax}
\newcommand{\qa}[1]{\rimg[width=\qw]{fig/qualitative/#1_crop.jpg}}
\newcommand{\qrow}[1]{\qa{#1_gt} & \qa{#1_cw} & \qa{#1_dd} & \qa{#1_cp} & \qa{#1_wct} & \qa{#1_wpm} & \qa{#1_h3}}
\renewcommand{\figlab}[1]{\makebox[0pt]{\parbox[b][15pt][c]{\qw}{\centering\scriptsize\renewcommand{\arraystretch}{0.9}\begin{tabular}{@{}c@{}}#1\end{tabular}}}}%
\renewcommand{\figours}[1]{\makebox[0pt]{\parbox[b][15pt][c]{\qw}{\centering{\setlength{\fboxsep}{1.2pt}\colorbox{tabours}{\scriptsize\renewcommand{\arraystretch}{0.9}\begin{tabular}{@{}c@{}}#1\end{tabular}}}}}}%
\renewcommand{\arraystretch}{0}
\begin{tabular}{@{}ccccccc@{}}
\figlab{GT} & \figlab{Ctrl-World} & \figlab{DreamDojo} & \figlab{Cosmos-Predict\\2.5} & \figours{SimPWM-\\Cosmos-Transfer} & \figours{SimPWM-\\LTX-2.3} & \figours{SimPWM-H3-C} \\[3pt]
\qrow{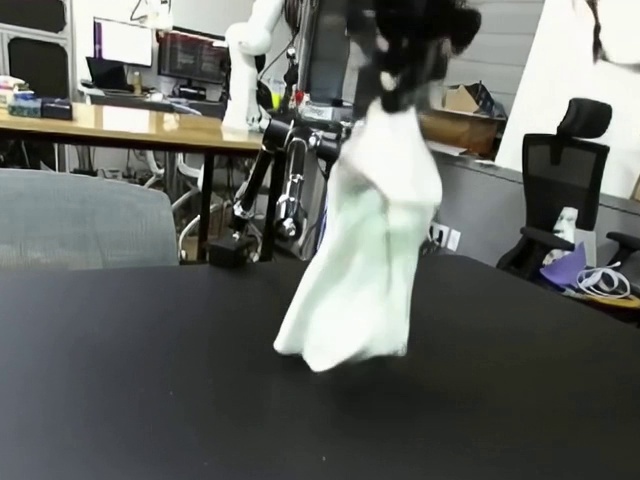} \\[2.5pt]
\multicolumn{7}{c}{\figdesc{Hang the cloth on the stand}} \\[5pt]
\qrow{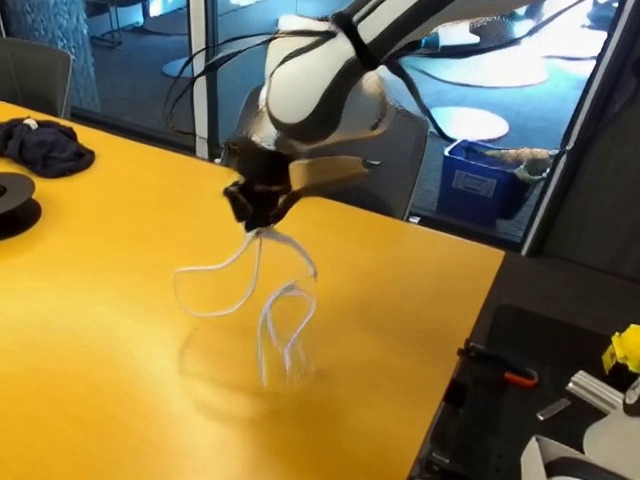} \\[2.5pt]
\multicolumn{7}{c}{\figdesc{Unfold the rope on the table}} \\
\end{tabular}
\captionof{figure}{\textbf{Qualitative comparison}, close-ups of a cloth and a rope. Ctrl-World blurs the rope; DreamDojo and Cosmos-Predict~2.5 misplace or distort the arm; SimPWM-H3-C, SimPWM-LTX-2.3 and SimPWM-Cosmos-Transfer carry the cloth and rope.}
\label{fig:qualitative}
}

\end{figure*}

\subsection{Setup}
\label{sec:exp_setup}
\paragraph{Data.}
We use DROID~\cite{khazatsky2024droid} (Franka with a Robotiq gripper), the first $97$ steps of each episode ($6.4$\,s at $15$\,fps), and group episodes by \emph{setup} (lab and camera pose).
The video models are trained on $3{,}000$ episodes from $839$ setups, the \mbox{-C} models on the consistent subset of $845$ of them (\hyperref[sec:method_render]{Sec.~\ref*{sec:method_render}}); the simulator has no trained parameters.
We test on $293$ episodes from $121$ setups that contribute neither training episodes nor the development episodes on which the simulator's thresholds were set, each with a closing gripper.

\paragraph{Metrics.}
\emph{Proxy} (\hyperref[tab:droid]{Table~\ref*{tab:droid}}): robot IoU against the robot mask from SAM~3~\cite{carion2025sam3}, and four measures of ours.
\emph{Grasp success}: the proxy grasps the object that the real gripper grasps (extracted with SAM~2~\cite{ravi2024sam2} where the closing gripper stalls), or neither grasps.
\emph{Object IoU}: the mask of the grasped object against the real target, over the clip, and $0$ unless both grasp the same object.
\emph{RGB L1} and \emph{depth MSE} (m$^2$): the proxy against the real video and its stereo depth~\cite{wen2025foundationstereo}, on the pixels that move in any row.
\emph{Generated video} (\hyperref[tab:droid_compare]{Tables~\ref*{tab:droid_compare}} and~\ref{tab:render}), against the real video: video quality (PSNR, LPIPS, FVD); motion (robot IoU; trajectory accuracy~\cite{shang2026worldarena} of the tracked robot; depth MSE of Depth Anything~V2~\cite{yang2024depthv2} on the generated frames against the stereo depth); physics (Physics-IQ~\cite{motamed2025physicsiq} with the robot masked out, interaction quality~\cite{shang2026worldarena}, the plausibility of the robot--object contact rated with the WorldArena prompt by a blind Claude judge; and our \emph{deformation IoU} of a held non-rigid target after aligning position and scale).

\paragraph{Baselines.}
All methods get the same first frame and actions; \hyperref[tab:droid_compare]{Table~\ref*{tab:droid_compare}} evaluates every video at $320{\times}192$ and $5$\,Hz ($33$ frames), the coarsest native output (Ctrl-World).
\emph{Ctrl-World}~\cite{guo2025ctrlworld} is trained on DROID except every $100$th episode.
\emph{DreamDojo}~\cite{gao2026dreamdojo} ($2$B) and \emph{Cosmos-Predict~2.5}~\cite{nvidia2025cosmospredict25} (its robot checkpoint, $2.1$B parameters, conditioned on end-effector motion and gripper opening) are post-trained on our $3{,}000$ clips with the settings of SimPWM-LTX-2.3 but their own action inputs and learning rates; at their native resolution they score similarly (\hyperref[tab:suppl_native]{Table~\ref*{tab:suppl_native}}).
SimPWM-Cosmos-Transfer shares its base with Cosmos-Predict~2.5.
Agentic Real2Sim~\cite{chen2026agentic} builds a MuJoCo scene from the recording with vision-language agents; on the $27$ test episodes it converts, its robot IoU is $0.574$ against $0.798$ for SimPWM-H3-C (\hyperref[sec:suppl_ar2s]{Sec.~\ref*{sec:suppl_ar2s}}).

\paragraph{Protocol and caveats.}
(i) Ctrl-World is trained on DROID including at least $234$ of our $293$ test episodes.
(ii) The models trained on all episodes use an earlier version of the simulator (grasp success $74.2\%$ vs.\ $69.5\%$; \hyperref[tab:variants]{Table~\ref*{tab:variants}}).
Details and per-row intervals: \hyperref[sec:suppl_impl]{Sec.~\ref*{sec:suppl_impl}}.

\begin{sidecapfloat}[t]{0.63}
\begin{table}[t]
\centering
\footnotesize
\setlength{\tabcolsep}{2.2pt}
\renewcommand{\arraystretch}{1.08}
\caption{\textbf{Ablation on simulator strategies} ($293$ test episodes). Each row changes one input or removes one part (grey: its replacement); best in bold.}
\label{tab:droid}
\tabstyle
\begin{tabular}{l||cc|c|cc}
\toprule
\multirow{2}{*}{\textbf{Variant}} & \multicolumn{2}{c|}{\textbf{IoU}$\uparrow$} & \textbf{Grasp} & \multicolumn{2}{c}{\textbf{Proxy}$\downarrow$} \\
\cmidrule(lr){2-3}\cmidrule(l){5-6}\tabcmidend
 & Robot & Object & succ.\,(\%)$\uparrow$ & RGB & Depth \\
\tabdouble
\tabgroup{6}{Our rule-based simulator} \\
\rowcolor{tabours} \textbf{Ours (full)} & \textbf{.787} & \textbf{.287} & \textbf{69.5} & \textbf{.097} & .165 \\
\hspace{0.6em}DROID logged calibration & .505 & .200 & 55.3 & .123 & .253 \\
\rowcolor{tabzebra} \hspace{0.6em}w/o stereo depth {\scriptsize\color{black!55}(mono)} & \textbf{.787} & .258 & 68.4 & \textbf{.097} & .222 \\
\hspace{0.6em}w/o action snap & \textbf{.787} & .185 & 67.6 & \textbf{.097} & .164 \\
\tabdouble
\tabgroup{6}{Learned dynamics} \\
PointWorld~\cite{huang2026pointworld} & \textbf{.787} & -- & -- & .117 & \textbf{.116} \\
\bottomrule
\end{tabular}
\end{table}

\end{sidecapfloat}
\subsection{Comparison}
\label{sec:exp_droid_compare}
\paragraph{DROID.}
\hyperref[tab:droid_compare]{Table~\ref*{tab:droid_compare}} compares SimPWM with the three baselines on the DROID test episodes with the metrics of \hyperref[sec:exp_setup]{Sec.~\ref*{sec:exp_setup}}; \hyperref[fig:teaser]{Figs.~\ref*{fig:teaser}}, \ref{fig:gallery} and~\ref{fig:qualitative} show examples.
The rendered robot makes the difference for the arm (\hyperref[sec:exp_render]{Sec.~\ref*{sec:exp_render}} isolates it): whatever their base model, the fine-tuned SimPWM variants give the lowest image error (PSNR, LPIPS) and follow the commanded motion most closely, whereas the end-to-end models drift from the commanded arm, DreamDojo most.
With the same base family, data and training steps, SimPWM-Cosmos-Transfer nearly doubles the robot IoU of Cosmos-Predict~2.5, so the gain comes from the proxy rather than from a larger model.
SimPWM-H3-C gives the lowest image error and FVD, level with SimPWM-H3: the consistent subset costs no image quality. Given our proxy without fine-tuning, MiniMax-H3 already follows the arm more closely than the end-to-end models, but fine-tuning improves image quality and arm following (FVD $194.1\to85.6$, robot IoU $0.718\to0.777$).
Ctrl-World scores highest on Physics-IQ, which counts every changing pixel; the SimPWM-H3 models and SimPWM-Cosmos-Transfer come close, whereas the output of SimPWM-LTX-2.3 flickers slightly in static regions.

\paragraph{Other robots.}
With LTX-2.3 retrained per robot, SimPWM ranks third of $19$ models in trajectory accuracy on the bimanual WorldArena benchmark but is weaker in image quality, and on the GR1 humanoid it follows the arms more closely than DreamDojo (robot IoU $0.861$ vs.\ $0.815$) with better PSNR and SSIM but worse LPIPS and FVD (\hyperref[sec:suppl_robots]{Sec.~\ref*{sec:suppl_robots}}).

\subsection{Real-to-Sim Evaluation}
\label{sec:exp_droid}
On the test episodes (\hyperref[tab:droid]{Table~\ref*{tab:droid}}, Ours), the proxy makes the same grasp as the real robot in $69.5\%$ of the decidable episodes; auditing $33$ rejections of the earlier simulator version found the real target extracted wrongly in $16$, which would raise its success from $74.2\%$ to $80.0\%$.
The rendered robot overlaps the real one closely, given the calibrated camera.
Object IoU is lower, since it also counts where the object is carried and placed, the main remaining error; the failures are grasps of a neighbouring object, motion beyond pick-and-place, and objects not in the first frame (\hyperref[sec:failures]{Sec.~\ref*{sec:failures}}).

\paragraph{Ablations.}
\begin{table*}[t]
\centering
\footnotesize
\setlength{\tabcolsep}{2.1pt}
\renewcommand{\arraystretch}{1.08}
\caption{\textbf{Sim-to-real ablation} with the ablation models (SimPWM-H3-C at lower resolution, one LoRA per reference video): generated at $640{\times}352$ and $24$\,fps, scored at $1280{\times}704$ and $15$\,fps (not comparable with \hyperref[tab:droid_compare]{Table~\ref*{tab:droid_compare}}). Deform.\ IoU: $24$ non-rigid episodes. Grasp: a blind Claude judge names the object lifted in eight frames of each video, or none; Acc.: the same answer as on the real video (object names matched by the judge); False: a pick-up where the real robot picks up none (unlike \hyperref[sec:evaluator]{Sec.~\ref*{sec:evaluator}}); Miss: no pick-up, and Right: a pick-up of the right object, where the real robot picks one up ($133$ episodes); bold: best (for False, excluding no proxy).}
\label{tab:render}
\tabstyle
\begin{adjustbox}{max width=\textwidth}
\begin{tabular}{l||ccc|ccc|ccc|cccc}
\toprule
\multirow{2}{*}{\textbf{Reference video}} & \multicolumn{3}{c|}{\textbf{Video quality}} & \multicolumn{3}{c|}{\textbf{Motion}} & \multicolumn{3}{c|}{\textbf{Physics}} & \multicolumn{4}{c}{\textbf{Grasp}} \\
\cmidrule(lr){2-4}\cmidrule(lr){5-7}\cmidrule(lr){8-10}\cmidrule(l){11-14}\tabcmidend
 & PSNR$\uparrow$ & LPIPS$\downarrow$ & FVD$\downarrow$ & Robot$\uparrow$ & Traj.$\uparrow$ & Depth$\downarrow$ & Phys-IQ$\uparrow$ & Interact.$\uparrow$ & Deform.\ IoU$\uparrow$ & Acc.\,(\%)$\uparrow$ & False\,(\%)$\downarrow$ & Miss\,(\%)$\downarrow$ & Right\,(\%)$\uparrow$ \\
\tabdouble
\tabgroup{14}{No simulated scene} \\
No proxy & 19.31 & .263 & 228.9 & .341 & .361 & .298 & 61.2 & .472 & .436 & 53.6 & 5.6 & 94.0 & 4.5 \\
\rowcolor{tabzebra} Robot only & 22.36 & .216 & 110.2 & .766 & .771 & .160 & 66.0 & .565 & .533 & 65.2 & \textbf{10.6} & 54.1 & 36.1 \\
\tabdouble
\tabgroup{14}{Simulated scene: learned dynamics, or our simulator in three encodings} \\
PointWorld~\cite{huang2026pointworld} & 22.59 & .211 & 105.5 & .766 & .756 & .161 & 67.1 & .564 & \textbf{.546} & 66.6 & \textbf{10.6} & 45.9 & 39.1 \\
\tabdash{1-14}
RGB proxy & 22.54 & .210 & 106.5 & .767 & \textbf{.775} & .161 & 66.6 & .566 & .507 & 67.6 & 12.5 & \textbf{37.6} & 43.6 \\
\rowcolor{tabours} \hspace{0.6em}$\Delta$ vs.\ robot only & $+$0.18 & $-$.006 & $-$3.7 & $+$.001 & $+$.004 & $+$.001 & $+$0.6 & $+$.001 & $-$.026 & $+$2.4 & $+$1.9 & $-16.5$ & $+7.5$ \\
\rowcolor{tabzebra} Depth proxy & 22.27 & .216 & 107.9 & .765 & .762 & .161 & 66.1 & .565 & .526 & \textbf{67.9} & 11.9 & 39.8 & 43.6 \\
RGB + depth & \textbf{22.80} & \textbf{.208} & \textbf{99.9} & \textbf{.777} & \textbf{.775} & \textbf{.156} & \textbf{67.4} & \textbf{.574} & .524 & 66.2 & 16.9 & 41.4 & \textbf{45.9} \\
\bottomrule
\end{tabular}
\end{adjustbox}
\par\vspace{10pt}
{\centering
\setlength{\tabcolsep}{1.2pt}
\newlength{\rdw}\setlength{\rdw}{\dimexpr(\textwidth-12\tabcolsep-15pt)/7\relax}
\newcommand{\rdimg}[1]{\rimg[width=\rdw]{fig/render/#1}}
\newcommand{\rdrow}[1]{\rdimg{#1_gt.jpg} & \rdimg{#1_noproxy.jpg} & \rdimg{#1_robot.jpg} & \rdimg{#1_pw.jpg} & \rdimg{#1_rgb.jpg} & \rdimg{#1_depth.jpg} & \rdimg{#1_rgb_depth.jpg}}
\renewcommand{\arraystretch}{0}
\begin{tabular}{@{}c@{\hspace{6pt}}cccccc@{}}
\figlab{GT} & \figlab{No proxy} & \figlab{Robot only} & \figlab{PointWorld} & \figlab{RGB proxy} & \figlab{Depth proxy} & \figlab{RGB + depth} \\[3pt]
\rdrow{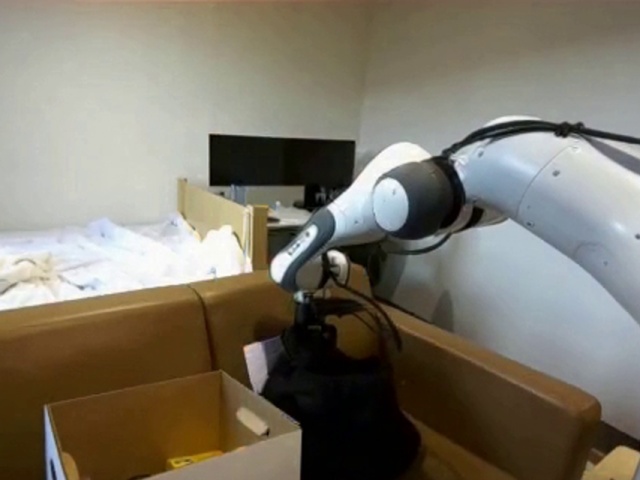} \\[2.5pt]
\multicolumn{7}{c}{\figdesc{Put the black hoodie inside the storage box}} \\[5pt]
\rdrow{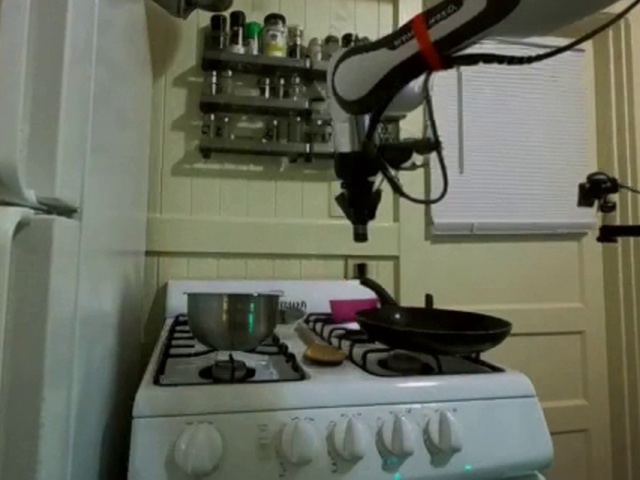} \\[2.5pt]
\multicolumn{7}{c}{\figdesc{Pour the contents from the purple bowl into the pan}} \\
\end{tabular}
\captionof{figure}{\textbf{Sim-to-real ablation} (\hyperref[tab:render]{Table~\ref*{tab:render}}). Without a proxy the arm goes astray; with any of the proxies the video follows the real arm.}
\label{fig:render}
}

\end{table*}
\hyperref[tab:droid]{Table~\ref*{tab:droid}} removes one part of the real-to-sim step at a time, and \hyperref[fig:ablation]{Fig.~\ref*{fig:ablation}} (Appendix) shows examples.
The rows replace our calibration by the one DROID logs, stereo by monocular depth, or remove the action snap; all are scored on the same episodes and targets.
Calibration matters most: with DROID's logged calibration, the closing point lands beside the target and many right grasps become wrong.
Without the action snap, grasp success changes little ($69.5\%$ vs.\ $67.6\%$, not significant), but the grasped objects are carried at an offset, since they stay at their first-frame depth: object IoU falls from $0.287$ to $0.185$.
Monocular depth raises the depth error but barely changes grasp success, as the snap corrects the depth of grasped objects.
PointWorld dynamics rarely lift objects, so its scene stays close to the first-frame depth and its depth error is lower.

\subsection{Sim-to-Real Evaluation}
\label{sec:exp_render}
\hyperref[tab:render]{Table~\ref*{tab:render}} varies the reference video of the ablation models (examples in \hyperref[fig:render]{Fig.~\ref*{fig:render}}): \emph{no proxy} (first frame and text), the robot alone, a proxy moved by PointWorld~\cite{huang2026pointworld} instead of our simulator, and our proxy as RGB, depth or both.
The robot-only row is the robot-render proxy of prior work~\cite{kim2026rofacto}; \hyperref[tab:render]{Table~\ref*{tab:render}} compares against it with the same base model, data and steps.
Global metrics (PSNR, FVD, Phys-IQ) barely separate the proxies: moved objects cover $0.36\%$ of the reference-video tokens (the arm $16\%$), so the simulator is judged on the grasp columns and the perturbation tests.
Any proxy improves on \emph{no proxy} by a wide margin (robot IoU $0.34\to0.77$, grasp agreement $+11.6$ to $+14.3$ points); the proxy forms differ little.
Against the robot alone, the simulated objects reduce missed pick-ups from $54.1\%$ to $37.6\%$ (paired difference $16.5$ points, $95\%$ CI $[8.3, 25.6]$) without significantly more false pick-ups ($+1.9$, $[-3.1, 6.9]$; $19$ of the $20$ occur where the simulator itself grasps); pick-ups of the right object rise by $+7.5$ points ($[-0.8, 15.8]$) and of another object by $+9.0$ ($[3.0, 15.8]$); where the simulator grasps the right object, the RGB proxy adds $24.6$ points of pick-ups ($[13.0, 36.2]$; \hyperref[tab:grasp_strata]{Table~\ref*{tab:grasp_strata}}).
With LTX-2.3 at full resolution, the RGB proxy also misses fewer pick-ups than the robot alone ($7.5$ points, $[1.5, 14.3]$), but only by picking up other objects ($+8.3$, $[0.8, 16.5]$), as LTX-2.3 follows the arm rather than the object state (\hyperref[sec:suppl_refs]{Sec.~\ref*{sec:suppl_refs}}).
\hyperref[sec:evaluator]{Sec.~\ref*{sec:evaluator}} tests whether the video follows the simulated objects when they no longer move with the arm.

\paragraph{Deformable objects.}
The simulator carries ropes and cloth rigidly, which can mislead the video (Deform.\ IoU $0.533\to0.507$ against the robot alone); on $24$ non-rigid episodes, a blind pairwise Claude judge prefers how the held object hangs in the videos of the ablation model and SimPWM-LTX-2.3 over their proxy, DreamDojo and Cosmos-Predict~2.5, and finds them about tied with Ctrl-World (\hyperref[sec:suppl_physics]{Sec.~\ref*{sec:suppl_physics}}).

\subsection{Following the Simulated State}
\label{sec:evaluator}
\begin{floatpair}[t]{0.45}{0.52}
\begin{figure}[t]
\centering
\caption{\textbf{Action perturbation} ($133$ episodes, blind Claude judge, \%): pick-ups under the recorded actions, with the gripper never closing, and with the simulated objects frozen. H3 zero-shot: MiniMax-H3 without fine-tuning; Ablation: models of \hyperref[tab:render]{Table~\ref*{tab:render}}; *: $640{\times}352$; proxy readout: the simulator's grasp state. More rows: \hyperref[sec:suppl_h3]{Sec.~\ref*{sec:suppl_h3}}.}
\label{fig:counterfactual}
\vspace{4pt}
\includegraphics[width=\columnwidth]{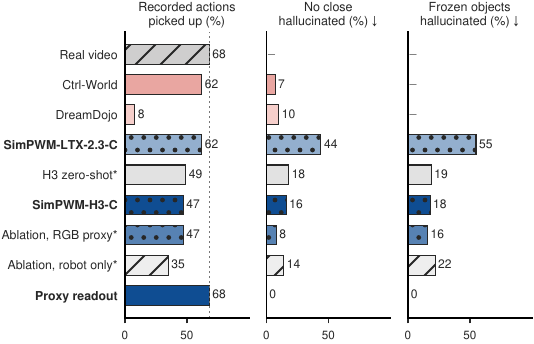}
\end{figure}
\pairnext
\begin{table}[t]
\centering\footnotesize\setlength{\tabcolsep}{2.4pt}
\renewcommand{\arraystretch}{1.08}
\caption{\textbf{Frozen objects by target type} (lifted, \%; $133$ episodes). Rec.: recorded actions; NC: no close; Fz.: frozen objects; Rel.: drop under frozen objects relative to Rec. A model that only lifts less often would keep its rate (Rel.\ near $0$).}
\label{tab:frozen}
\tabstyle
\begin{tabular}{l||cccc|cccc}
\toprule
\multirow{2}{*}{\textbf{Model}} & \multicolumn{4}{c|}{\textbf{Rigid} ($96$)} & \multicolumn{4}{c}{\textbf{Deformable} ($37$)} \\
\cmidrule(lr){2-5}\cmidrule(l){6-9}\tabcmidend
 & Rec. & NC & Fz. & Rel. & Rec. & NC & Fz. & Rel. \\
\tabdouble
\rowcolor{tabours} SimPWM-H3-C & 42 & 11 & 10 & 75\% & 59 & 27 & 38 & 36\% \\
H3 zero-shot$^*$ & 45 & 16 & 17 & 63\% & 59 & 24 & 24 & 59\% \\
\rowcolor{tabzebra} SimPWM-H3$^\S$ & 44 & 21 & 25 & 43\% & 70 & 59 & 59 & 15\% \\
SimPWM-LTX-2.3-C & 57 & 39 & 49 & 15\% & 73 & 59 & 70 & 4\% \\
\rowcolor{tabzebra} SimPWM-Cosmos-Transfer$^\S$ & 55 & 41 & 44 & 21\% & 68 & 46 & 57 & 16\% \\
\tabdouble
\rowcolor{tabzebra} Ablation$^*$, RGB proxy & 43 & 3 & 13 & 71\% & 59 & 22 & 24 & 59\% \\
Ablation$^*$, robot only & 27 & 10 & 15 & 46\% & 54 & 22 & 41 & 25\% \\
\bottomrule
\end{tabular}
\par\vspace{2pt}{\scriptsize H3 zero-shot: MiniMax-H3 without fine-tuning. $^*$\,$640{\times}352$. $^\S$\,Trained on all $3{,}000$ episodes, earlier simulator version. Rigid / deformable from keywords of the task instruction; all entries computed from the counts.}
\end{table}

\end{floatpair}
We take the $133$ test episodes in which the real robot lifts an object (grasp judge of \hyperref[tab:render]{Table~\ref*{tab:render}}) and perturb the input so that it should not (\hyperref[fig:perturb]{Fig.~\ref*{fig:perturb}} shows one episode, \hyperref[fig:counterfactual]{Fig.~\ref*{fig:counterfactual}} the rates).
A Claude judge, blind to the method, sees the first frame and eight frames from the first stall of the gripper on and says whether the target is lifted; seeing only these, it confirms $68\%$ of the lifts that the judge of \hyperref[tab:render]{Table~\ref*{tab:render}} finds on the whole clip, so we compare its rates only across methods. A blind human label agrees with it on $44$ of $46$ decided random sheets ($\kappa=0.91$; \hyperref[sec:suppl_h3]{Sec.~\ref*{sec:suppl_h3}}).

\paragraph{Open gripper.}
With the gripper never closing (\emph{no close}), SimPWM-H3-C shows a lift in $16\%$ of the videos, SimPWM-LTX-2.3-C in $44\%$ (on rigid targets $11\%$ vs.\ $39\%$; \hyperref[tab:frozen]{Table~\ref*{tab:frozen}}). $10$ of the $21$ lifts of SimPWM-H3-C are of deformable targets, whose shape the simulator does not model.
The ablation models lift in $8\%$ (RGB proxy) and $14\%$ (robot only): the video follows the rendered gripper.
Ctrl-World lifts even less ($7\%$) but has no object state to freeze; DreamDojo lifts little under any actions, Cosmos-Predict~2.5 in $11\%$ ($31\%$ when recorded).

\paragraph{Frozen objects.}
Freezing the simulated objects while the arm moves as recorded separates them from the gripper. SimPWM-H3-C lifts less often than SimPWM-LTX-2.3-C and the real videos ($47\%$, $62\%$ and $68\%$ under the recorded actions): the simulator lifts an object in $90$ of the $133$ episodes, and SimPWM-H3-C shows the lift in $54\%$ of these ($[43, 64]$) but in $30\%$ of the other $43$ (SimPWM-LTX-2.3-C $63\%$ and $58\%$). It might thus merely be conservative; but a conservative model would keep its rate when the objects are frozen.
On the $96$ rigid targets, SimPWM-H3-C drops from $42\%$ to $10\%$, and SimPWM-LTX-2.3-C only from $57\%$ to $49\%$ (\hyperref[tab:frozen]{Table~\ref*{tab:frozen}}); over all $133$ episodes, the paired difference of the drops is $22$ points ($95\%$ CI $[12, 31]$).
More than half of the remaining lifts of SimPWM-H3-C are of deformable targets ($14$ of $24$), which the simulator carries rigidly.
Freezing also closes the fingers fully when nothing is grasped, which the robot-only proxy isolates: in the ablation models it loses $12.8$ points against $31.6$ with the RGB proxy, so $18.8$ points ($95\%$ CI $[10, 28]$) come from the simulated object state.

\paragraph{Why base models differ.}
In consistent training data, a closing gripper that rises and a lifted object always occur together, so following the arm and following the simulated object fit the data equally well, and which one a model learns is set by its prior.
LTX-2.3 predicts moved objects mainly from the arm, which gives $91\%$ of the proxy's gain in their denoising loss.
MiniMax-H3 follows the object state without fine-tuning (H3 zero-shot loses $63\%$ of its lifts on rigid targets).
Episodes that disagree with the simulation tip the balance towards the arm: trained on all $3{,}000$ episodes (a quarter disagree on the grasp), SimPWM-H3 loses only $43\%$ of its rigid-target lifts under frozen objects, against $75\%$ for SimPWM-H3-C (\hyperref[tab:frozen]{Table~\ref*{tab:frozen}}). SimPWM-Cosmos-Transfer, trained on the data of SimPWM-H3, loses $21\%$; on all $133$ episodes its drop is $17$ points ($[8, 26]$) below that of SimPWM-H3-C and close to those of SimPWM-LTX-2.3-C and SimPWM-H3.
We hypothesise that the prior of MiniMax-H3 comes from its Ref2VA pretraining, which asks the output to follow a reference video, whereas LTX-2.3 learns to use a reference only in fine-tuning (details: \hyperref[sec:suppl_h3]{Sec.~\ref*{sec:suppl_h3}}).

\section{Discussion}
\label{sec:conclusion}
Does the video follow the simulated state? For the arm, yes, with each of three base video models; for the objects, only if the prior of the base model favours the reference video (MiniMax-H3, possibly from pretraining), whereas LTX-2.3 infers grasps from the arm motion. Consistent data alone cannot tell the two apart, so our recipe is: fine-tune on consistent episodes and choose a base model that passes the perturbation tests.

\paragraph{Limitations.}
\label{sec:limitations}
(1) Pushing, pouring, folding and articulation are not modelled, only objects in the first frame are simulated, ropes and cloth are carried rigidly, and the video model does not add what the simulator lacks.
(2) The simulator trusts the gripper signal, and a closing point a few pixels off can make it grasp a neighbouring object; its grasp matches the real one in ${\sim}70\%$ of episodes, too rarely to flag errors under recorded actions.
(3) Poor calibration or monocular depth lowers accuracy (\hyperref[tab:droid]{Table~\ref*{tab:droid}}).
(4) Actions are replayed, not closed-loop.
(5) Metrics rely on SAM tracks and a VLM judge (checked against $60$ human labels).

\clearpage
\bibliographystyle{plainnat}
\bibliography{main}

\begin{thebibliography}{69}
\providecommand{\natexlab}[1]{#1}
\providecommand{\url}[1]{\texttt{#1}}
\expandafter\ifx\csname urlstyle\endcsname\relax
  \providecommand{\doi}[1]{doi: #1}\else
  \providecommand{\doi}{doi: \begingroup \urlstyle{rm}\Url}\fi

\bibitem[{Alibaba Cloud}(2025)]{alibaba2025wan26}
{Alibaba Cloud}.
\newblock Alibaba unveils {Wan2.6} series enabling everyone to star in videos.
\newblock
  \url{https://www.alibabacloud.com/press-room/alibaba-unveils-wan2-6-series-enabling-everyone},
  2025.
\newblock Press release, December 16, 2025.

\bibitem[Bansal et~al.(2026)Bansal, Peng, Bitton, Goldenberg, Grover, and
  Chang]{bansal2025videophy2}
Hritik Bansal, Clark Peng, Yonatan Bitton, Roman Goldenberg, Aditya Grover, and
  Kai-Wei Chang.
\newblock {VideoPhy-2}: A challenging action-centric physical commonsense
  evaluation in video generation.
\newblock In \emph{ICLR}, 2026.

\bibitem[Bruce et~al.(2024)Bruce, Dennis, Edwards, Parker-Holder, Shi,
  et~al.]{bruce2024genie}
Jake Bruce, Michael Dennis, Ashley Edwards, Jack Parker-Holder, Yuge Shi,
  et~al.
\newblock Genie: Generative interactive environments.
\newblock In \emph{ICML}, 2024.

\bibitem[Carion et~al.(2026)Carion, Gustafson, Hu, Debnath, Hu,
  et~al.]{carion2025sam3}
Nicolas Carion, Laura Gustafson, Yuan-Ting Hu, Shoubhik Debnath, Ronghang Hu,
  et~al.
\newblock {SAM} 3: Segment anything with concepts.
\newblock In \emph{ICLR}, 2026.

\bibitem[Chen et~al.(2026)Chen, Xia, Peng, Zhang, Jing, Ma,
  et~al.]{chen2026agentic}
Guanxiong Chen, Qianjun Xia, Jiawei Peng, Heng Zhang, Pengyu Jing, Bole Ma,
  et~al.
\newblock Agentic {Real2Sim}: Physics-based world modeling with vision-language
  agents.
\newblock \emph{arXiv preprint arXiv:2607.19190}, 2026.

\bibitem[Chi et~al.(2025)Chi, Jia, Fan, Ju, Mi, Zhang, et~al.]{chi2025wow}
Xiaowei Chi, Peidong Jia, Chun-Kai Fan, Xiaozhu Ju, Weishi Mi, Kevin Zhang,
  et~al.
\newblock {WoW}: Towards a world omniscient world model through embodied
  interaction.
\newblock \emph{arXiv preprint arXiv:2509.22642}, 2025.

\bibitem[Du et~al.(2023)Du, Yang, Dai, Dai, Nachum, Tenenbaum, Schuurmans, and
  Abbeel]{du2023unipi}
Yilun Du, Mengjiao Yang, Bo~Dai, Hanjun Dai, Ofir Nachum, Josh Tenenbaum, Dale
  Schuurmans, and Pieter Abbeel.
\newblock Learning universal policies via text-guided video generation.
\newblock In \emph{NeurIPS}, 2023.

\bibitem[Ebert et~al.(2018)Ebert, Finn, Dasari, Xie, Lee, and
  Levine]{ebert2018visualforesight}
Frederik Ebert, Chelsea Finn, Sudeep Dasari, Annie Xie, Alex Lee, and Sergey
  Levine.
\newblock Visual foresight: Model-based deep reinforcement learning for
  vision-based robotic control.
\newblock \emph{arXiv preprint arXiv:1812.00568}, 2018.

\bibitem[Esser et~al.(2023)Esser, Chiu, Atighehchian, Granskog, and
  Germanidis]{esser2023gen1}
Patrick Esser, Johnathan Chiu, Parmida Atighehchian, Jonathan Granskog, and
  Anastasis Germanidis.
\newblock Structure and content-guided video synthesis with diffusion models.
\newblock In \emph{ICCV}, 2023.

\bibitem[Feng et~al.(2025)Feng, Tan, Mao, Xiang, Liu, Huang, Su, and
  Zhu]{feng2025vidar}
Yao Feng, Hengkai Tan, Xinyi Mao, Chendong Xiang, Guodong Liu, Shuhe Huang,
  Hang Su, and Jun Zhu.
\newblock {Vidar}: Embodied video diffusion model for generalist manipulation.
\newblock \emph{arXiv preprint arXiv:2507.12898}, 2025.

\bibitem[Finn et~al.(2016)Finn, Goodfellow, and Levine]{finn2016unsupervised}
Chelsea Finn, Ian Goodfellow, and Sergey Levine.
\newblock Unsupervised learning for physical interaction through video
  prediction.
\newblock In \emph{NeurIPS}, 2016.

\bibitem[Fu et~al.(2026)Fu, Wang, Liu, Bai, Xu, Wan, Zhang, and
  Lin]{fu2025robomaster}
Xiao Fu, Xintao Wang, Xian Liu, Jianhong Bai, Runsen Xu, Pengfei Wan, Di~Zhang,
  and Dahua Lin.
\newblock Learning video generation for robotic manipulation with collaborative
  trajectory control.
\newblock In \emph{ICLR}, 2026.

\bibitem[Gao et~al.(2026)Gao, Liang, Zheng, Malik, Ye, Yu,
  et~al.]{gao2026dreamdojo}
Shenyuan Gao, William Liang, Kaiyuan Zheng, Ayaan Malik, Seonghyeon Ye, Sihyun
  Yu, et~al.
\newblock {DreamDojo}: A generalist robot world model from large-scale human
  videos.
\newblock In \emph{ICML}, 2026.

\bibitem[{GigaWorld Team} et~al.(2025){GigaWorld Team}, Ye, Wang, Ni, Huang,
  Zhao, et~al.]{gigaworld2025gigaworld0}
{GigaWorld Team}, Angen Ye, Boyuan Wang, Chaojun Ni, Guan Huang, Guosheng Zhao,
  et~al.
\newblock {GigaWorld-0}: World models as data engine to empower embodied {AI}.
\newblock \emph{arXiv preprint arXiv:2511.19861}, 2025.

\bibitem[{GigaWorld Team} et~al.(2026){GigaWorld Team}, Ma, Wang, Li,
  et~al.]{gigaworld2026gigaworld1}
{GigaWorld Team}, Angyuan Ma, Boyuan Wang, Bohan Li, et~al.
\newblock {GigaWorld-1}: A roadmap to build world models for robot policy
  evaluation.
\newblock \emph{arXiv preprint arXiv:2607.02642}, 2026.

\bibitem[{Google DeepMind}(2025)]{google2025veo31}
{Google DeepMind}.
\newblock {Veo 3.1}.
\newblock \url{https://deepmind.google/models/veo/}, 2025.
\newblock Model page; announced October 15, 2025.

\bibitem[Gu et~al.(2025)Gu, Yan, Lu, Li, Dou, Si, Dong, Liu, Lin, Liu, Wang,
  and Liu]{gu2025das}
Zekai Gu, Rui Yan, Jiahao Lu, Peng Li, Zhiyang Dou, Chenyang Si, Zhen Dong,
  Qifeng Liu, Cheng Lin, Ziwei Liu, Wenping Wang, and Yuan Liu.
\newblock Diffusion as shader: {3D}-aware video diffusion for versatile video
  generation control.
\newblock In \emph{SIGGRAPH}, 2025.

\bibitem[Guo et~al.(2025)Guo, Ma, Wang, Yang, Liu, and Li]{guo2025flowdreamer}
Jun Guo, Xiaojian Ma, Yikai Wang, Min Yang, Huaping Liu, and Qing Li.
\newblock {FlowDreamer}: A {RGB-D} world model with flow-based motion
  representations for robot manipulation.
\newblock \emph{arXiv preprint arXiv:2505.10075}, 2025.

\bibitem[Guo et~al.(2026)Guo, Shi, Chen, and Finn]{guo2025ctrlworld}
Yanjiang Guo, Lucy~Xiaoyang Shi, Jianyu Chen, and Chelsea Finn.
\newblock {Ctrl-World}: A controllable generative world model for robot
  manipulation.
\newblock In \emph{ICLR}, 2026.

\bibitem[Guo et~al.(2024)Guo, Yang, Rao, Agrawala, Lin, and
  Dai]{guo2024sparsectrl}
Yuwei Guo, Ceyuan Yang, Anyi Rao, Maneesh Agrawala, Dahua Lin, and Bo~Dai.
\newblock {SparseCtrl}: Adding sparse controls to text-to-video diffusion
  models.
\newblock In \emph{ECCV}, 2024.

\bibitem[Ha and Schmidhuber(2018)]{ha2018worldmodels}
David Ha and J{\"u}rgen Schmidhuber.
\newblock Recurrent world models facilitate policy evolution.
\newblock In \emph{NeurIPS}, 2018.

\bibitem[HaCohen et~al.(2026)HaCohen, Brazowski, Chiprut, Bitterman, Kvochko,
  Berkowitz, et~al.]{hacohen2026ltx2}
Yoav HaCohen, Benny Brazowski, Nisan Chiprut, Yaki Bitterman, Andrew Kvochko,
  Avishai Berkowitz, et~al.
\newblock {LTX-2}: Efficient joint audio-visual foundation model.
\newblock \emph{arXiv preprint arXiv:2601.03233}, 2026.

\bibitem[Hafner et~al.(2025)Hafner, Pasukonis, Ba, and
  Lillicrap]{hafner2023dreamerv3}
Danijar Hafner, Jurgis Pasukonis, Jimmy Ba, and Timothy Lillicrap.
\newblock Mastering diverse control tasks through world models.
\newblock \emph{Nature}, 640:\penalty0 647--653, 2025.

\bibitem[Huang et~al.(2024)Huang, Wang, Wu, Shi, Dou, Liang,
  et~al.]{huang2024iclora}
Lianghua Huang, Wei Wang, Zhi-Fan Wu, Yupeng Shi, Huanzhang Dou, Chen Liang,
  et~al.
\newblock In-context {LoRA} for diffusion transformers.
\newblock \emph{arXiv preprint arXiv:2410.23775}, 2024.

\bibitem[Huang et~al.(2026)Huang, Chao, Mousavian, Liu, Fox, Mo, and
  Fei-Fei]{huang2026pointworld}
Wenlong Huang, Yu-Wei Chao, Arsalan Mousavian, Ming-Yu Liu, Dieter Fox, Kaichun
  Mo, and Li~Fei-Fei.
\newblock {PointWorld}: Scaling {3D} world models for in-the-wild robotic
  manipulation.
\newblock In \emph{CVPR}, 2026.

\bibitem[Jiang et~al.(2025{\natexlab{a}})Jiang, Hsu, Zhang, Yu, Wang, and
  Li]{jiang2025phystwin}
Hanxiao Jiang, Hao-Yu Hsu, Kaifeng Zhang, Hsin-Ni Yu, Shenlong Wang, and Yunzhu
  Li.
\newblock {PhysTwin}: Physics-informed reconstruction and simulation of
  deformable objects from videos.
\newblock In \emph{ICCV}, 2025{\natexlab{a}}.

\bibitem[Jiang et~al.(2025{\natexlab{b}})Jiang, Chen, Huang, Chen, Zhou,
  et~al.]{jiang2025enerverseac}
Yuxin Jiang, Shengcong Chen, Siyuan Huang, Liliang Chen, Pengfei Zhou, et~al.
\newblock {EnerVerse-AC}: Envisioning embodied environments with action
  condition.
\newblock In \emph{NeurIPS}, 2025{\natexlab{b}}.

\bibitem[Jiang et~al.(2026)Jiang, Zhou, Jiang, Huang, et~al.]{jiang2026wovr}
Zhennan Jiang, Shangqing Zhou, Yutong Jiang, Zefang Huang, et~al.
\newblock {WoVR}: World models as reliable simulators for post-training {VLA}
  policies with {RL}.
\newblock \emph{arXiv preprint arXiv:2602.13977}, 2026.

\bibitem[Khazatsky et~al.(2024)Khazatsky, Pertsch, Nair, Balakrishna, Dasari,
  et~al.]{khazatsky2024droid}
Alexander Khazatsky, Karl Pertsch, Suraj Nair, Ashwin Balakrishna, Sudeep
  Dasari, et~al.
\newblock {DROID}: A large-scale in-the-wild robot manipulation dataset.
\newblock In \emph{RSS}, 2024.

\bibitem[Kim et~al.(2026)Kim, Kim, Cha, and Joo]{kim2026rofacto}
Byungjun Kim, Taeksoo Kim, Hyunsoo Cha, and Hanbyul Joo.
\newblock Robot-factored world models via robot rendering.
\newblock In \emph{CoRL}, 2026.

\bibitem[Ko et~al.(2024)Ko, Mao, Du, Sun, and Tenenbaum]{ko2024avdc}
Po-Chen Ko, Jiayuan Mao, Yilun Du, Shao-Hua Sun, and Joshua~B. Tenenbaum.
\newblock Learning to act from actionless videos through dense correspondences.
\newblock In \emph{ICLR}, 2024.

\bibitem[Li et~al.(2024)Li, Hsu, Gu, Pertsch, Mees, et~al.]{li2024simpler}
Xuanlin Li, Kyle Hsu, Jiayuan Gu, Karl Pertsch, Oier Mees, et~al.
\newblock Evaluating real-world robot manipulation policies in simulation.
\newblock In \emph{CoRL}, 2024.

\bibitem[Li et~al.(2025{\natexlab{a}})Li, Wei, Chi, Li, Zhao,
  et~al.]{li2025manipdreamer}
Ying Li, Xiaobao Wei, Xiaowei Chi, Yuming Li, Zhongyu Zhao, et~al.
\newblock {ManipDreamer}: Boosting robotic manipulation world model with action
  tree and visual guidance.
\newblock \emph{arXiv preprint arXiv:2504.16464}, 2025{\natexlab{a}}.

\bibitem[Li et~al.(2025{\natexlab{b}})Li, Yu, Liu, Yang, Herrmann, Wetzstein,
  and Wu]{li2025wonderplay}
Zizhang Li, Hong-Xing Yu, Wei Liu, Yin Yang, Charles Herrmann, Gordon
  Wetzstein, and Jiajun Wu.
\newblock {WonderPlay}: Dynamic {3D} scene generation from a single image and
  actions.
\newblock In \emph{ICCV}, 2025{\natexlab{b}}.

\bibitem[Liao et~al.(2026{\natexlab{a}})Liao, Liang, Li, and
  Liu]{liao2026memorizon}
Tingting Liao, Xuezhi Liang, Hao Li, and Guangyi Liu.
\newblock Memorizon: Training world models beyond their context window.
\newblock \emph{arXiv preprint arXiv:2610.00544}, 2026{\natexlab{a}}.

\bibitem[Liao et~al.(2026{\natexlab{b}})Liao, Zhou, Huang, Yang, Chen, Jiang,
  et~al.]{liao2025genieenvisioner}
Yue Liao, Pengfei Zhou, Siyuan Huang, Donglin Yang, Shengcong Chen, Yuxin
  Jiang, et~al.
\newblock {Genie Envisioner}: A unified world foundation platform for robotic
  manipulation.
\newblock In \emph{ICLR}, 2026{\natexlab{b}}.

\bibitem[Liu et~al.(2024)Liu, Ren, Gupta, and Wang]{liu2024physgen}
Shaowei Liu, Zhongzheng Ren, Saurabh Gupta, and Shenlong Wang.
\newblock {PhysGen}: Rigid-body physics-grounded image-to-video generation.
\newblock In \emph{ECCV}, 2024.

\bibitem[Mallya et~al.(2020)Mallya, Wang, Sapra, and
  Liu]{mallya2020worldconsistent}
Arun Mallya, Ting-Chun Wang, Karan Sapra, and Ming-Yu Liu.
\newblock World-consistent video-to-video synthesis.
\newblock In \emph{ECCV}, 2020.

\bibitem[{MiniMax}(2026)]{minimax2026h3}
{MiniMax}.
\newblock {MiniMax H3}: An open model breaking the boundaries between tasks and
  modalities.
\newblock Blog post, \url{https://minimax.io/blog/minimax-h3}; model weights
  and code: \url{https://huggingface.co/MiniMaxAI/MiniMax-H3},
  \url{https://github.com/MiniMax-AI/MiniMax-H3}, July 2026.
\newblock Accessed November 2026.

\bibitem[Motamed et~al.(2026)Motamed, Culp, Swersky, Jaini, and
  Geirhos]{motamed2025physicsiq}
Saman Motamed, Laura Culp, Kevin Swersky, Priyank Jaini, and Robert Geirhos.
\newblock Do generative video models understand physical principles?
\newblock In \emph{WACV}, 2026.

\bibitem[{NVIDIA} et~al.(2025{\natexlab{a}}){NVIDIA}, Abu~Alhaija, Alvarez,
  Bala, et~al.]{nvidia2025cosmostransfer}
{NVIDIA}, Hassan Abu~Alhaija, Jose Alvarez, Maciej Bala, et~al.
\newblock Cosmos-{Transfer1}: Conditional world generation with adaptive
  multimodal control.
\newblock \emph{arXiv preprint arXiv:2503.14492}, 2025{\natexlab{a}}.

\bibitem[{NVIDIA} et~al.(2025{\natexlab{b}}){NVIDIA}, Agarwal, Ali, Bala,
  et~al.]{agarwal2025cosmos}
{NVIDIA}, Niket Agarwal, Arslan Ali, Maciej Bala, et~al.
\newblock Cosmos world foundation model platform for physical {AI}.
\newblock \emph{arXiv preprint arXiv:2501.03575}, 2025{\natexlab{b}}.

\bibitem[{NVIDIA} et~al.(2025{\natexlab{c}}){NVIDIA}, Ali, Bai, Bala, Balaji,
  Blakeman, et~al.]{nvidia2025cosmospredict25}
{NVIDIA}, Arslan Ali, Junjie Bai, Maciej Bala, Yogesh Balaji, Aaron Blakeman,
  et~al.
\newblock World simulation with video foundation models for physical {AI}.
\newblock \emph{arXiv preprint arXiv:2511.00062}, 2025{\natexlab{c}}.

\bibitem[{NVIDIA} et~al.(2025{\natexlab{d}}){NVIDIA}, Bjorck, Casta{\~n}eda,
  Cherniadev, Da, Ding, et~al.]{nvidia2025gr00t}
{NVIDIA}, Johan Bjorck, Fernando Casta{\~n}eda, Nikita Cherniadev, Xingye Da,
  Runyu Ding, et~al.
\newblock {GR00T N1}: An open foundation model for generalist humanoid robots.
\newblock \emph{arXiv preprint arXiv:2503.14734}, 2025{\natexlab{d}}.

\bibitem[Qiu et~al.(2026)Qiu, Chen, Liao, Wang, et~al.]{qiu2026gesim2}
Boxiang Qiu, Liliang Chen, Yue Liao, Nan Wang, et~al.
\newblock {GE-Sim 2.0}: A roadmap towards comprehensive closed-loop video world
  simulators for robotic manipulation.
\newblock \emph{arXiv preprint arXiv:2605.27491}, 2026.

\bibitem[Quevedo et~al.(2026)Quevedo, Sharma, Sun, Suryavanshi, Liang, and
  Yang]{quevedo2026worldgym}
Julian Quevedo, Ansh~Kumar Sharma, Yixiang Sun, Varad Suryavanshi, Percy Liang,
  and Sherry Yang.
\newblock {WorldGym}: World model as an environment for policy evaluation.
\newblock In \emph{ICLR}, 2026.

\bibitem[Ranawaka et~al.(2026)Ranawaka, Wong, Pai, Chu, Dai,
  et~al.]{ranawaka2026simfoundry}
Nadun Ranawaka, Josiah Wong, Wei-Lin Pai, Wei-Teng Chu, Tianyuan Dai, et~al.
\newblock {SimFoundry}: Modular and automated scene generation for policy
  learning and evaluation.
\newblock \emph{arXiv preprint arXiv:2606.28276}, 2026.

\bibitem[Ravi et~al.(2025)Ravi, Gabeur, Hu, Hu, Ryali, et~al.]{ravi2024sam2}
Nikhila Ravi, Valentin Gabeur, Yuan-Ting Hu, Ronghang Hu, Chaitanya Ryali,
  et~al.
\newblock {SAM} 2: Segment anything in images and videos.
\newblock In \emph{ICLR}, 2025.

\bibitem[Shang et~al.(2025)Shang, Zhang, Tang, Jin, Gao, Wu, and
  Li]{shang2025roboscape}
Yu~Shang, Xin Zhang, Yinzhou Tang, Lei Jin, Chen Gao, Wei Wu, and Yong Li.
\newblock {RoboScape}: Physics-informed embodied world model.
\newblock In \emph{NeurIPS}, 2025.

\bibitem[Shang et~al.(2026)Shang, Li, Ma, Su, Jin, Wang, Jin, Zhang, Tang, Su,
  Gao, Wu, Liu, Shah, Zhang, Chen, Zhu, Tian, Chua, Zhu, and
  Li]{shang2026worldarena}
Yu~Shang, Zhuohang Li, Yiding Ma, Weikang Su, Xin Jin, Ziyou Wang, Lei Jin, Xin
  Zhang, Yinzhou Tang, Haisheng Su, Chen Gao, Wei Wu, Xihui Liu, Dhruv Shah,
  Zhaoxiang Zhang, Zhibo Chen, Jun Zhu, Yonghong Tian, Tat-Seng Chua, Wenwu
  Zhu, and Yong Li.
\newblock {WorldArena}: A unified benchmark for evaluating perception and
  functional utility of embodied world models.
\newblock \emph{arXiv preprint arXiv:2602.08971}, 2026.

\bibitem[{Team Wan} et~al.(2025){Team Wan}, Wang, Ai, Wen, Mao, Xie,
  et~al.]{wan2025wan}
{Team Wan}, Ang Wang, Baole Ai, Bin Wen, Chaojie Mao, Chen-Wei Xie, et~al.
\newblock {Wan}: Open and advanced large-scale video generative models.
\newblock \emph{arXiv preprint arXiv:2503.20314}, 2025.

\bibitem[Todorov et~al.(2012)Todorov, Erez, and Tassa]{todorov2012mujoco}
Emanuel Todorov, Tom Erez, and Yuval Tassa.
\newblock {MuJoCo}: A physics engine for model-based control.
\newblock In \emph{IROS}, 2012.

\bibitem[Torne et~al.(2024)Torne, Simeonov, Li, Chan, Chen, Gupta, and
  Agrawal]{torne2024rialto}
Marcel Torne, Anthony Simeonov, Zechu Li, April Chan, Tao Chen, Abhishek Gupta,
  and Pulkit Agrawal.
\newblock Reconciling reality through simulation: A real-to-sim-to-real
  approach for robust manipulation.
\newblock In \emph{RSS}, 2024.

\bibitem[Valevski et~al.(2025)Valevski, Leviathan, Arar, and
  Fruchter]{valevski2025gamengen}
Dani Valevski, Yaniv Leviathan, Moab Arar, and Shlomi Fruchter.
\newblock Diffusion models are real-time game engines.
\newblock In \emph{ICLR}, 2025.

\bibitem[Wang et~al.(2026)Wang, Chou, Zhang, Wang, Huang, Yu, Nabi, and
  Li]{wang2026worldsim3d}
Jimmy Wang, Yifan Chou, Kaifeng Zhang, Yixuan Wang, Binghao Huang, Tao Yu,
  Saleh Nabi, and Yunzhu Li.
\newblock Hybrid particle-voxel dynamics for multimodal {3D} world simulation.
\newblock \emph{arXiv preprint arXiv:2611.04492}, 2026.

\bibitem[Wang et~al.(2018)Wang, Liu, Zhu, Liu, Tao, Kautz, and
  Catanzaro]{wang2018vid2vid}
Ting-Chun Wang, Ming-Yu Liu, Jun-Yan Zhu, Guilin Liu, Andrew Tao, Jan Kautz,
  and Bryan Catanzaro.
\newblock Video-to-video synthesis.
\newblock In \emph{NeurIPS}, 2018.

\bibitem[Wang et~al.(2023)Wang, Yuan, Zhang, Chen, Wang,
  et~al.]{wang2023videocomposer}
Xiang Wang, Hangjie Yuan, Shiwei Zhang, Dayou Chen, Jiuniu Wang, et~al.
\newblock {VideoComposer}: Compositional video synthesis with motion
  controllability.
\newblock In \emph{NeurIPS}, 2023.

\bibitem[Wen et~al.(2025)Wen, Trepte, Aribido, Kautz, Gallo, and
  Birchfield]{wen2025foundationstereo}
Bowen Wen, Matthew Trepte, Joseph Aribido, Jan Kautz, Orazio Gallo, and Stan
  Birchfield.
\newblock {FoundationStereo}: Zero-shot stereo matching.
\newblock In \emph{CVPR}, 2025.

\bibitem[Whitney et~al.(2024)Whitney, Varley, Jain, Choromanski, Singh, and
  Sindhwani]{whitney2024hdvpd}
William~F. Whitney, Jacob Varley, Deepali Jain, Krzysztof Choromanski, Sumeet
  Singh, and Vikas Sindhwani.
\newblock Modeling the real world with high-density visual particle dynamics.
\newblock \emph{arXiv preprint arXiv:2406.19800}, 2024.

\bibitem[Wu et~al.(2024)Wu, Yin, Feng, He, Li, Hao, and Long]{wu2024ivideogpt}
Jialong Wu, Shaofeng Yin, Ningya Feng, Xu~He, Dong Li, Jianye Hao, and
  Mingsheng Long.
\newblock {iVideoGPT}: Interactive {VideoGPTs} are scalable world models.
\newblock In \emph{NeurIPS}, 2024.

\bibitem[Wu et~al.(2022)Wu, Escontrela, Hafner, Abbeel, and
  Goldberg]{wu2022daydreamer}
Philipp Wu, Alejandro Escontrela, Danijar Hafner, Pieter Abbeel, and Ken
  Goldberg.
\newblock {DayDreamer}: World models for physical robot learning.
\newblock In \emph{CoRL}, 2022.

\bibitem[Yang et~al.(2024{\natexlab{a}})Yang, Kang, Huang, Zhao, Xu, Feng, and
  Zhao]{yang2024depthv2}
Lihe Yang, Bingyi Kang, Zilong Huang, Zhen Zhao, Xiaogang Xu, Jiashi Feng, and
  Hengshuang Zhao.
\newblock Depth anything {V2}.
\newblock In \emph{NeurIPS}, 2024{\natexlab{a}}.

\bibitem[Yang et~al.(2024{\natexlab{b}})Yang, Du, Ghasemipour, Tompson,
  Kaelbling, Schuurmans, and Abbeel]{yang2024unisim}
Sherry Yang, Yilun Du, Kamyar Ghasemipour, Jonathan Tompson, Leslie Kaelbling,
  Dale Schuurmans, and Pieter Abbeel.
\newblock Learning interactive real-world simulators.
\newblock In \emph{ICLR}, 2024{\natexlab{b}}.

\bibitem[Yang et~al.(2026)Yang, Li, Xu, Wang, Ye, Chen, et~al.]{yang2026orv}
Xiuyu Yang, Bohan Li, Shaocong Xu, Nan Wang, Chongjie Ye, Zhaoxi Chen, et~al.
\newblock {ORV}: {4D} occupancy-centric robot video generation.
\newblock In \emph{CVPR}, 2026.

\bibitem[Yang et~al.(2025)Yang, Teng, Zheng, Ding, Huang, Xu,
  et~al.]{yang2025cogvideox}
Zhuoyi Yang, Jiayan Teng, Wendi Zheng, Ming Ding, Shiyu Huang, Jiazheng Xu,
  et~al.
\newblock {CogVideoX}: Text-to-video diffusion models with an expert
  transformer.
\newblock In \emph{ICLR}, 2025.

\bibitem[Yu et~al.(2025)Yu, Fu, Huang, El-Refai, Ambrus, Cheng, Irshad, and
  Goldberg]{yu2025r2r2r}
Justin Yu, Letian Fu, Huang Huang, Karim El-Refai, Rares~Andrei Ambrus, Richard
  Cheng, Muhammad~Zubair Irshad, and Ken Goldberg.
\newblock {Real2Render2Real}: Scaling robot data without dynamics simulation or
  robot hardware.
\newblock In \emph{CoRL}, 2025.

\bibitem[Zhang et~al.(2023)Zhang, Rao, and Agrawala]{zhang2023controlnet}
Lvmin Zhang, Anyi Rao, and Maneesh Agrawala.
\newblock Adding conditional control to text-to-image diffusion models.
\newblock In \emph{ICCV}, 2023.

\bibitem[Zhen et~al.(2025)Zhen, Sun, Zhang, Li, Zhou, Du, and
  Gan]{zhen2025tesseract}
Haoyu Zhen, Qiao Sun, Hongxin Zhang, Junyan Li, Siyuan Zhou, Yilun Du, and
  Chuang Gan.
\newblock {TesserAct}: Learning {4D} embodied world models.
\newblock In \emph{ICCV}, 2025.

\bibitem[Zhu et~al.(2025)Zhu, Wu, Guo, Liu, Cheang, and Kong]{zhu2024irasim}
Fangqi Zhu, Hongtao Wu, Song Guo, Yuxiao Liu, Chilam Cheang, and Tao Kong.
\newblock {IRASim}: A fine-grained world model for robot manipulation.
\newblock In \emph{ICCV}, 2025.

\end{thebibliography}

\clearpage
\beginappendix
\setcounter{dbltopnumber}{3}\renewcommand{\dbltopfraction}{0.95}\renewcommand{\dblfloatpagefraction}{0.85}
\setcounter{topnumber}{3}\renewcommand{\topfraction}{0.95}\setcounter{totalnumber}{5}\renewcommand{\textfraction}{0.05}\renewcommand{\floatpagefraction}{0.85}
\setcounter{figure}{0}
\setcounter{table}{0}
\renewcommand{\thesection}{\Alph{section}}\renewcommand{\theHsection}{\Alph{section}}
\counterwithin{figure}{section}\counterwithin{table}{section}\counterwithin{equation}{section}
\renewcommand{\thefigure}{\thesection\arabic{figure}}\renewcommand{\theHfigure}{\thesection\arabic{figure}}
\renewcommand{\thetable}{\thesection\arabic{table}}\renewcommand{\theHtable}{\thesection\arabic{table}}
\renewcommand{\theequation}{\thesection\arabic{equation}}\renewcommand{\theHequation}{\thesection\arabic{equation}}

\section{
Implementation Details}
\label{sec:suppl_impl}
\begin{wrapside}{0.6}
\begin{table}[t]
\centering\scriptsize\setlength{\tabcolsep}{1.6pt}
\renewcommand{\arraystretch}{1.08}
\caption{\textbf{Model variants.} Data: training episodes ($845$: the consistent subset); Sim.: current or earlier simulator version, for training and test proxies; Steps: steps\,$\times$\,batch.}
\label{tab:variants}
\tabstyle
\begin{tabular}{l||c|c|c|c|c|c}
\toprule
\textbf{Model} & \textbf{Data} & \textbf{Sim.} & \textbf{Res.} & \textbf{Steps} & \textbf{Tab.} & \textbf{Fig.} \\
\tabdouble
\rowcolor{tabours} SimPWM-H3-C & 845 & cur. & 1280 & $250{\times}32$ & \ref{tab:droid_compare}, \ref{tab:frozen} & \ref{fig:perturb}--\ref{fig:overview}, \ref{fig:qualitative}, \ref{fig:counterfactual} \\
SimPWM-H3 & 3,000 & earl. & 1280 & $250{\times}32$ & \ref{tab:droid_compare}, \ref{tab:frozen} & -- \\
\rowcolor{tabzebra} H3 zero-shot & -- & cur. & 1280 / 640$^*$ & -- & \ref{tab:droid_compare}, \ref{tab:frozen}$^*$ & \ref{fig:counterfactual}$^*$ \\
Ablation models & 845 & cur. & 640 & $250{\times}32$ & \ref{tab:render}, \ref{tab:frozen} & \ref{fig:render}, \ref{fig:counterfactual} \\
\rowcolor{tabzebra} SimPWM-LTX-2.3-C & 845 & cur. & 1280 & $1{,}000{\times}16$ & \ref{tab:droid_compare}, \ref{tab:frozen} & \ref{fig:perturb}, \ref{fig:counterfactual} \\
SimPWM-LTX-2.3 & 3,000 & earl. & 1280 & $2{,}000{\times}16$ & \ref{tab:droid_compare} & \ref{fig:qualitative} \\
\rowcolor{tabzebra} SimPWM-Cosmos-Transfer & 3,000 & earl. & 1280 & $2{,}000{\times}16$ & \ref{tab:droid_compare}, \ref{tab:frozen} & \ref{fig:qualitative} \\
\bottomrule
\end{tabular}
\par\vspace{2pt}{\scriptsize Res.: $1280{\times}704$ or $640{\times}352$.}
\end{table}

\end{wrapside}
\paragraph{Camera calibration.}
\label{sec:suppl_cam}
The camera poses logged by DROID can be far off (robot-silhouette IoU as low as $0.05$), so we re-calibrate each episode with the robot itself: we render the robot posed by forward kinematics and fit the extrinsics (and optionally the radial distortion) to SAM~3 robot masks on $20$ frames of the recording ($5$ of them inside the evaluated window on test episodes).
A calibration per episode reaches a robot IoU of at least $0.6$ on $86\%$ of $5{,}469$ DROID episodes, against $54\%$ for one calibration per camera setup; episodes below this threshold ($15\%$ of those from the test setups) are left out before the test candidates are drawn.
Refitting each test episode only on frames outside the evaluated window changes the camera by a median of $0.98$\,cm and $0.93^\circ$. With the refitted camera, the robot IoU of the proxy inside the window falls from $0.787$ to $0.768$ (median change $-0.002$; $-0.006$ without the $18$ episodes, $6\%$, in which the refit ends in a local minimum), and the grasp success of the proxy (the version of SimPWM-H3-C) from $69.5\%$ to $68.0\%$ (not significant, $p=0.50$); on the frames outside the window the original camera fits as well ($0.814$ vs.\ $0.806$). The proxy is thus essentially unchanged. The videos follow: regenerated from the refitted proxy (current simulator version for SimPWM-H3-C, earlier one for SimPWM-LTX-2.3), SimPWM-H3-C and SimPWM-LTX-2.3 reach a robot IoU of $0.763$ and $0.764$ instead of $0.777$ (paired differences $-0.013$ for both, $[-0.021,-0.006]$), a PSNR lower by $0.17$ and $0.14$\,dB, a trajectory score lower by $0.023$ ($[-0.044,-0.004]$) and $0.003$ (not significant), and an FVD higher by $2.4$ and $1.0$ (not significant). The drops come from the $18$ episodes with a failed refit: on the other $275$, PSNR changes by $-0.06$ and $-0.04$\,dB and no other metric changes significantly. For reference, the best end-to-end model reaches a robot IoU of $0.690$ in its videos (\hyperref[tab:droid_compare]{Table~\ref*{tab:droid_compare}}). Using test frames still favours robot IoU slightly (Evaluation caveat (iii)); the DROID-logged row of \hyperref[tab:droid]{Table~\ref*{tab:droid}} shows what a poor calibration costs.

\paragraph{Runtime.}
Building the scene, simulating and rendering the proxy take a median of $264$\,s per episode on one A100 (over $376$ episodes, including stereo depth and SAM~3 object extraction); this stage is shared by all base models.
Generating one video then takes $329$\,s for SimPWM-LTX-2.3 and $822$\,s for SimPWM-Cosmos-Transfer on one A100 ($97$ frames at $1280{\times}704$), about $460$\,s per A100 for the ablation models ($158$ frames at $640{\times}352$, run on four A100s), and about $1{,}570$\,s on one A100 for SimPWM-H3 ($158$ frames at $1280{\times}704$).
Cosmos-Predict~2.5 and DreamDojo take about $720$\,s at the same output size, and Ctrl-World ($320{\times}192$ at $5$\,Hz) $50$\,s.

\paragraph{Baselines at native resolution.}
\hyperref[tab:suppl_native]{Table~\ref*{tab:suppl_native}} gives DreamDojo and Cosmos-Predict~2.5 post-trained at their native output with the same data and budget. Resolution changes the image metrics a little but not how closely they follow the motion.
\begin{table*}[t]
\centering\footnotesize\setlength{\tabcolsep}{5pt}
\caption{\textbf{Baselines at native and at our resolution} (DROID test, protocol of \hyperref[tab:droid_compare]{Table~\ref*{tab:droid_compare}}).}
\label{tab:suppl_native}
\renewcommand{\arraystretch}{1.08}\tabstyle
\begin{tabular}{l||ccc|ccc|ccc}
\toprule
\multirow{2}{*}{\textbf{Method}} & \multicolumn{3}{c|}{\textbf{Video quality}} & \multicolumn{3}{c|}{\textbf{Motion}} & \multicolumn{3}{c}{\textbf{Physics}} \\
\cmidrule(lr){2-4}\cmidrule(lr){5-7}\cmidrule(l){8-10}\tabcmidend
 & PSNR$\uparrow$ & LPIPS$\downarrow$ & FVD$\downarrow$ & Robot$\uparrow$ & Traj.$\uparrow$ & Depth$\downarrow$ & Phys-IQ$\uparrow$ & Interact.$\uparrow$ & Deform.\ IoU$\uparrow$ \\
\tabdouble
\tabgroup{10}{DreamDojo} \\
Native ($640{\times}480$) & 19.31 & .156 & 170.3 & .324 & .278 & .304 & 65.2 & .466 & .414 \\
\rowcolor{tabzebra} Ours ($1280{\times}704$) & 19.74 & .142 & 155.5 & .304 & .271 & .320 & 67.8 & .470 & .440 \\
\tabdouble
\tabgroup{10}{Cosmos-Predict~2.5} \\
Native ($320{\times}256$) & 19.80 & .148 & 150.6 & .397 & .370 & .263 & 63.4 & .386 & .412 \\
\rowcolor{tabzebra} Ours ($1280{\times}704$) & 20.65 & .128 & 134.3 & .398 & .384 & .272 & 66.8 & .474 & .404 \\
\bottomrule
\end{tabular}
\end{table*}
\paragraph{Simulator.}
The gripper is \emph{closing} when the commanded closure $c_t\in[0,1]$ exceeds $0.12$, and \emph{opening} when $c_t$ drops by more than $0.12$ below its maximum in the current closure, or below $0.12$.
An object is graspable when at least $k=6$ of its points lie between the pads and its local width does not exceed the opening ($8.5$\,cm); a full closure (median closure $\ge 0.975$) holds nothing. The thresholds were set on development episodes in early versions of the pipeline and have not changed since.
The test set shares no setup with the training or development episodes: of the $500$ candidates, drawn from setups without training episodes among the episodes that pass the calibration check, we keep the $293$ episodes from the $121$ setups that contribute no development episode.
In the earlier simulator version, turning off one option, keeping refined objects in place until they are grasped, lowered grasp success from $74.2\%$ to $70.9\%$ on the $275$ decidable episodes (paired 95\% CI $[1.1, 5.8]$) and the object IoU from $0.30$ to $0.27$; all variants in \hyperref[tab:droid]{Table~\ref*{tab:droid}} use it.

Objects are rendered as z-buffered point splats with gap filling.
\paragraph{Video model.}
The IC-LoRA has rank $32$ on the attention and feed-forward layers and is trained with learning rate $2\times10^{-4}$, linear decay and batch size $16$: SimPWM-LTX-2.3 is stopped at step $2{,}000$ of a $3{,}000$-step schedule, SimPWM-LTX-2.3-C at step $1{,}000$ of a $1{,}500$-step schedule, and the LTX-2.3 robot-only model of \hyperref[sec:suppl_refs]{Sec.~\ref*{sec:suppl_refs}} decays fully over $2{,}000$ steps. The depth proxy is inverse depth over $0.2$--$3$\,m. The prompt is ``A static camera films a Franka robot arm with a two-finger gripper \ldots\ The robot arm performs the task: $\langle$task$\rangle$.''
At inference, the reference video is encoded with the tiled VAE encoder of LTX-2.3 (the training latents are untiled); untiled encoding changes no metric significantly.
\begin{wrapside}{0.45}
\begin{figure}[t]
\centering
\setlength{\tabcolsep}{1.2pt}
\newcommand{\ab}[1]{\rimg[width=\dimexpr(\linewidth-8\tabcolsep-3pt)/5\relax]{fig/ablation/#1}}
\renewcommand{\arraystretch}{0}
\begin{tabular}{@{}ccccc@{}}
\figlab{GT} & \figours{Full} & \figlab{Logged calib.} & \figlab{w/o stereo} & \figlab{w/o snap} \\[3pt]
\ab{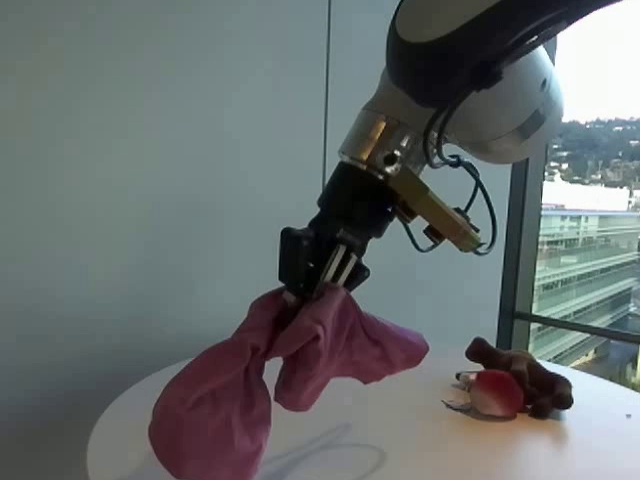} & \ab{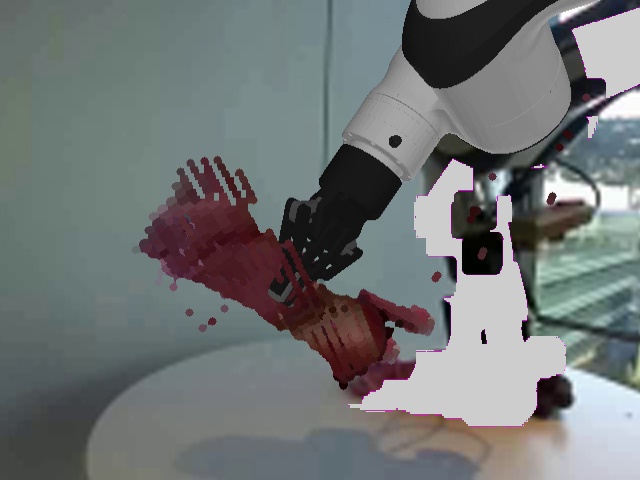} & \ab{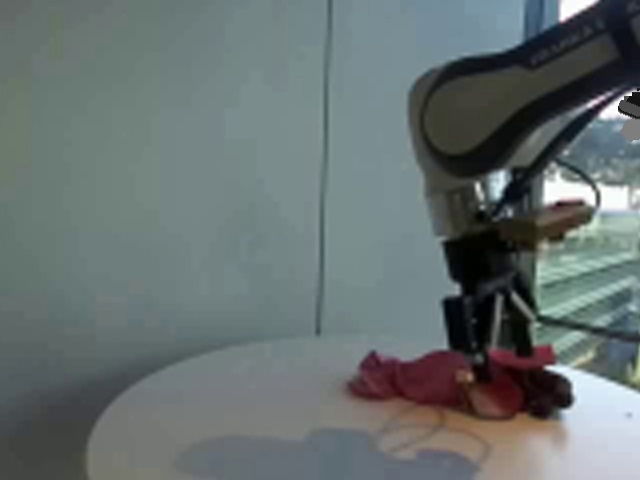} & \ab{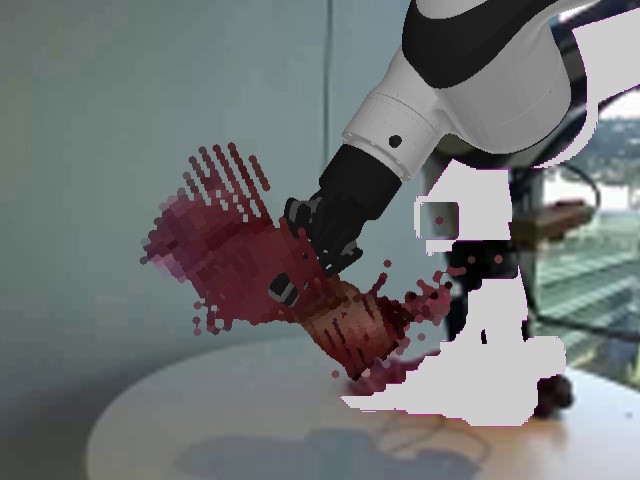} & \ab{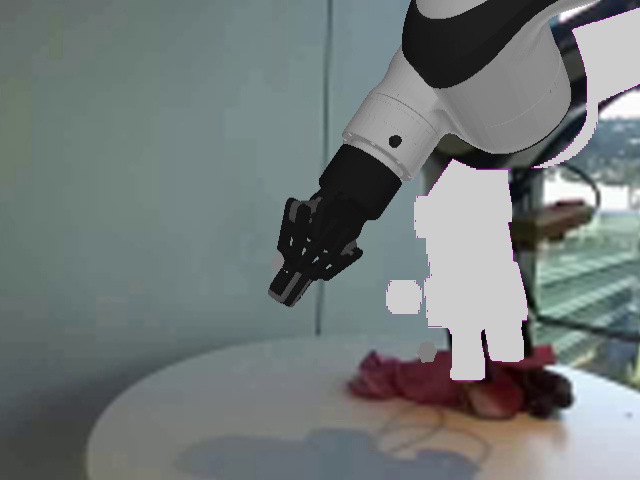} \\[2.5pt]
\ab{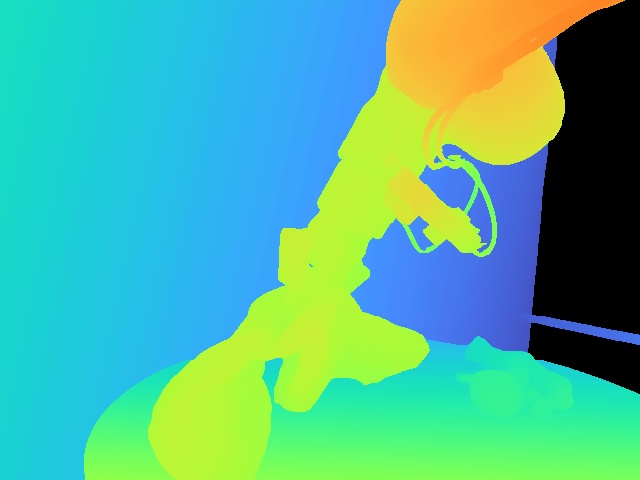} & \ab{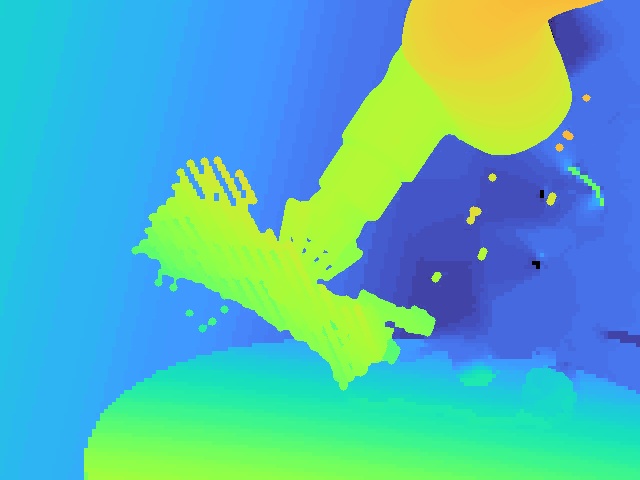} & \ab{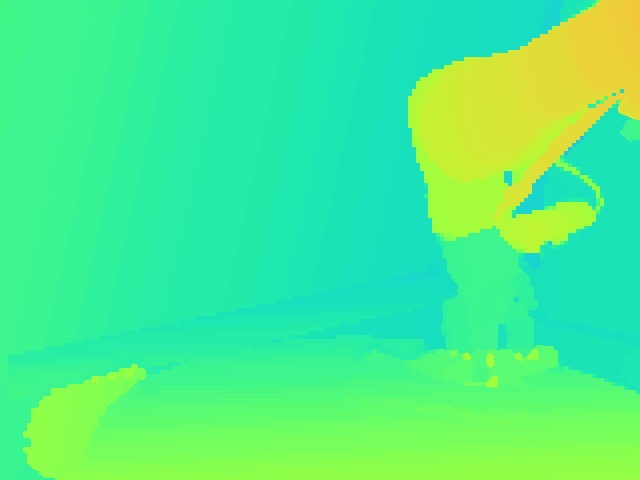} & \ab{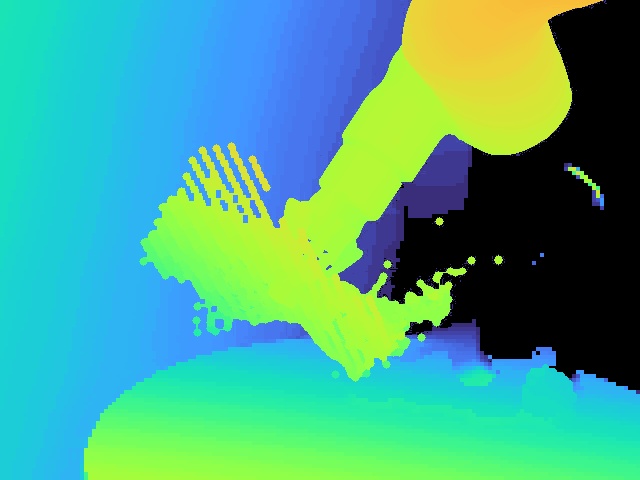} & \ab{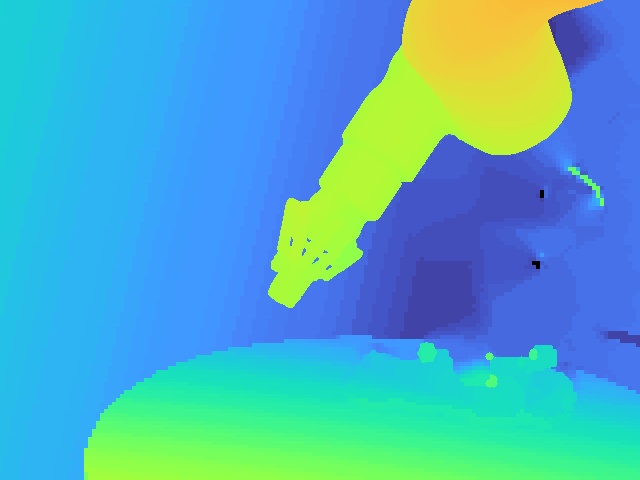} \\
\end{tabular}
\caption{\textbf{Simulator ablations} (\hyperref[tab:droid]{Table~\ref*{tab:droid}}; frames from the earlier simulator version): the DROID-logged calibration and no snap grasp a wrong object and leave the cloth on the table; mono depth misplaces the background.}
\label{fig:ablation}
\end{figure}

\end{wrapside}
\paragraph{Evaluation caveats.}
(i) Ctrl-World is trained on DROID~1.0.1 except every $100$th episode; matching episodes by instruction and length, at least $234$ of our $293$ test episodes ($96$ of the $133$ of \hyperref[fig:counterfactual]{Fig.~\ref*{fig:counterfactual}}) are in its training split.
(ii) SimPWM-H3-C is trained on the $845$ consistent training episodes with a later version of the proxy, which has a lower grasp success on the test episodes ($69.5\%$ vs.\ $74.2\%$; $21$ episodes become wrong and $8$ right, McNemar $p=0.024$) at the same robot IoU and a lower RGB error; among other changes, it grasps only objects inside the jaw volume and segments parts such as lids and handles.
(iii) All other methods are trained on our training episodes and scored on the same test episodes, so data and test selection are common to all rows; only the calibrated camera, fit partly on frames of the test recording, is specific to SimPWM and favours robot IoU.
(iv) The automatic grasp check errs both ways: a wrongly extracted real target rejects a right grasp, and a grasp of a neighbouring object can be accepted. For the earlier simulator version ($74.2\%$), an audit of $33$ of its $71$ rejections found the real target extracted wrongly in $16$, which would raise its grasp success to $80.0\%$; unaudited rejections stay wrong.
(v) With $293$ clips, fewer than the $400$ feature dimensions, FVD is biased upwards; only differences between methods on the same clips are meaningful.
\paragraph{Evaluation.}
DreamDojo and Cosmos-Predict~2.5 are post-trained like SimPWM-LTX-2.3 ($1280{\times}704$, $97$ frames, first frame and task text, $2{,}000$ steps, global batch $16$). Non-rigid episodes are those whose task instruction, without parenthesised examples, contains a deformable-object keyword (towel, cloth, fold, napkin, shirt, clothes, sweater, sweatshirt, fabric, rag, tissue, sock, jacket, scarf, blanket, curtain, cable, cord, wire, rope, string, garment, trousers, serviette, bandana, sheet); this rule is used for \hyperref[tab:frozen]{Table~\ref*{tab:frozen}} and all rigid / deformable counts. It counts bags and objects not in the list (\eg a hoodie) as rigid; $13$ of the $96$ rigid targets among the $133$ episodes involve a bag.
Without the action snap, grasp success drops by $1.9$ points ($95\%$ interval $[-2.5, 6.5]$, not significant) and object IoU by $0.102$ ($[0.063, 0.142]$); in the earlier simulator version, by $4.0$ points ($[0.4, 8.0]$). Per-row $95\%$ bootstrap intervals of \hyperref[tab:droid]{Table~\ref*{tab:droid}} are about $\pm.015$ robot IoU, $\pm.04$ object IoU, $\pm5$ grasp success, $\pm.005$ RGB L1 and $\pm.025$ depth MSE. RGB L1 and depth MSE are computed on one pixel set per episode shared by all rows; on the real moving pixels alone, our RGB L1 is level with that of PointWorld ($0.179$ vs.\ $0.180$) and the depth medians are similar ($0.122$ vs.\ $0.108$); the gap in the mean depth MSE comes mostly from a long tail of our failed simulations.
FVD uses $16$ frames per clip, resized to $224{\times}224$, I3D features ($400$-D) and the real clips of the same episodes as reference; with $20$-group jackknife $95\%$ intervals, the paired differences of SimPWM-H3-C in \hyperref[tab:droid_compare]{Table~\ref*{tab:droid_compare}} are $-6.05$ $[-15.07, 2.98]$ to SimPWM-LTX-2.3 and $-20.3$ $[-31.4, -9.1]$ to Ctrl-World. Because of its sample bias (caveat (v)), we do not compare FVD on the $27$ episodes of \hyperref[sec:suppl_ar2s]{Sec.~\ref*{sec:suppl_ar2s}}, and the FVD of the $100$ GR1 windows is only indicative.
The depth MSE (m$^2$) of \hyperref[tab:droid_compare]{Tables~\ref*{tab:droid_compare}} and~\ref{tab:render} compares Depth Anything~V2 (Metric-Indoor-Large) on the generated frames, scaled per step to the stereo depth on static pixels, with the FoundationStereo depth of the real video at steps $24$, $48$, $72$ and $96$, on the pixels in $0.2$--$3$\,m that change by more than $2$\,cm.
For Physics-IQ, a re-encoded real video scores $97.4$ (on $58$ test episodes), so the metric is not saturated, and the RGB proxy alone scores $69.2$; the gap of SimPWM-LTX-2.3 to Ctrl-World comes from flicker in static regions of its output.

\section{Stereo Depth}
\label{sec:suppl_stereo}
\begin{figure*}[t]
\centering
\includegraphics[width=0.92\textwidth]{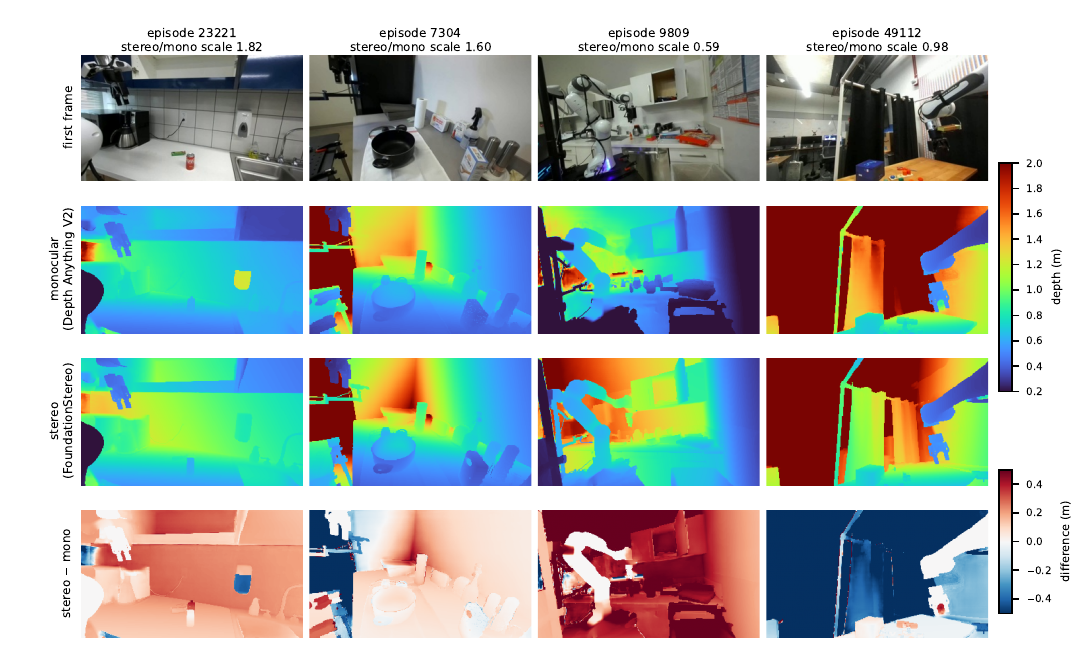}
\caption{\textbf{First-frame depth, monocular vs.\ stereo} (development episodes). Rows: first frame, monocular depth (robot as scale ruler), stereo depth, difference; titles: median stereo/mono ratio. Chosen to show large disagreements (first three) and a ratio near $1$ (last).}
\label{fig:suppl_depth}
\end{figure*}

Our DROID experiments take the first-frame depth from the stereo pair of the external camera (\hyperref[sec:method_init]{Sec.~\ref*{sec:method_init}}).
\hyperref[fig:suppl_depth]{Figure~\ref*{fig:suppl_depth}} compares it with the monocular depth on four development episodes: the monocular depth is plausible in shape, but its scale is often off by tens of centimetres, which the stereo depth corrects; the \emph{w/o stereo depth} row of \hyperref[tab:droid]{Table~\ref*{tab:droid}} measures the effect on the proxy.
Without a stereo pair, Depth Anything~V2~\cite{yang2024depthv2} is corrected for the camera's focal length and scaled by one global factor, clipped to $[0.6,1.6]$, so that the robot pixels match the depth of the rendered robot.
On the first frames of the $500$ candidate test episodes, this depth has an absolute relative error of $0.296$ against stereo, and $0.173$ with the best scale per image, so most of the error is the global scale.

\section{Additional DROID Results}
\label{sec:suppl_droid}

\subsection{More Results}
\label{sec:suppl_more}
\begin{figure*}[t]
\centering
\setlength{\tabcolsep}{1pt}
\newlength{\galsw}\setlength{\galsw}{\dimexpr(384.06pt-6\tabcolsep)/4\relax}
\newcommand{\galsimg}[1]{\includegraphics[width=\galsw]{fig/gallery/#1}}
\newcommand{\galsrow}[1]{\galsimg{#1_gt.jpg} & \galsimg{#1_rgb.jpg} & \galsimg{#1_depth.jpg} & \galsimg{#1_gen.jpg}}
\renewcommand{\arraystretch}{0}
\hspace*{-4.44pt}\begin{tabular}{@{}cccc@{}}
{\small GT} & {\small RGB proxy} & {\small Depth proxy} & {\small\textbf{Generated}} \\[3pt]
\galsrow{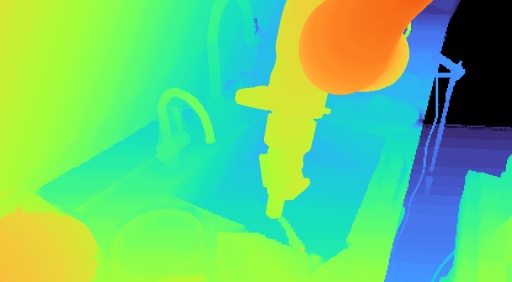} \\[2pt]
\galsrow{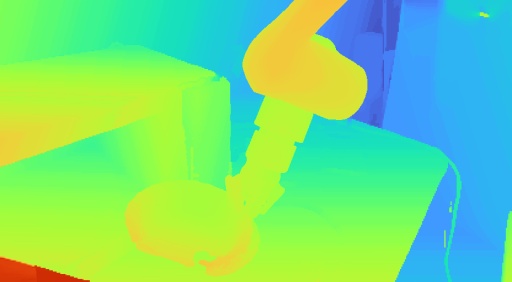} \\[2pt]
\galsrow{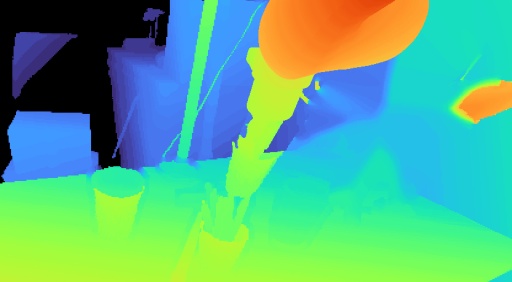} \\[2pt]
\galsrow{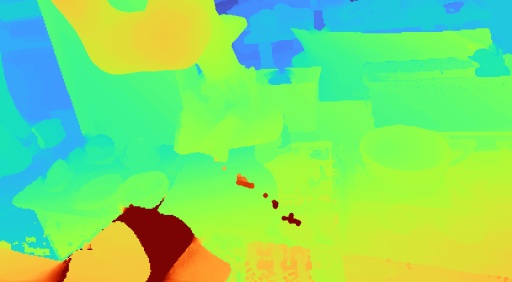} \\[2pt]
\galsrow{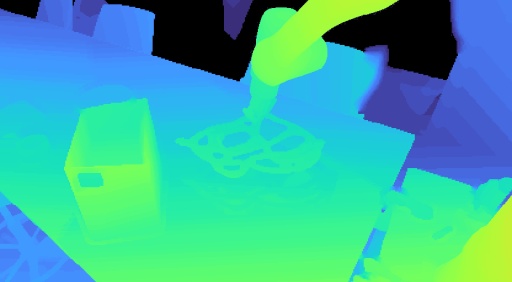} \\[2pt]
\galsrow{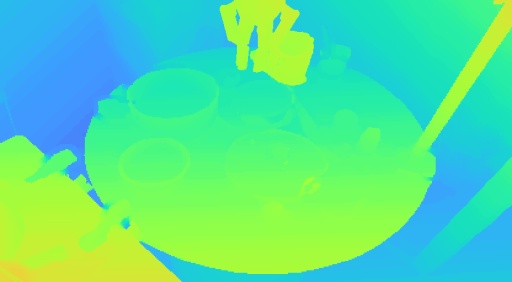} \\
\end{tabular}
\caption{\textbf{More results of SimPWM-H3-C} on test episodes from unseen setups (magenta: unseen; depth proxy: inverse depth over $0.2$--$3$\,m, blue near, red far).}
\label{fig:suppl_gallery}
\end{figure*}

\hyperref[fig:suppl_gallery]{Figure~\ref*{fig:suppl_gallery}} shows six more test episodes in the format of \hyperref[fig:gallery]{Fig.~\ref*{fig:gallery}}.

\begin{figure*}[t]
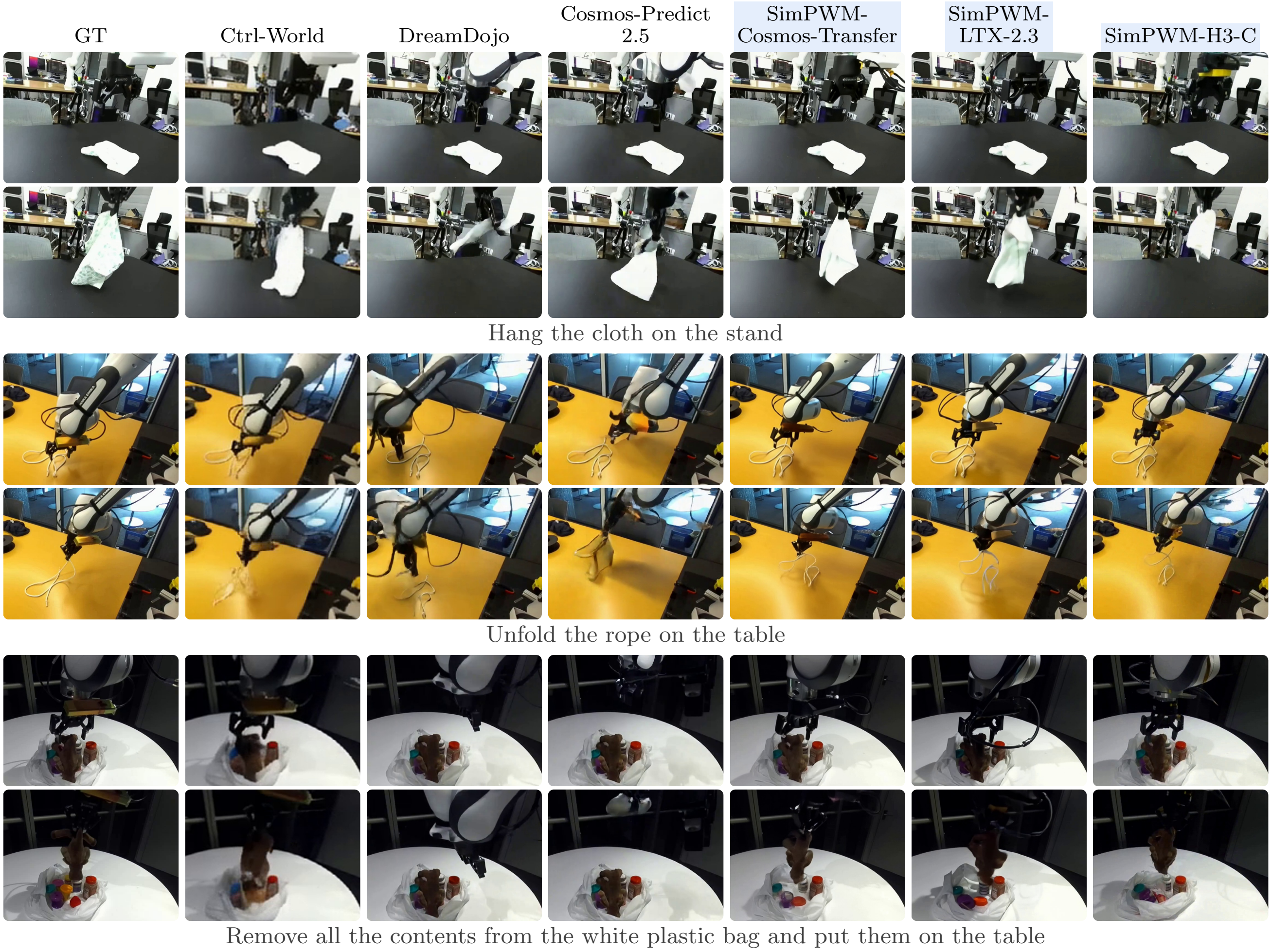

\centering
\setlength{\tabcolsep}{1.2pt}
\newlength{\sqw}\setlength{\sqw}{\dimexpr(\textwidth-12\tabcolsep-11pt)/7\relax}
\newcommand{\sqa}[1]{\rimg[width=\sqw]{fig/qualitative/#1_crop40.jpg}}
\newcommand{\sqb}[1]{\rimg[width=\sqw]{fig/qualitative/#1_crop.jpg}}
\renewcommand{\figlab}[1]{\makebox[0pt]{\scriptsize\renewcommand{\arraystretch}{0.9}\begin{tabular}[b]{@{}c@{}}#1\end{tabular}}}%
\renewcommand{\figours}[1]{\makebox[0pt]{{\setlength{\fboxsep}{1.2pt}\colorbox{tabours}{\scriptsize\renewcommand{\arraystretch}{0.9}\begin{tabular}[b]{@{}c@{}}#1\end{tabular}}}}}%
\newcommand{\sqrow}[2]{#1{#2_gt} & #1{#2_cw} & #1{#2_dd} & #1{#2_cp} & #1{#2_wct} & #1{#2_wpm} & #1{#2_h3}}
\renewcommand{\arraystretch}{0}
\begin{tabular}{@{}ccccccc@{}}
\figlab{GT} & \figlab{Ctrl-World} & \figlab{DreamDojo} & \figlab{Cosmos-Predict\\2.5} & \figours{SimPWM-\\Cosmos-Transfer} & \figours{SimPWM-\\LTX-2.3} & \figours{SimPWM-H3-C} \\[3pt]
\sqrow{\sqa}{8231} \\[1.2pt]
\sqrow{\sqb}{8231} \\[2.5pt]
\multicolumn{7}{c}{\figdesc{Hang the cloth on the stand}} \\[5pt]
\sqrow{\sqa}{4984} \\[1.2pt]
\sqrow{\sqb}{4984} \\[2.5pt]
\multicolumn{7}{c}{\figdesc{Unfold the rope on the table}} \\[5pt]
\sqrow{\sqa}{1956} \\[1.2pt]
\sqrow{\sqb}{1956} \\[2.5pt]
\multicolumn{7}{c}{\figdesc{Remove all the contents from the white plastic bag and put them on the table}} \\
\end{tabular}
\caption{\textbf{Qualitative comparison at two steps}: the two episodes of \hyperref[fig:qualitative]{Fig.~\ref*{fig:qualitative}} and a third one, top row of each pair while the gripper reaches and closes, bottom row at the frame shown in \hyperref[fig:qualitative]{Fig.~\ref*{fig:qualitative}}.}
\label{fig:suppl_qualitative}
\end{figure*}

\hyperref[fig:suppl_qualitative]{Figure~\ref*{fig:suppl_qualitative}} shows the two episodes of \hyperref[fig:qualitative]{Fig.~\ref*{fig:qualitative}} and a third one at two steps of each episode: step $40$, while the gripper reaches and closes, and the step of \hyperref[fig:qualitative]{Fig.~\ref*{fig:qualitative}}.

\subsection{Proxies of the Sim-to-Real Ablation}
\label{sec:suppl_refs}
\paragraph{Robot-only proxy with LTX-2.3.}
An LTX-2.3 model trained with the robot-only proxy shares the data ($3{,}000$ episodes, earlier simulator version), steps, batch size and sampling settings of SimPWM-LTX-2.3; only the learning-rate schedule differs (\hyperref[sec:suppl_impl]{Sec.~\ref*{sec:suppl_impl}}).
On the $293$ test episodes, scored at $1280{\times}704$, the robot-only model is not worse in image quality (robot only vs.\ RGB proxy: PSNR $23.11$ vs.\ $22.91$, LPIPS $0.123$ vs.\ $0.128$, FVD $80.7$ vs.\ $93.3$, difference $-12.6$ $[-19.2, -6.0]$; robot IoU $0.794$ vs.\ $0.788$; Phys-IQ $66.0$ vs.\ $63.4$).
With the grasp judge of \hyperref[tab:render]{Table~\ref*{tab:render}}, grasp agreement is the same ($72.7\%$ vs.\ $73.7\%$); on the $133$ real pick-ups the RGB proxy misses fewer ($12.8\%$ vs.\ $20.3\%$, $[1.5, 14.3]$) but picks up another object more often ($23.3\%$ vs.\ $15.0\%$, $[0.8, 16.5]$), and picks up the right object equally often ($63.9\%$ vs.\ $64.7\%$); with the gripper never closing, both lift in $64\%$ of $100$ earlier test episodes (difference $0$, $[-8, 8]$).
LTX-2.3 thus uses the moved objects of the proxy to decide that something is lifted, but not which object, in line with \hyperref[sec:evaluator]{Sec.~\ref*{sec:evaluator}}.

\begin{wrapside}{0.5}
\begin{table}[h]
\centering
\footnotesize
\setlength{\tabcolsep}{3pt}
\caption{\textbf{Grasp outcome of the ablation models by proxy grasp} on the $133$ real pick-ups (\%; right / wrong object / missed), robot-only vs.\ RGB proxy; $10$ episodes with an undecided proxy grasp and $1$ labelled separately as a missed grasp not shown.}
\label{tab:grasp_strata}
\begin{tabular}{lccc}
\toprule
Proxy grasp & $n$ & Robot only & RGB proxy \\
\midrule
Right object & $69$ & $46$ / $6$ / $48$ & $57$ / $20$ / $23$ \\
Other object & $31$ & $23$ / $26$ / $52$ & $35$ / $26$ / $39$ \\
None & $22$ & $23$ / $5$ / $73$ & $14$ / $9$ / $77$ \\
\bottomrule
\end{tabular}
\end{table}
\end{wrapside}
\paragraph{Grasp columns of \hyperref[tab:render]{Table~\ref*{tab:render}}.}
Against the robot alone, the RGB proxy reduces missed pick-ups by $16.5$ points ($95\%$ interval $[8.3, 25.6]$) without significantly more false pick-ups ($+1.9$, $[-3.1, 6.9]$); $19$ of its $20$ false pick-ups occur where the simulator itself grasps, so they come from the proxy.
Only about half of the added pick-ups are of the right object ($+7.5$ points, $[-0.8, 15.8]$; another object: $+9.0$, $[3.0, 15.8]$); in $14$ of the $25$ wrong-object pick-ups the simulator grasps the right object (another object in $8$, none in $2$, undecided in $1$), so they come from the video or from the judge's matching of object names. The RGB and depth proxies tie in grasp agreement.
\hyperref[tab:grasp_strata]{Table~\ref*{tab:grasp_strata}} splits the $133$ real pick-ups by the proxy grasp of the ablation models (the later simulator version): where the proxy grasps the right object, the RGB proxy adds lifts over the robot alone ($+24.6$ points, $[13.0, 36.2]$), about half of them of another object; where it grasps nothing, it adds none ($-4.5$, $[-22.7, 9.1]$). The video thus lifts where the simulator grasps.

\begin{figure*}[t]
\centering
\setlength{\tabcolsep}{0.8pt}
\newlength{\rfw}\setlength{\rfw}{\dimexpr(0.92\textwidth-8\tabcolsep)/5\relax}
\newcommand{\rfimg}[1]{\includegraphics[width=\rfw]{fig/render/13218_#1.jpg}}
\renewcommand{\arraystretch}{0}
\begin{tabular}{@{}ccccc@{}}
{\scriptsize First frame} & {\scriptsize Robot only} & {\scriptsize PointWorld} & {\scriptsize RGB proxy} & {\scriptsize Depth proxy} \\[3pt]
\rfimg{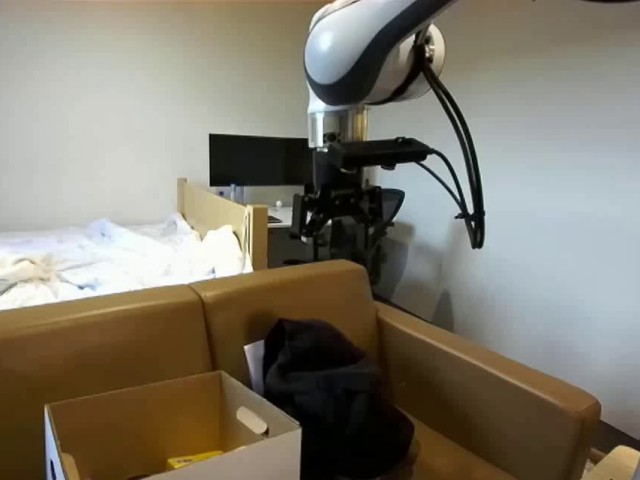} & \rfimg{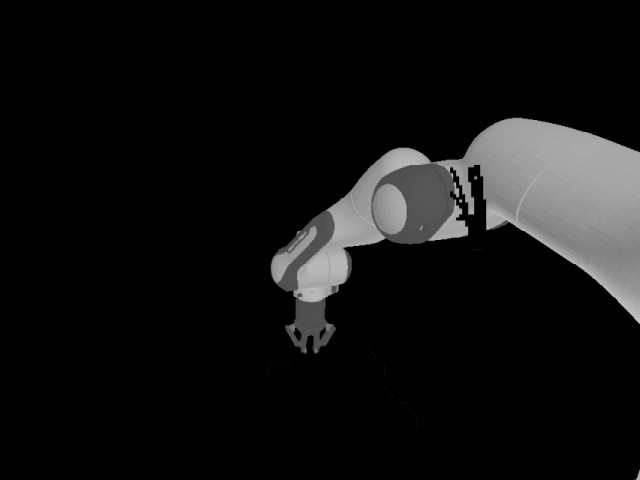} & \rfimg{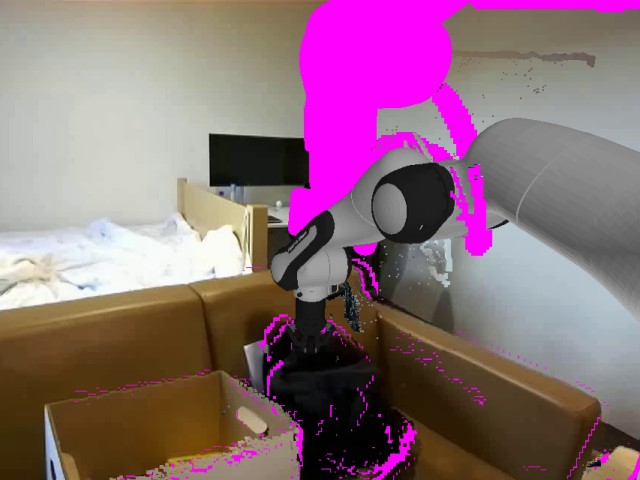} & \rfimg{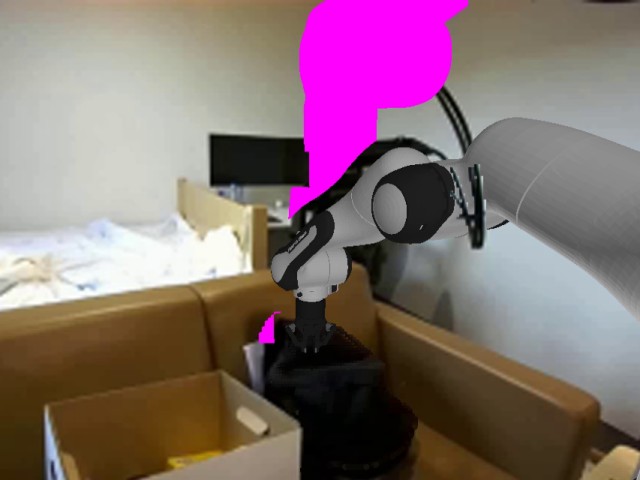} & \rfimg{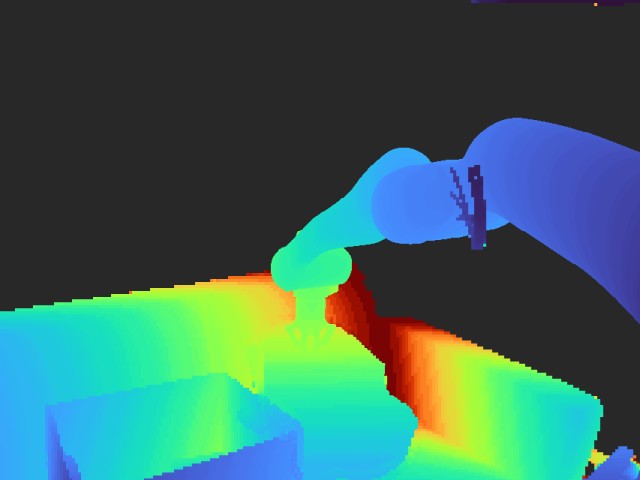} \\
\end{tabular}
\caption{\textbf{Reference videos} of the variants of \hyperref[tab:render]{Table~\ref*{tab:render}} for the first episode of \hyperref[fig:render]{Fig.~\ref*{fig:render}} (depth proxy: inverse depth over $0.2$--$3$\,m, blue near, red far), shown in the proxy version of LTX-2.3; MiniMax-H3 receives the same views rendered by the later version.}
\label{fig:suppl_refs}
\end{figure*}

\hyperref[fig:suppl_refs]{Figure~\ref*{fig:suppl_refs}} shows what the video model receives in each row of \hyperref[tab:render]{Table~\ref*{tab:render}}.
\emph{Robot only} is the shaded robot without the scene; the \emph{RGB} and \emph{depth proxies} render the scene moved by our simulator.
PointWorld~\cite{huang2026pointworld} predicts from the same first frame, stereo depth and actions how the scene points move (chained over five clips); every pixel of the first frame follows its nearest predicted point, drawn in the format of our RGB proxy with our robot rendering.
The two RGB proxies thus differ only in how the scene moves.

\subsection{Physical Plausibility by Pairwise Comparison}
\label{sec:suppl_physics}
The pairwise judge of \hyperref[sec:exp_render]{Sec.~\ref*{sec:exp_render}} compares how a held cloth or rope deforms.
Of the $61$ test episodes with a non-rigid target, $40$ have a window in which the real gripper holds the target, and $24$ of these have a real target that the strict extraction of the deformation IoU can track.
For each of these $24$, we crop a $4{:}3$ window around the real gripper and the target, the same for all methods, and take eight frames evenly from the holding window.
A blind Claude judge (Claude Opus~5.5; batches of $30$ sheets in random order, no method names) sees the real sequence and two candidates and answers which one shows the held object deforming and hanging more like the real video.
Each pair is asked in both orders and counts as a win only if both answers agree, otherwise as a tie.

\begin{sidecapfloat}[t]{0.92}
\begin{table}[t]
\centering\footnotesize\setlength{\tabcolsep}{4pt}\renewcommand{\arraystretch}{1.08}
\captionof{table}{\textbf{Pairwise plausibility} (number of the $24$ episodes). H3: the ablation model ($640{\times}352$, RGB proxy); LTX, Transfer: SimPWM-LTX-2.3 and SimPWM-Cosmos-Transfer; H3 judged in separate batches.}
\label{tab:suppl_physics}
\tabstyle
\begin{tabular}{l||ccc@{\hspace{18pt}}l||ccc}
\toprule
\textbf{Pair} & \textbf{Win} & \textbf{Tie} & \textbf{Loss} & \textbf{Pair} & \textbf{Win} & \textbf{Tie} & \textbf{Loss} \\
\tabdouble
H3 vs.\ RGB proxy & 17 & 4 & 3 & H3 vs.\ DreamDojo & 20 & 2 & 2 \\
\rowcolor{tabzebra} LTX vs.\ RGB proxy & 18 & 1 & 5 & LTX vs.\ DreamDojo & 21 & 1 & 2 \\
Transfer vs.\ RGB proxy & 20 & 1 & 3 & H3 vs.\ Cosmos-Predict~2.5 & 21 & 1 & 2 \\
\rowcolor{tabzebra} H3 vs.\ Ctrl-World & 10 & 4 & 10 & LTX vs.\ Cosmos-Predict~2.5 & 20 & 2 & 2 \\
LTX vs.\ Ctrl-World & 11 & 6 & 7 &  & & &  \\
\bottomrule
\end{tabular}
\end{table}
\end{sidecapfloat}
On all $40$ episodes with a holding window, the win rates against the proxy, DreamDojo and Cosmos-Predict~2.5 are $70$--$90\%$, and against Ctrl-World $35\%$ for the H3 model and $52\%$ for SimPWM-LTX-2.3. H3 and LTX-2.3 were judged in separate runs and never against each other.
Two caveats: against the proxy, the judge may prefer the real-looking rendering regardless of how the cloth deforms; and against DreamDojo and Cosmos-Predict~2.5, part of the wins come from videos in which the object is not lifted at all.

\paragraph{VideoPhy-2.} The physical-commonsense judge of VideoPhy-2~\cite{bansal2025videophy2} (official model and prompt) does not separate the methods: on the same $24$ episodes all score $3.71$--$3.96$ out of $5$, and the real videos $3.96$. It does not mark the fragmented proxy cloth as implausible either, so a general physics score misses the difference that the pairwise comparison measures.

\subsection{Action Perturbation with LTX-2.3 and MiniMax-H3}
\label{sec:suppl_h3}
\begin{figure*}[t]
\centering
\setlength{\tabcolsep}{1.2pt}
\newlength{\hw}\setlength{\hw}{\dimexpr(\textwidth-16\tabcolsep-2\baselineskip-14pt)/8\relax}
\newcommand{\himg}[1]{\rimg[width=\hw]{fig/h3/#1}}
\newcommand{\hrow}[1]{\himg{#1_32.jpg} & \himg{#1_56.jpg} & \himg{#1_72.jpg} & \himg{#1_96.jpg}}
\newcommand{\hlab}[1]{\rotatebox{90}{\makebox[0.55\hw]{\figlab{#1}}}}
\renewcommand{\arraystretch}{0}
\begin{tabular}{@{}c@{\hspace{2pt}}cccc@{\hspace{10pt}}c@{\hspace{2pt}}cccc@{}}
\multicolumn{5}{c}{\figlab{Recorded actions; the proxy misses the rope}} & \multicolumn{5}{c}{\figlab{No close: the gripper never closes}} \\[3pt]
\hlab{Real} & \hrow{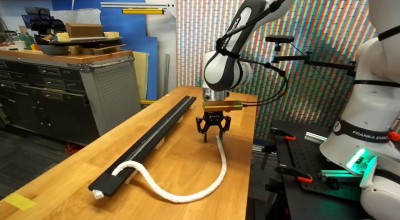} & \hlab{Proxy} & \hrow{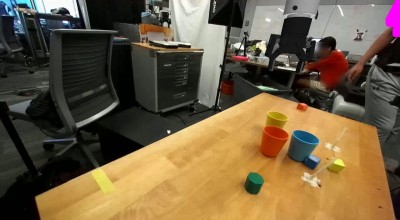} \\[2pt]
\hlab{Proxy} & \hrow{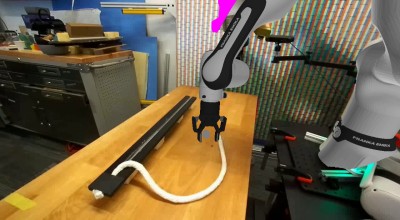} & \hlab{LTX-2.3} & \hrow{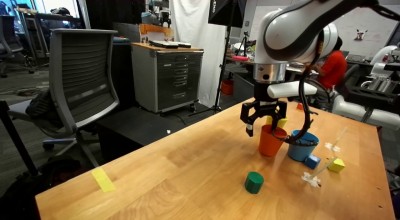} \\[2pt]
\hlab{LTX-2.3} & \hrow{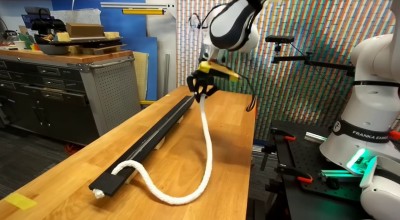} & \hlab{H3} & \hrow{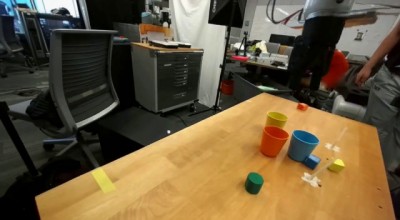} \\[2pt]
\hlab{H3} & \hrow{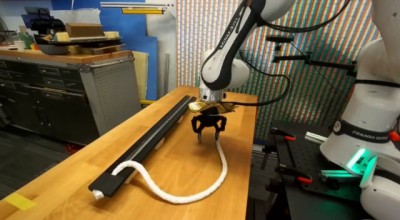} & & \multicolumn{4}{c}{\raisebox{0.45\hw}{\parbox[t]{4\hw}{\centering\figdesc{Pick up the orange cup, place it into the yellow cup, and finally place both into the blue cup}}}} \\[2.5pt]
& \multicolumn{4}{c}{\figdesc{Pick up the white rope and put it in between the black object}} & & \multicolumn{4}{c}{} \\
\end{tabular}
\caption{\textbf{SimPWM-LTX-2.3-C and the ablation model} ($640{\times}352$), both fine-tuned on the consistent subset (\hyperref[sec:suppl_h3]{Sec.~\ref*{sec:suppl_h3}}). Left: the proxy does not lift the rope; LTX-2.3 lifts it as the real robot does, the ablation model leaves it on the table as the proxy does. Right: with the gripper held open, LTX-2.3 still lifts the orange cup; the ablation model leaves it in place.}
\label{fig:suppl_h3}
\end{figure*}

\hyperref[fig:counterfactual]{Figure~\ref*{fig:counterfactual}} uses the $133$ test episodes in which the real robot lifts an object according to the grasp judge of \hyperref[tab:render]{Table~\ref*{tab:render}}; for $62$ of them no target object meets our strict definition (the object held when the gripper stalls), and the judged frames start at the first stall of the gripper instead.
The perturbation judge (Claude Opus~5.5) sees the first frame and eight frames from that point to the end of the clip, in temporal order, with the videos of all methods shuffled and blind to the method, with the target object named from the task; rates include low-confidence answers. To check the judge, one author labelled $60$ of its frame sheets, drawn at random from SimPWM-H3-C and SimPWM-LTX-2.3-C ($10$ per model and condition), blind to model, condition and verdict, as yes, no or uncertain (sheets on which the author and the judge disagreed were labelled a second time, again blind). On the $46$ sheets labelled yes or no, the judge agrees on $44$ ($96\%$, Cohen's $\kappa=0.91$; first pass $41$ of $50$, $\kappa=0.65$), and under perturbed actions the author counts as many lifts as the judge: $4$ of $20$ for SimPWM-H3-C and $9$ of $20$ for SimPWM-LTX-2.3-C. $14$ of the $60$ sheets are uncertain, $5$ of the $10$ frozen-object sheets of SimPWM-H3-C.
\hyperref[tab:suppl_cf_sets]{Table~\ref*{tab:suppl_cf_sets}} lists the rates of all variants.

\begin{wrapside}{0.44}
\begin{table}[t]
\centering\footnotesize\setlength{\tabcolsep}{3.2pt}
\caption{\textbf{Action-perturbation rates per variant} (Claude judge, \%; recorded actions / no close / frozen objects; $133$ episodes of \hyperref[fig:counterfactual]{Fig.~\ref*{fig:counterfactual}}); --: not applicable.}
\label{tab:suppl_cf_sets}
\renewcommand{\arraystretch}{1.05}\tabstyle
\begin{tabular}{l||c}
\toprule
\textbf{Variant} & \textbf{Rec. / NC / Fz.} \\
\tabdouble
Real video & 68 / -- / -- \\
\rowcolor{tabzebra} Ctrl-World & 62 / 7 / -- \\
DreamDojo & 8 / 10 / -- \\
\rowcolor{tabzebra} Cosmos-Predict~2.5 & 31 / 11 / -- \\
SimPWM-H3-C & 47 / 16 / 18 \\
\rowcolor{tabzebra} Ablation model ($640{\times}352$) & 47 / 8 / 16 \\
\hspace{0.6em}robot-only proxy & 35 / 14 / 22 \\
SimPWM-H3 & 51 / 32 / 35 \\
\rowcolor{tabzebra} SimPWM-LTX-2.3-C & 62 / 44 / 55 \\
Proxy readout & 68 / 0 / 0 \\
\tabdouble
\rowcolor{tabzebra} MiniMax-H3 zero-shot ($640{\times}352$) & 49 / 18 / 19 \\
\hspace{0.6em}robot-only proxy & 55 / -- / 35 \\
SimPWM-Cosmos-Transfer & 59 / 42 / 47 \\
\bottomrule
\end{tabular}
\end{table}
\end{wrapside}
\emph{Arm offset} shifts the arm by $5$\,cm so that the gripper closes beside the object. The simulator trusts the gripper signal, which the offset leaves unchanged: it still lifts an object in $75$ of the $133$ episodes, because the object is wider than the offset ($33$), the action snap pulls it into the gripper ($29$), or a segment is added where the gripper closes ($13$). The offset thus does not define whether the object should be lifted, and we report it only here: on the $133$ episodes, the video shows the lift in $34\%$ (SimPWM-H3-C), $37\%$ and $35\%$ (ablation models with the RGB and the robot-only proxy), $42\%$ (SimPWM-H3), $61\%$ (SimPWM-LTX-2.3-C), $55\%$ (Ctrl-World) and $13\%$ (DreamDojo) of the videos.

\paragraph{No-close lifts of SimPWM-H3-C.}
Under no close, SimPWM-H3-C lifts the object $8$ points more often than the ablation model on the $133$ episodes ($95\%$ interval $[1, 15]$). Of its $21$ no-close lifts, $10$ are of deformable targets (task-instruction keywords, as in \hyperref[tab:frozen]{Table~\ref*{tab:frozen}}), and $17$ are also shown by SimPWM-LTX-2.3-C.

\paragraph{Zero-shot and full-data H3 models.}
Without fine-tuning, freezing the objects lowers the lift rate of MiniMax-H3 by $10$ points more with the RGB proxy than with the robot-only proxy ($[1, 19]$; \hyperref[tab:frozen]{Tables~\ref*{tab:frozen}} and~\ref{tab:suppl_cf_sets}); fine-tuning lowers its no-close rate by $10$ points ($[3, 17]$).
SimPWM-H3, trained on all $3{,}000$ episodes with the earlier simulator version, lifts the object $16$ ($[9, 23]$) and $17$ ($[11, 23]$) points more often than SimPWM-H3-C under no close and with frozen objects, at the same image quality; in $763$ of its training episodes the consistency check (with the later simulator version) finds that the simulated grasp disagrees with the real one.
All H3 models are trained for $250$ steps; trained on to step $500$, SimPWM-H3 improves in quality (FVD $56.9\to52.9$ on the $500$ candidates) and lifts less often under no close.

\paragraph{Where the simulator grasps nothing.}
The video follows the readout where the simulator grasps nothing: on the $22$ real pick-ups where the proxy grasps no object (row None of \hyperref[tab:grasp_strata]{Table~\ref*{tab:grasp_strata}}), SimPWM-H3-C misses $14$ ($64\%$, $[45, 82]$), SimPWM-LTX-2.3-C only $4$ ($18\%$, $[5, 36]$), inferring the lift from the arm motion. This subset differs from the $43$ episodes of \hyperref[sec:evaluator]{Sec.~\ref*{sec:evaluator}} in which the simulator does not lift the target, which also include grasps of another object.

\paragraph{Token-level loss of LTX-2.3.}
We compute the denoising loss of SimPWM-LTX-2.3 on $30$ training episodes at noise level $\sigma=0.9$, split by token (background, static objects, the arm, and objects that move in the real video), with the reference video set to the proxy, to the proxy with frozen objects, to the proxy of another episode, or left out.
On the moved objects the loss is $0.319$ with the proxy, $0.325$ with frozen objects, $0.385$ with none and $0.396$ with a shuffled proxy: the arm motion accounts for $91\%$ of what the proxy lowers the loss on the moved objects, so LTX-2.3 predicts the moved objects mainly from the arm.
Moved objects cover only $0.36\%$ of the reference-video tokens of the $3{,}000$ training episodes, the arm $16\%$.
This describes what LTX-2.3 learns, not why: MiniMax-H3, fine-tuned on the same proxies, follows the object state.

\paragraph{MiniMax-H3 fine-tuning.}
We fine-tune MiniMax-H3~\cite{minimax2026h3}, a $33$B video model whose Ref2VA variant takes a reference video, a reference image and text: the proxy is the reference video, the real first frame the reference image and the first latent frame of the target, and the text asks the model to follow the arm, gripper and objects of the reference video, followed by the task.
The consistent subset keeps the $845$ training episodes whose proxies agree with the real video (the proxy grasp agrees with the real one, robot IoU at least $0.8$, held-object IoU at least $0.5$, the $20\%$ with the largest RGB error removed).
H3 is trained with a LoRA of rank $32$ for $250$ steps at batch size $32$ ($24$\,fps, $158$ frames resampled from the $97$ recorded steps), and SimPWM-LTX-2.3-C with the IC-LoRA for $1{,}000$ steps at batch size $16$ (SimPWM-LTX-2.3: $2{,}000$), so H3 sees half as many samples ($8{,}000$ vs.\ $16{,}000$).
In episode 1333 (\hyperref[fig:suppl_h3]{Fig.~\ref*{fig:suppl_h3}}) the simulator does not lift the rope although the real robot does, and LTX-2.3 lifts it, following the arm, while the ablation model leaves it on the table, following the proxy.
Six of the $133$ episodes were looked at during development; without them, SimPWM-H3-C gives $46/16/33/18$ (recorded / no close / offset / frozen objects) and the ablation model $46/9/35/17$.

\subsection{Comparison with Agentic Real2Sim}
\label{sec:suppl_ar2s}
Agentic Real2Sim~\cite{chen2026agentic} builds a simulator scene from a recorded episode: vision-language agents (here the open Qwen3.6-35B-A3B) turn the episode into a MuJoCo scene, tracking the objects through the whole recorded stereo video, and the recorded actions are replayed.
We use the official released pipeline with its built-in MuJoCo renderer (the photorealistic renderer of their paper is not released), without the scene shift that the official version applies to make grasps succeed.
Unlike the world models, it thus sees the objects in future frames (our calibration uses robot masks only, though also from frames inside the window); converting one episode takes a median of $46$ minutes.
We ran it on $50$ of the $500$ candidate episodes; it converts $44$ (the $6$ failures come from the method itself), $27$ of them test episodes, on which \hyperref[tab:suppl_ar2s]{Table~\ref*{tab:suppl_ar2s}} compares it.
\hyperref[fig:teaser]{Fig.~\ref*{fig:teaser}} shows test episode 5375, one of these $27$, at step $92$; in its Agentic Real2Sim panel, the empty background of the render is replaced by the input first frame with the real robot and objects inpainted, while the robot and objects shown are those of the render.
Its arm follows the actions better than DreamDojo and Cosmos-Predict~2.5, since it replays them, but its rebuilt scene still misses objects and grasps (\hyperref[fig:suppl_ar2s]{Fig.~\ref*{fig:suppl_ar2s}}).
\begin{fixedpair}{0.5}{0.46}
\begin{table}[t]
\centering
\footnotesize
\setlength{\tabcolsep}{5pt}
\renewcommand{\arraystretch}{1.12}
\caption{\textbf{Agentic Real2Sim} on the $27$ test episodes it converts: motion against the real video (protocol of \hyperref[tab:droid_compare]{Table~\ref*{tab:droid_compare}}). Its flat-coloured, untextured MuJoCo rendering is not meant to look real, so image metrics are not compared. Best in bold.}
\label{tab:suppl_ar2s}
\tabstyle
\begin{tabular}{l||ccc}
\toprule
Method & Robot$\uparrow$ & Traj.$\uparrow$ & Depth$\downarrow$ \\
\tabdouble
Agentic Real2Sim (Qwen3.6) & .574 & .426 & .350 \\
\rowcolor{tabzebra} Ctrl-World & .666 & .630 & .178 \\
DreamDojo & .328 & .344 & .330 \\
\rowcolor{tabzebra} Cosmos-Predict~2.5 & .386 & .378 & .244 \\
\tabdash{1-4}
\rowcolor{tabours} SimPWM-H3-C & \textbf{.798} & .789 & .149 \\
\rowcolor{tabours} SimPWM-LTX-2.3 & .790 & \textbf{.830} & \textbf{.148} \\
\bottomrule
\end{tabular}
\end{table}
\pairnext
\begin{figure}[t]
\centering
\newcommand{\qs}[1]{\includegraphics[width=0.325\linewidth]{fig/qualitative/#1}}
\setlength{\tabcolsep}{1pt}
\renewcommand{\arraystretch}{0.6}
\caption{\textbf{Agentic Real2Sim} (its MuJoCo render composited on the input background, as in \hyperref[fig:teaser]{Fig.~\ref*{fig:teaser}}): wrong object motion.}
\label{fig:suppl_ar2s}
\vspace{4pt}
\begin{tabular}{@{}ccc@{}}
\scriptsize GT & \scriptsize Agentic Real2Sim & \scriptsize SimPWM-H3-C \\[2pt]
\qs{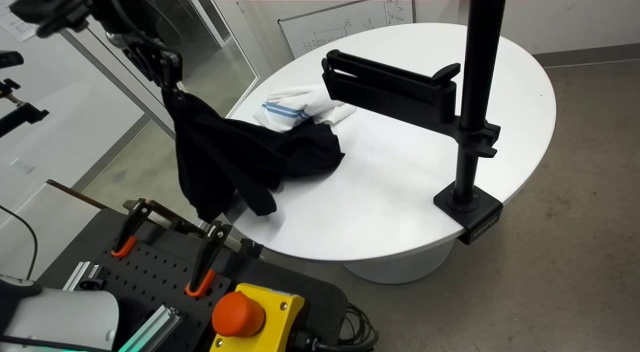} & \qs{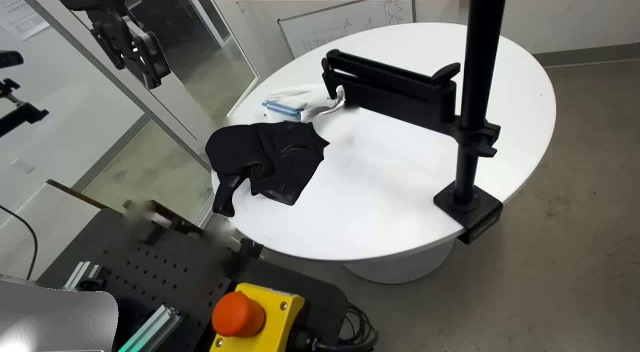} & \qs{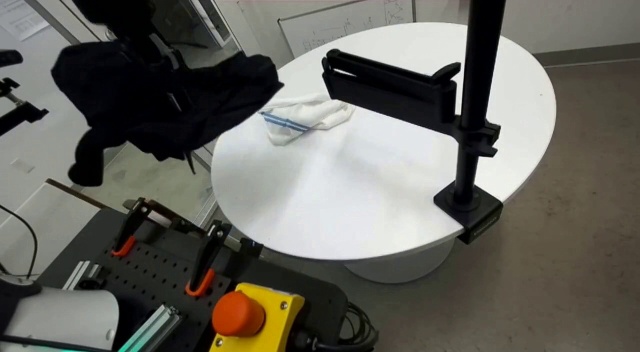} \\
\qs{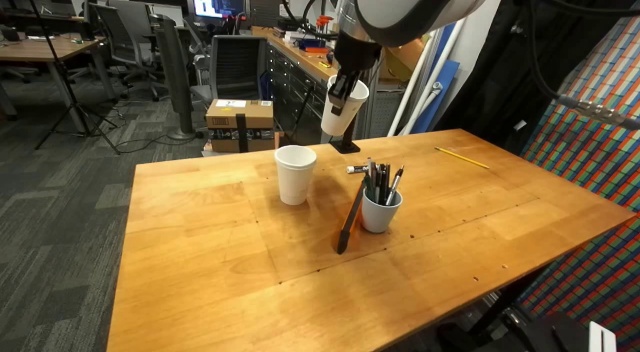} & \qs{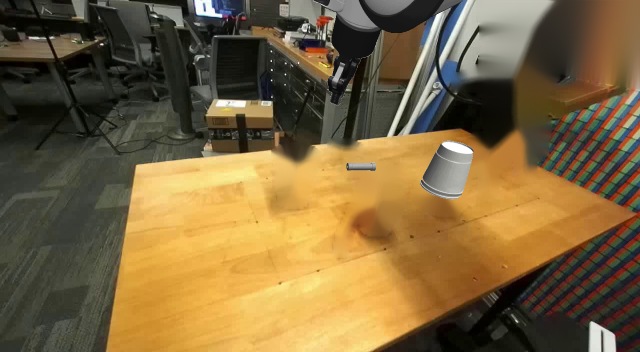} & \qs{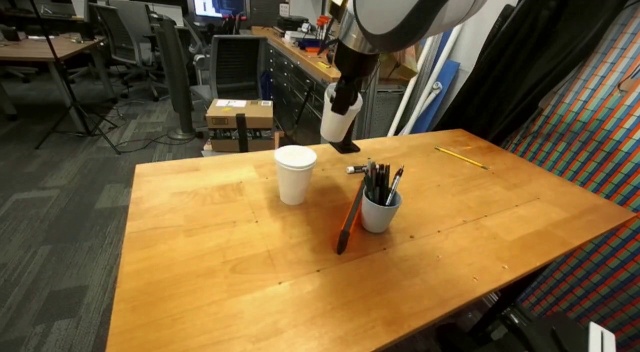} \\
\end{tabular}
\end{figure}
\end{fixedpair}
\section{Failure Cases}
\label{sec:failures}
\begin{sidecapfloat}[t]{0.56}
\begin{figure}[t]
\centering
\setlength{\tabcolsep}{1.2pt}
\newlength{\fw}\setlength{\fw}{\dimexpr(\linewidth-6\tabcolsep-5pt)/4\relax}
\newcommand{\fimg}[1]{\rimg[width=\fw]{fig/failures/#1}}
\renewcommand{\arraystretch}{0}
\begin{tabular}{@{}cc@{\hspace{5pt}}cc@{}}
\figlab{GT} & \figlab{Proxy} & \figlab{GT} & \figlab{Proxy} \\[2pt]
\fimg{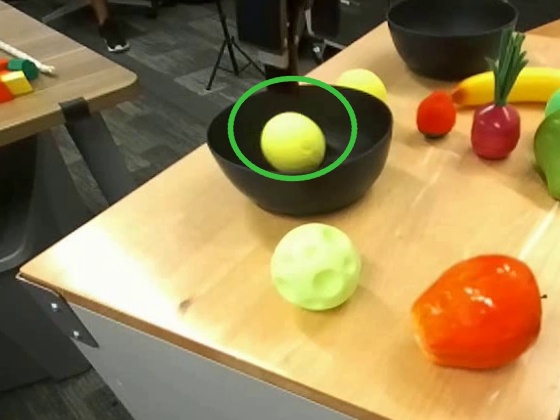} & \fimg{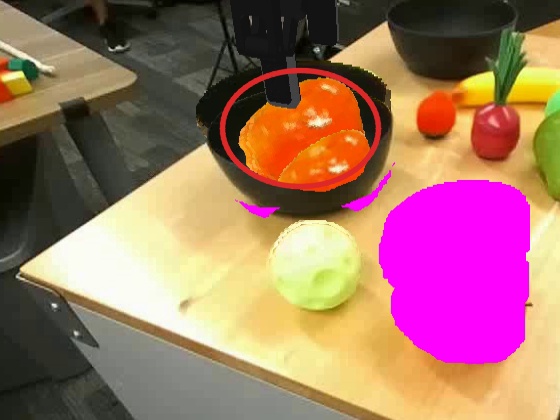} & \fimg{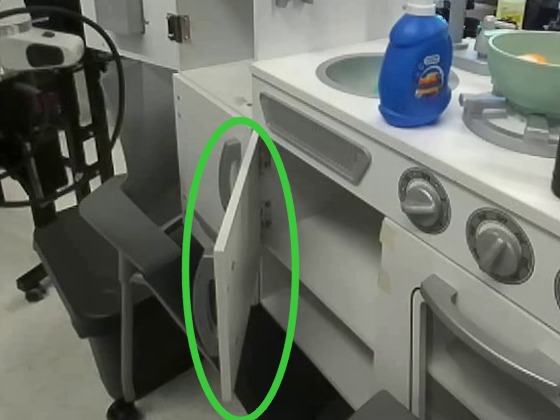} & \fimg{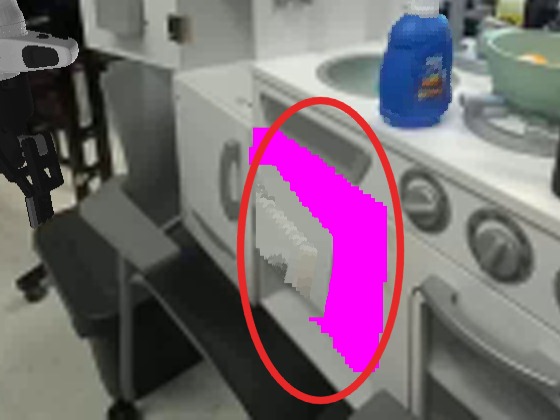} \\[4pt]
\figlab{First frame} & \figlab{GT} & \figlab{Proxy} & \figlab{GT, later} \\[2pt]
\fimg{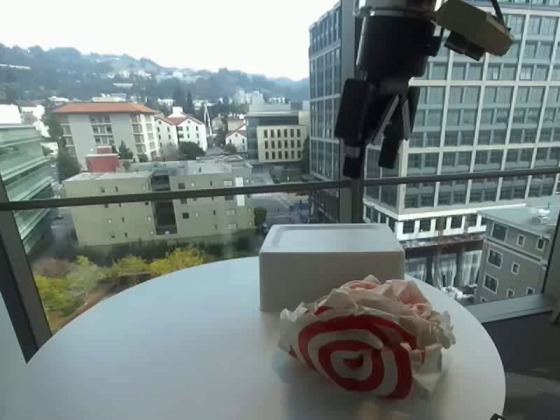} & \fimg{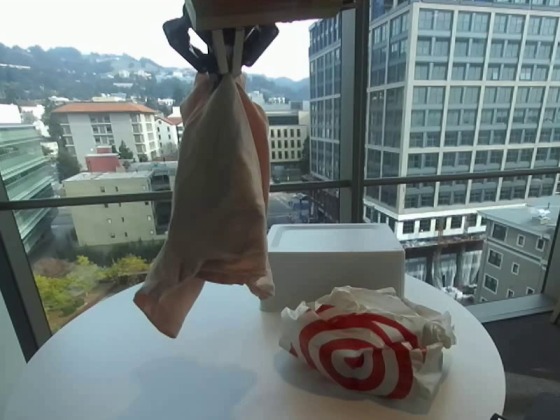} & \fimg{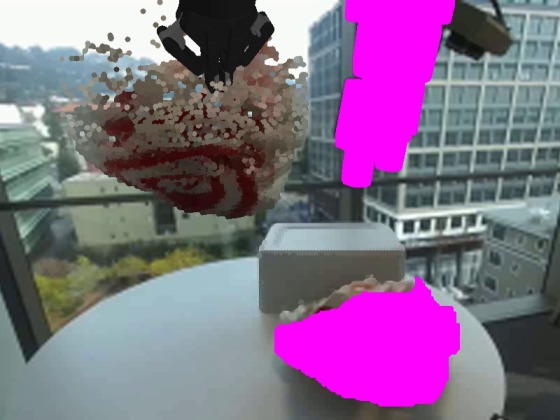} & \fimg{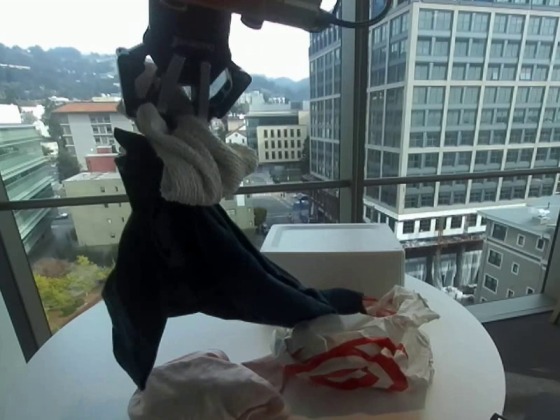} \\
\end{tabular}
\caption{\textbf{Failure cases.} Top left: a wrong object is grasped (red). Top right: the door opens (green) but stays shut in the proxy (red). Bottom: the robot pulls clothes out of a bag, whose contents the first frame does not show; the proxy lifts the bag itself.}
\label{fig:failures}
\end{figure}

\end{sidecapfloat}
\hyperref[fig:failures]{Figure~\ref*{fig:failures}} shows examples of the three main failure cases.
(1) \emph{Motion beyond pick-and-place.} Pushing, pouring, folding and articulation are not modelled; the video model adds folds but not motion that the proxy lacks (episode 5252: the robot swings a cabinet door open, which stays shut in the proxy). Object IoU on the right grasps is similar for the $110$ test tasks that name such an action and for pick-and-place ($0.47$ vs.\ $0.45$), as it measures only the carried object.
(2) \emph{Wrong grasps and placement.} A closing point a few pixels beside the target makes the proxy grasp a neighbour (episode 6115: the red pepper instead of the lemon);
of the $33$ audited episodes, the proxy grasps a wrong object in $9$ and a container instead of its content (or the reverse) in $1$, and $7$ stay unclear. Even with the right object, the carried object drifts from the real one (object IoU $0.46$).
(3) \emph{Unobserved objects.} The simulator holds only what the first frame shows: in training episode 2385 (bottom of \hyperref[fig:failures]{Fig.~\ref*{fig:failures}}), the clothes the robot pulls out of a bag are not in the first frame, so in the proxy the gripper lifts the bag itself.

\section{Other Robots}
\label{sec:suppl_robots}
In this section, SimPWM-LTX-2.3 denotes LTX-2.3 fine-tuned with our proxy on the data of the respective robot, not the DROID model of \hyperref[tab:droid_compare]{Table~\ref*{tab:droid_compare}}.
We test SimPWM, with the LTX-2.3 IC-LoRA, on two further robots: the full pipeline on the bimanual Aloha-AgileX of WorldArena (\hyperref[sec:suppl_worldarena]{Sec.~\ref*{sec:suppl_worldarena}}), and only the sim-to-real step on the GR1 humanoid (\hyperref[sec:suppl_hands]{Sec.~\ref*{sec:suppl_hands}}).

\subsection{Bimanual Robot: WorldArena Benchmark}
\label{sec:suppl_worldarena}

On the WorldArena~1.0 benchmark~\cite{shang2026worldarena} (the $1{,}000$ episodes of its challenge test set, from $50$ RoboTwin~2.0 bimanual tasks, each given a first frame and a $14$-D joint-action sequence), we compare with $18$ of its models (\hyperref[tab:compare]{Table~\ref*{tab:compare}}): SimPWM is best on $3$ of the $10$ metrics it is ranked on and weaker in image quality.
\hyperref[fig:suppl_wa_scatter]{Figure~\ref*{fig:suppl_wa_scatter}} places it among them: its trajectory accuracy is third, after GE-Sim~2.0~\cite{qiu2026gesim2} and GigaWorld-1~\cite{gigaworld2026gigaworld1}, which are pretrained on large real-robot corpora.
Baseline numbers are taken from the public leaderboard (the $14$ models of the benchmark paper and four further action-conditioned entries with a paper), not re-run. Action following is copied from the benchmark paper, which computes it on its own split, so we do not rank SimPWM on it.
We post-train our video model on all $2{,}500$ Clean-50 expert demonstrations of RoboTwin~2.0 ($50$ per task), i.e.\ both the $2{,}000$ training and the $500$ test episodes of the benchmark paper's $40/10$ split; the benchmark paper's baselines are post-trained on the $2{,}000$ training episodes only, so the comparison is not controlled for training data. We evaluate only on the separate $1{,}000$-episode challenge test set, none of whose action sequences matches any of these demonstrations (closest per-step deviation $0.02$\,rad). Only the robot configuration of the simulator changes (\hyperref[fig:suppl_wa]{Fig.~\ref*{fig:suppl_wa}}).

\begin{table*}[t]
\centering
\footnotesize
\setlength{\tabcolsep}{4.2pt}
\renewcommand{\arraystretch}{1.12}
\caption{\textbf{WorldArena~1.0 benchmark.} Official metrics (Interact., Persp.\ and Instr.\ with the benchmark's Qwen3-VL judge, unlike the Claude judge of \hyperref[tab:render]{Table~\ref*{tab:render}}); higher is better, best in bold; baselines from the public leaderboard (Track~1, 7 Nov.\ 2026); Wan~2.2 is cited by the Wan technical report~\cite{wan2025wan}. $^\dagger$From the benchmark paper, on its own $500$ test episodes, which are in our training data. $^\ddagger$Ours, on the $1{,}000$ challenge test episodes.}
\label{tab:compare}
\tabstyle
\begin{adjustbox}{max width=\textwidth}
\begin{tabular}{l||ccc|c|cc|cc|ccc}
\toprule
\multirow{2}{*}{\textbf{Method}} & \multicolumn{3}{c|}{\textbf{Visual quality}} & \textbf{Consistency} & \multicolumn{2}{c|}{\textbf{Physics}} & \multicolumn{2}{c|}{\textbf{3D}} & \multicolumn{3}{c}{\textbf{Control}} \\
\cmidrule(lr){2-4}\cmidrule(lr){6-7}\cmidrule(lr){8-9}\cmidrule(lr){10-12}\tabcmidend
 & Image$\uparrow$ & Aesth.$\uparrow$ & JEPA$\uparrow$ & Photom.$\uparrow$ & Interact.$\uparrow$ & Traj.$\uparrow$ & Depth$\uparrow$ & Persp.$\uparrow$ & Instr.$\uparrow$ & Sem.$\uparrow$ & Action$^\dagger\uparrow$ \\
\tabdouble
\tabgroup{12}{Text-conditioned embodied} \\
GigaWorld-0~\cite{gigaworld2025gigaworld0} & .4881 & .4015 & .3865 & .1284 & .5620 & .1707 & .6378 & .7806 & .5614 & .8664 & .1134 \\
\rowcolor{tabzebra} Genie Envisioner~\cite{liao2025genieenvisioner} & .2818 & .2639 & .3381 & .1604 & .2274 & .0263 & .8683 & .5432 & .2036 & .8598 & .0109 \\
TesserAct~\cite{zhen2025tesseract} & .3545 & .4130 & .3994 & .2031 & .6270 & .1639 & .7385 & .8426 & .6680 & .8868 & .0311 \\
\rowcolor{tabzebra} RoboMaster~\cite{fu2025robomaster} & .3605 & .3787 & .2983 & .2828 & .5480 & .1235 & .8377 & .7794 & .5114 & .8778 & .0352 \\
Vidar~\cite{feng2025vidar} & .4175 & .3987 & .6110 & .2057 & .6084 & .2126 & .7977 & .8188 & .6154 & .8846 & .0819 \\
\rowcolor{tabzebra} Cosmos-Predict~2.5 (text)~\cite{nvidia2025cosmospredict25} & .6549 & .4433 & .4669 & .3539 & .6514 & .1160 & .7071 & .9822 & .6470 & .8634 & \textbf{.1418} \\
WoW~\cite{chi2025wow} & .4012 & .3659 & .9169 & .0579 & .7071 & .2763 & .7890 & .8827 & .8100 & .8847 & .0434 \\
\tabdouble
\tabgroup{12}{General video models} \\
CogVideoX~\cite{yang2025cogvideox} & .3623 & .3630 & .9484 & .3379 & .6696 & .3479 & .9109 & .8546 & .7558 & .8970 & .0076 \\
\rowcolor{tabzebra} Wan~2.2~\cite{wan2025wan} & .4232 & .3953 & .8805 & .3162 & .7262 & .2650 & .8591 & .9262 & .8110 & .8878 & .0512 \\
Wan~2.6~\cite{alibaba2025wan26} & \textbf{.6736} & .4440 & .7257 & .3457 & .7316 & .1218 & .7543 & .9394 & .8996 & .8809 & .0992 \\
\rowcolor{tabzebra} Veo~3.1~\cite{google2025veo31} & .6557 & \textbf{.4879} & .5797 & .3752 & \textbf{.8150} & .1136 & .7427 & \textbf{.9964} & \textbf{.9714} & .8379 & .0852 \\
\tabdouble
\tabgroup{12}{Action-conditioned} \\
IRASim~\cite{zhu2024irasim} & .4185 & .3286 & .9372 & .4570 & .6276 & .3592 & .9350 & .8416 & .6540 & .8943 & .0526 \\
\rowcolor{tabzebra} Ctrl-World~\cite{guo2025ctrlworld} & .4244 & .3705 & .9277 & .1288 & .6262 & .4820 & .9325 & .8366 & .6768 & .8868 & .0210 \\
Cosmos-Predict~2.5 (action)~\cite{nvidia2025cosmospredict25} & .4948 & .3414 & .9065 & .4521 & .6046 & .2749 & .9035 & .8468 & .6054 & .8986 & .0133 \\
\rowcolor{tabzebra} DreamDojo~\cite{gao2026dreamdojo} & .3319 & .3653 & .7157 & .3534 & .4996 & .1514 & .8733 & .7616 & .4352 & .8884 & -- \\
WoVR~\cite{jiang2026wovr} & .3960 & .3490 & .9581 & \textbf{.4651} & .6456 & .4408 & .9608 & .8456 & .6998 & .8984 & -- \\
\rowcolor{tabzebra} GigaWorld-1~\cite{gigaworld2026gigaworld1} & .5118 & .4117 & .9677 & .2811 & .7510 & .5427 & .9844 & .9560 & .8214 & .8942 & -- \\
GE-Sim~2.0~\cite{qiu2026gesim2} & .4673 & .3907 & .9716 & .0705 & .7360 & \textbf{.5863} & .9649 & .9364 & .8270 & .8905 & -- \\
\tabdouble
\rowcolor{tabours} \textbf{SimPWM-LTX-2.3 (ours)} & .3841 & .3741 & \textbf{.9874} & .3862 & .7118 & .5271 & \textbf{.9895} & .9044 & .8128 & \textbf{.9031} & .0239$^\ddagger$ \\
\bottomrule
\end{tabular}
\end{adjustbox}
\end{table*}
\begin{sidecapfloat}[t]{0.6}
\begin{figure}[t]
\centering
\includegraphics[width=0.8\linewidth]{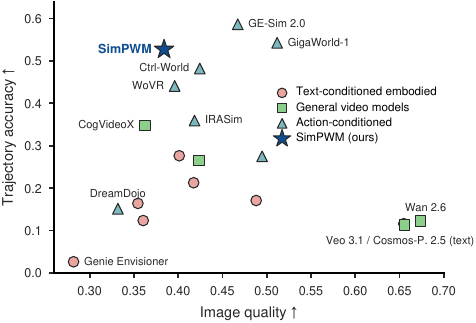}
\caption{\textbf{WorldArena: image quality against trajectory accuracy} for SimPWM-LTX-2.3 and the $18$ models of the WorldArena benchmark (all metrics in \hyperref[tab:compare]{Table~\ref*{tab:compare}}). SimPWM follows the trajectories more closely than all models but GE-Sim~2.0 and GigaWorld-1, at a lower image quality; Wan~2.6 and Veo~3.1 have the best image quality but follow the trajectories poorly.}
\label{fig:suppl_wa_scatter}
\end{figure}
\end{sidecapfloat}
\newlength{\waw}\setlength{\waw}{\dimexpr(384.06pt-8pt)/6\relax}%
\newcommand{\waimg}[1]{\includegraphics[width=\waw,trim=0 14 0 42,clip]{fig/worldarena/#1.jpg}}%
\newcommand{\walab}[1]{\rotatebox{90}{\makebox[0.575\waw]{\scriptsize #1}}}
\newcommand{\waline}[3]{\walab{#3} & \waimg{#1_s0_#2} & \waimg{#1_s1_#2} & \waimg{#1_s2_#2} & \waimg{#1_s3_#2} & \waimg{#1_s4_#2} & \waimg{#1_s5_#2} \\[1pt]}
\newcommand{\warow}[2]{\waline{#1}{gt}{GT} \waline{#1}{px}{RGB proxy} \waline{#1}{gen}{\textbf{SimPWM}}\multicolumn{7}{c}{\rule{0pt}{9pt}\figdesc{#2}} \\[4pt]}
\begin{figure*}[tbp]
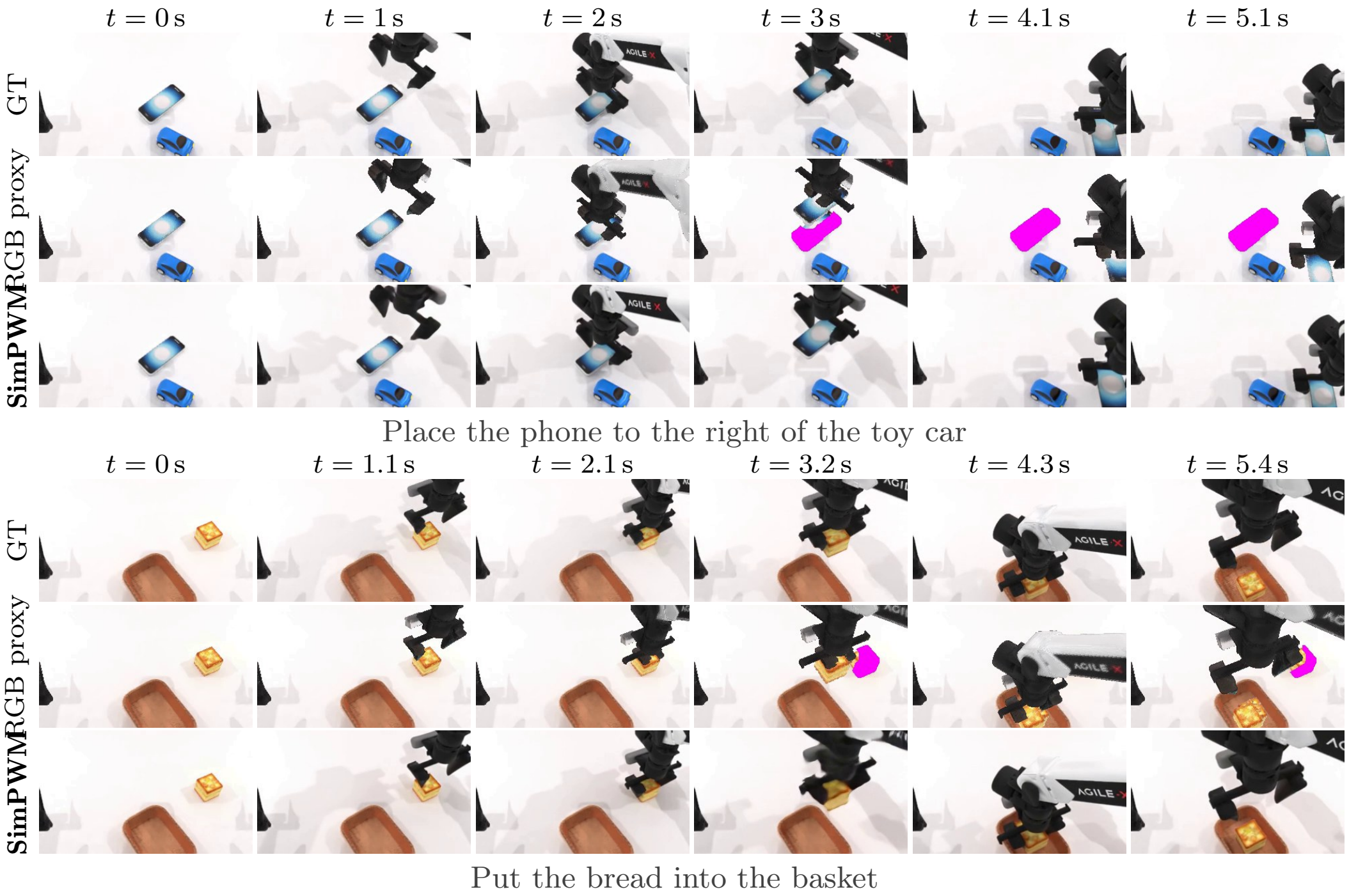

\centering
\setlength{\tabcolsep}{0.8pt}
\renewcommand{\arraystretch}{0}
\hspace*{-13.6pt}\begin{tabular}{@{}c@{\hspace{3pt}}cccccc@{}}
& {\scriptsize $t=0$\,s} & {\scriptsize $t=1$\,s} & {\scriptsize $t=2$\,s} & {\scriptsize $t=3$\,s} & {\scriptsize $t=4.1$\,s} & {\scriptsize $t=5.1$\,s} \\[2pt]
\warow{524}{Place the phone to the right of the toy car}
& {\scriptsize $t=0$\,s} & {\scriptsize $t=1.1$\,s} & {\scriptsize $t=2.1$\,s} & {\scriptsize $t=3.2$\,s} & {\scriptsize $t=4.3$\,s} & {\scriptsize $t=5.4$\,s} \\[2pt]
\warow{415}{Put the bread into the basket}
\end{tabular}
\caption{\textbf{SimPWM-LTX-2.3 on WorldArena} test episodes: real video (GT), the simulator\textquotesingle s RGB proxy, and the generated video, from the first frame ($t=0$) to the end of each episode (magenta: unseen). The video model fills in what the proxy lacks, such as the unseen regions and the appearance of moved objects.}
\label{fig:suppl_wa}
\end{figure*}
\begin{figure*}[tbp]
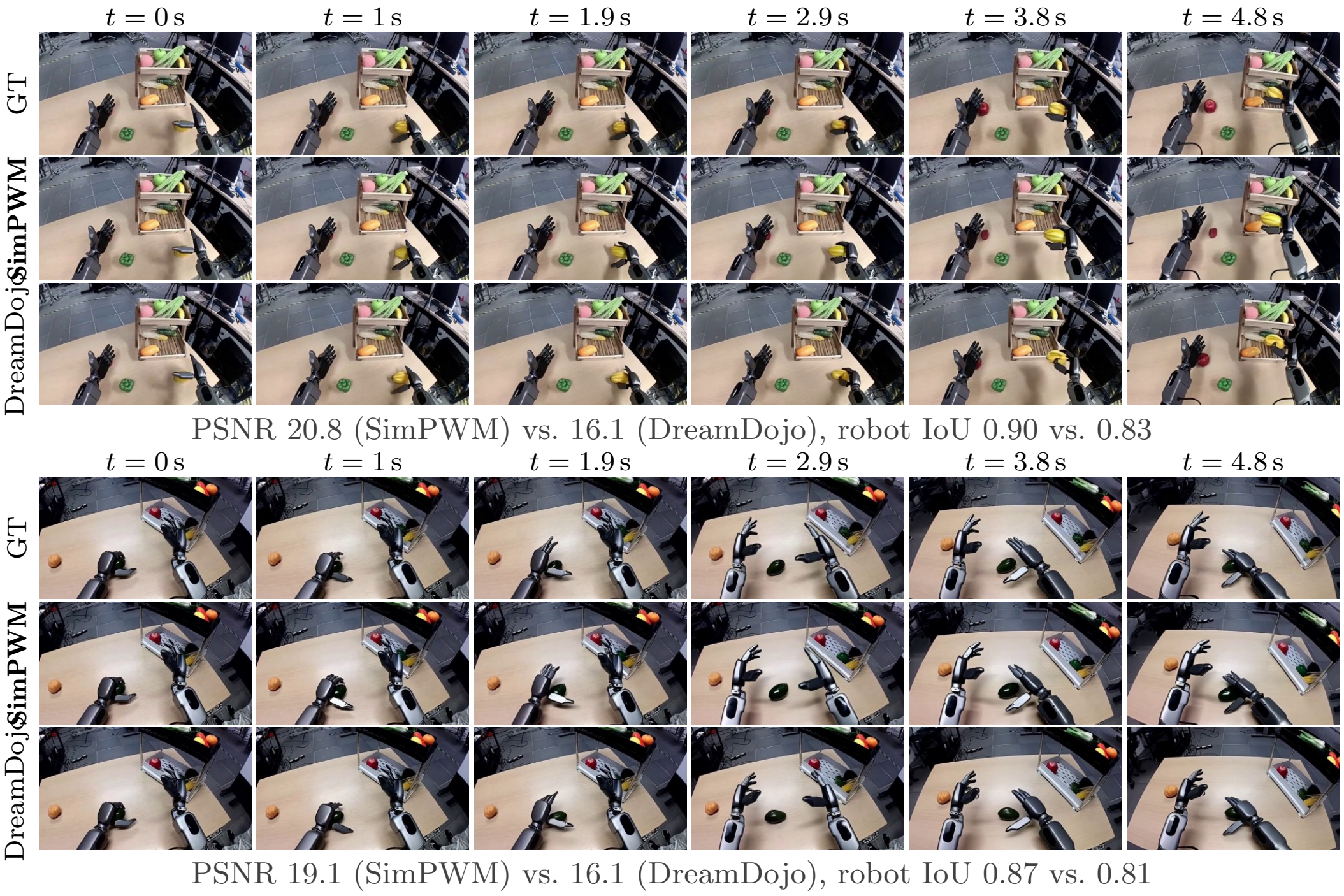

\centering
\setlength{\tabcolsep}{0.8pt}
\renewcommand{\arraystretch}{0}
\newlength{\grw}\setlength{\grw}{\dimexpr(384.06pt-8pt)/6\relax}%
\newcommand{\grimg}[1]{\includegraphics[width=\grw,trim=0 14 0 42,clip]{fig/gr1/#1.jpg}}%
\newcommand{\grlab}[1]{\rotatebox{90}{\makebox[0.575\grw]{\scriptsize #1}}}
\newcommand{\grline}[3]{\grlab{#3} & \grimg{#1_s0_#2} & \grimg{#1_s1_#2} & \grimg{#1_s2_#2} & \grimg{#1_s3_#2} & \grimg{#1_s4_#2} & \grimg{#1_s5_#2} \\[1pt]}
\newcommand{\grex}[2]{\grline{#1}{gt}{GT} \grline{#1}{ours}{\textbf{SimPWM}} \grline{#1}{dd}{DreamDojo}\multicolumn{7}{c}{\rule{0pt}{9pt}\figdesc{#2}} \\[4pt]}
\hspace*{-13.6pt}\begin{tabular}{@{}c@{\hspace{3pt}}cccccc@{}}
& {\scriptsize $t=0$\,s} & {\scriptsize $t=1$\,s} & {\scriptsize $t=1.9$\,s} & {\scriptsize $t=2.9$\,s} & {\scriptsize $t=3.8$\,s} & {\scriptsize $t=4.8$\,s} \\[2pt]
\grex{w072}{PSNR $20.8$ (SimPWM) vs.\ $16.1$ (DreamDojo), robot IoU $0.90$ vs.\ $0.83$}
& {\scriptsize $t=0$\,s} & {\scriptsize $t=1$\,s} & {\scriptsize $t=1.9$\,s} & {\scriptsize $t=2.9$\,s} & {\scriptsize $t=3.8$\,s} & {\scriptsize $t=4.8$\,s} \\[2pt]
\grex{w024}{PSNR $19.1$ (SimPWM) vs.\ $16.1$ (DreamDojo), robot IoU $0.87$ vs.\ $0.81$}
\end{tabular}
\caption{\textbf{GR1 test windows}: real video (GT), SimPWM-LTX-2.3 and DreamDojo at six points of each $4.8$\,s window, for two of the windows in which SimPWM scores clearly higher than DreamDojo (selected by PSNR, SSIM, LPIPS and robot IoU; \hyperref[tab:suppl_gr1]{Table~\ref*{tab:suppl_gr1}} gives the averages over all $100$ windows).}
\label{fig:suppl_gr1}
\end{figure*}
\subsection{Dexterous Hands: GR1}
\label{sec:suppl_hands}

\label{sec:suppl_gr1}
\begin{fixedpair}{0.62}{0}
\begin{table}[t]
\centering\footnotesize\setlength{\tabcolsep}{4pt}\renewcommand{\arraystretch}{1.08}
\caption{\textbf{GR1 video results} ($100$ test windows, DreamDojo's protocol; FVD on $100$ clips is only indicative). Best in bold.}
\label{tab:suppl_gr1}
\tabstyle
\begin{adjustbox}{max width=\linewidth}
\begin{tabular}{l||ccccc}
\toprule
Method & PSNR$\uparrow$ & SSIM$\uparrow$ & LPIPS$\downarrow$ & FVD$\downarrow$ & Robot IoU$\uparrow$ \\
\tabdouble
DreamDojo~\cite{gao2026dreamdojo} & 17.95 & .631 & \textbf{.261} & \textbf{147.4} & .815 \\
\rowcolor{tabours} SimPWM-LTX-2.3 & \textbf{18.23} & \textbf{.656} & .272 & 177.7 & \textbf{.861} \\
\bottomrule
\end{tabular}
\end{adjustbox}
\end{table}
\end{fixedpair}
We also train the video model on a humanoid with two dexterous hands, the GR1 robot of the GR00T teleoperation data~\cite{nvidia2025gr00t} (head camera, $1280{\times}800$ at $20$\,fps), with a proxy of the robot and the first frame but no simulated objects, so this experiment tests only the sim-to-real step.
The head camera is calibrated per session from SAM~3 robot masks, and the model is trained on $3{,}000$ episodes from $450$ sessions with the recipe of \hyperref[sec:suppl_impl]{Sec.~\ref*{sec:suppl_impl}}, at $10$\,fps and $800{\times}480$.
On DreamDojo's $100$ test windows, against its official GR1 checkpoint (\hyperref[tab:suppl_gr1]{Table~\ref*{tab:suppl_gr1}}, \hyperref[fig:suppl_gr1]{Fig.~\ref*{fig:suppl_gr1}}), SimPWM follows the arms more closely, while DreamDojo, post-trained on about seven times as many GR1 episodes, has the better LPIPS and FVD.
As in DreamDojo's protocol, the test windows come from sessions also seen in training.

\end{document}